# LEARNING STRUCTURED REPRESENTATIONS OF VISUAL SCENES

**MENG-JIUN CHIOU**

*(B.Sc., National Chiao Tung University)*

A THESIS SUBMITTED FOR THE DEGREE OF

DOCTOR OF PHILOSOPHY

DEPARTMENT OF COMPUTER SCIENCE

SCHOOL OF COMPUTING

NATIONAL UNIVERSITY OF SINGAPORE

2022

Supervisor:
Professor Roger Zimmermann

Examiners:
Professor Chua Tat Seng
Associate Professor Ng Teck Khim

# Declaration

I hereby declare that this thesis is my original work and it has been written by me in its entirety. I have duly acknowledged all the sources of information which have been used in the thesis.

This thesis has also not been submitted for any degree in any university previously.

Meng-Jiun Chiou

24 May 2022

*To my beloved family*

# Acknowledgments

My Ph.D. journey has been a wild ride. It has been motivational, exciting and promising, but it has also been tiring, despairing and worrying. I am extremely grateful to my supervisor, Prof. Roger Zimmermann, who is very wise and kind and has guided me throughout the whole academic journey. Without his support and advice, especially on the way of doing research and academic writing/thinking, it would not be possible for me to complete many of the projects in this thesis. I also greatly appreciate my co-supervisor, Prof. Jiashi Feng, who is very knowledgeable in the subject of this thesis and has also constantly supported me. His advice and comments are at the very core of many of the studies in this thesis. I also would like to thank many of the collaborators, labmates of both the Media Management Research Lab (MMRL) and Learning and Vision Lab (LV Lab), and my friends, including Hanshu Yan, Junhao Liew, Xiao Ma, Devamanyu Hazarika, Yifang Yin, Zhenguang Liu, Henghui Ding, Chun-Yu Liao, and Li-Wei Wang, who have helped me during this journey. Last but not least, I sincerely thank Prof. Chua Tat-Seng and Prof. Ng Teck Khim for their insightful comments on both my qualifying exam report and this thesis.

This thesis is dedicated to my parents, who have been continuously supporting me throughout my life with whatever decisions I made, and my wife, who always motivated me whenever I was overwhelmed by negative emotions.

Finally, I thank staffs and professors at the School of Computing and the Department of Electrical and Computer Engineering, including but not limited to Ms. Loo Line Fong, Prof. Angela Yao and Mr. Hoon Keng Chuan (Francis) who have in some ways helped me, and the National University of Singapore as a whole, who offered me this invaluable, once-in-a-lifetime opportunity to pursue the Ph.D. degree.

# Contents

# Abstract

Learning Structured Representations of
Visual Scenes

by

Meng-Jiun Chiou

Doctor of Philosophy in Computer Science

National University of Singapore

Recent advances in deep learning, large-scale data, and significantly more powerful computing abilities have brought numerous breakthroughs in computer vision in the past few years. For instance, machines have achieved near-human performance, or even outperform humans, in certain lower-level visual recognition tasks including image classification, segmentation, and object detection. However, for other higher-level vision tasks requiring a more detailed understanding of visual contents, such as visual question answering (VQA) and visual captioning (VC), machines still lag behind humans. This is partly because, unlike human beings, machines lack the ability to establish a comprehensive, structured understanding of the contents, on which reasoning can be performed. To be specific, higher-level vision tasks are usually overly simplified by operating models directly on images and are tackled by end-to-end neural networks without taking the compositional semantics of scenes into account. It has been shown that deep neural networks based models sometimes make serious mistakes caused by taking shortcuts learned from biased datasets. Moreover, the "black-box" nature of neural networks means their predictions are barely explainable, which is unfavorable for visual reasoning tasks like VQA. As the intermediate-level representations bridging the two levels, structured representations of visual scenes, such as *visual relationships* between pairwise objects, have been shown to not only benefit compositional models in learning to reason along with the structures but provide higher interpretability for model decisions. Nevertheless, these representations receive much less attention than traditional recognition tasks, leaving numerous open challenges (*e.g.*, imbalance predicate class problem) unsolved.

In the thesis, we study how to describe the content of the individual image or video with visual relationships as the structured representations. A visual relationship between two objects (`subject` and `object`, respectively) is defined by a triplet of the form (`subject`, `predicate`, `object`), which include subject's and object's bounding boxes and category labels along with the predicate label. The triplet form of visual relationships naturally resembles how humans describe the interaction between two objects with a language sentence: an adverb (predicate) connects a subject to an object, *e.g.*, "person is sitting on a chair" is represented by (`person`, `sitting on`, `chair`). To establish a holistic representation of a scene, a graph structure named *scene graph* constructed with visual objects as nodes and predicates as directed edges is usually utilized to take object and relation contexts into account. For instance, *e.g.*, "person sitting on a chair is holding a glass" can be represented by (`person`, `sitting on`, `chair`) and (`person`, `holding`, `glass`) where the `persons` in the two visual relation triplets refer to the same entity.

In the first part of the thesis, we spend two chapters on learning structured representations of images with visual relationships and scene graphs, which are formulated as Visual Relationship Detection (VRD) and Scene Graph Generation (SGG), respectively. First, we delve into how to incorporate external knowledge to perform VRD. Inspired by the recent success of pretrained representations, we propose a Transformer-based multimodal model which recognizes visual relations with both visual and language knowledge learned via pretraining on the large-scale corpus. The proposed model is also equipped with an effective spatial module and a novel mask attention module to explicitly capture spatial information among the objects. These designs are shown to benefit VRD and help the model achieve competitive results on two challenging VRD datasets. Second, we re-think the role of datasets' knowledge and argue that some of them are "bad" knowledge bringing in biases for predicting visual relationships and should be removed. Specifically, we tackle the critical data imbalance problem from a novel perspective of *reporting bias*, which arises from datasets itselves and causes machines to prefer easy predicates such as (`person`, `on`, `chair`) or (`bird`, `in`, `room`), to more informative predicates (`person`, `sitting on`, `chair`) or (`bird`, `flying in`, `room`). To remove this reporting bias, we develop a model-agnostic debiasing method for generating more informative scene graphs by taking the chances of predicate classes being labeled into account. Also, we shift the focus from VRD to SGG to generate holistic, graph-structured representations and leverage message passing networks for incorporating the contexts. Extensive experiments show that our approach significantly alleviates the long tail, achieves state-of-the-art SGG debiasing performance, and produces prominently more fine-grained scene graphs.

In the second part of the thesis, we extend the static-image VRD setting into temporal domain and consider human-object interaction (HOI) detection – a special case of VRD where the subjects of visual relationships are restricted to humans. Conventional HOI methods operating on only static images have been used by researchers to predict temporal-related HOIs in videos; however, in this way models neglect temporal contexts and may provide sub-optimal performance. Another related task, video visual relationship detection (VidVRD), is also not a suitable setting as i) VidVRD methods neglect human-related features in general, ii) video object detection remains challenging and iii) action boundaries labeling itself might be inconsistent. We thus propose to bridge these gaps by explicitly considering temporal information and adopting a keyframe-based detection for video HOI detection. We also show that a naive temporal-aware variant of a common action detection baseline underperforms in video-based HOIs due to a feature-inconsistency issue. We then propose a novel, neural network-based model utilizing temporal information such as human and object trajectories, frame-wise localized visual features, and spatial-temporal masked human pose features. Experiments show that our approach is not only a solid baseline in our proposed video HOI benchmark, but also a competitive alternative in a popular video relationship detection benchmark.

Overall, in these works, we explore how structured representations of visual scenes can be effectively constructed and learned in both the static-image and video settings, with improvements resulting from external knowledge incorporation, bias-reducing mechanism, and/or enhanced representation models. At the end of this thesis, we also discuss some open challenges and limitations to shed light on future directions of structured representation learning for visual scenes.

# List of Figures

# List of Tables

# Chapter 1

# Introduction

Thanks to recent advances in deep learning [81, 84], machines are able to achieve near-human performance in some computer vision tasks that requires machines to *recognize* objects, *e.g.*, image classification [52, 22, 21]. Some researchers have thus been shifting focus to other *higher-level* vision tasks requiring machines to, based on object-level representations [101, 23], *understand* scenes. Some examples of scene understanding tasks are visual question answering (VQA) [2] and visual captioning (VC) [101, 184]. Early approaches to VQA and VC encode visual inputs into either a global Convolutional Neural Networks (CNNs) features [158, 73], object-based CNNs features [1], or attention-weighted grids of CNNs features [175, 124, 111]. Although these features are proven effective, some researchers [1] argue that they do not correspond to the human visual system where salient objects or regions form a more natural basis for human attention [35, 129], and they thus propose to utilize bottom-up (object-based) attention over CNNs features. However, it has still been shown that these deep neural networks (DNNs) based models make mistakes on questions that are obvious to humans [116, 55], partly because they tend to incorrectly exploit dataset biases (*e.g.*, co-occurrence) as shortcuts to answer the questions that require *reasoning* [64, 43, 70]. For instance, given a VQA query comprising a scene where a `book` is put on a `table` and a `glass` under the `table`, a naive VQA model that was trained with many examples of a `glass` on a `table` would still prefer answering `glass` is on the `table`. Note that this issue cannot be addressed by the above-mentioned (naive) visual encoding methods. What is worse is that attention-based approaches still lack interpretability: while attention maps visualized with input images somehow show *where* the model was looking at, it is still hard to tell *why* machines make the prediction.

To deal with the aforementioned issues, researchers draw inspiration from text-only QA [58] where questions are usually decomposed into logical expressions that can be evaluated against a structured representation of the world (*e.g.*, knowledge bases) [94]. One plausible remedy to VQA models' lack of reasoning ability is thus to represent visual scenes with structured representations on which logical expressions can be performed. *Scenes* are collections of salient objects or regions annotated with attributes such as category, color, material, texture, shape, etc. [70]. *Scene graphs* [71, 109] take one step further to also take into account visual relationships among objects to represent scenes as directed graph structures, where nodes are objects and edges are object pair-wise relationships (predicates). Note that visual relationships in the form of triplets (`subject,predicate,object`) are analogous to relational facts in knowledge base (*e.g.*, WordNet [114]). Scene graphs have been shown as effective intermediate representations to benefit downstream tasks including VQA [136, 198, 95] and VC [182, 181, 176] in terms of both performance and interpretability.

While visual relationship detection (VRD) and scene graph generation (SGG) [109, 88, 29, 193, 118, 187, 183, 191, 179, 57, 15, 16, 45, 13, 34, 189, 74, 148, 102, 164, 177, 77, 53, 165, 190, 134, 185, 122, 167, 62, 152, 188] have been extensively studied over the past few years, how to generate rich and unbiased scene graphs remains to be challenging. First, considering the increasing larger-scale vision and language datasets [142, 201] and the success of pre-training BERT-style [156, 31] models for multi-modal downstream tasks [110, 85, 17], a natural, but still an open question is that how we can exploit these external visual-linguistic knowledge to enhance VRD performance. Second, we shift the focus to SGG for more holistic, structured representations, and we re-think the role of datasets' knowledge and argue that some of them are "bad" prior knowledge bringing in unwanted biases. Concretely, we deal with a critical issue – the long-tail problem [42, 138, 26] which causes a system to prefer frequent classes to less frequent ones. We note that the long tail in SGG is actually different from the one in common long-tailed recognition tasks, in that the former is significantly affected by the imbalance of missing labels in existing datasets (*i.e.*, "bad" labeling bias), instead of the unbalanced class prior distributions [148]. However, existing SGG debiasing approaches [45, 34, 102, 148, 177, 164] do not take this mismatch into account, resulting in sub-optimal debiasing performance. We briefly explain these VRD and SGG challenges we solve in Section 1.1.1, and we dive deeper into them in Chapter 3 and 4.

While some visual relationships (especially, spatial predicates such as `on` and `in`) can be easily understood from a single image, others (especially, temporal predicates such as `open` and `close`) might require temporal information to be distinguished. As humans are usually the focus in video clips, we are interested in how do machines construct human-centered structured representations in videos. Video visual relationship detection (VidVRD) [133, 131, 121, 144, 103, 14, 151] pushes VRD forward to focus on detecting visual relations in videos. However, existing VidVRD approaches detect general visual relations among all kinds of objects, omitting human-centered features (*e.g.*, human poses, body meshes) which could be important for human-related visual relation detection. Moreover, compared to VidVRD's video-level annotations, frame-level atomic labels like AVA dataset [44] might be more suitable as i) action boundaries labeling could be inconsistent due to inter-annotator disagreement [63, 44], and ii) accurate video object detection [98] remains challenging. On the other hand, image-based human-object interaction detection (HOID) [141, 11, 39, 120, 92, 162, 159, 49, 91, 163, 97, 155, 199], which can be regarded as a special case of VRD where the subjects' category in visual relation triplets are restricted to `human`, has been extensively studied over the past years. While conventional HOI detection methods operating on only static images have been used to predict temporal-related interactions (*e.g.*, [159]), their models are essentially guessing without temporal contexts, leading to sub-optimal performance. In addition, we note that the existing works in video HOI retrieval (VidHOIR) [78, 66, 120, 145] only focus on HOI tagging without spatial-temporal localization, failing to represent video scenes in a structured manner. Despite these related efforts, there still lacks a study on how machines can learn to predict HOIs in videos in a human feature-aware and spatial-temporal localization-aware manner. We briefly discuss this gap in the following Section 1.1.1 and delve into it in Chapter 5.

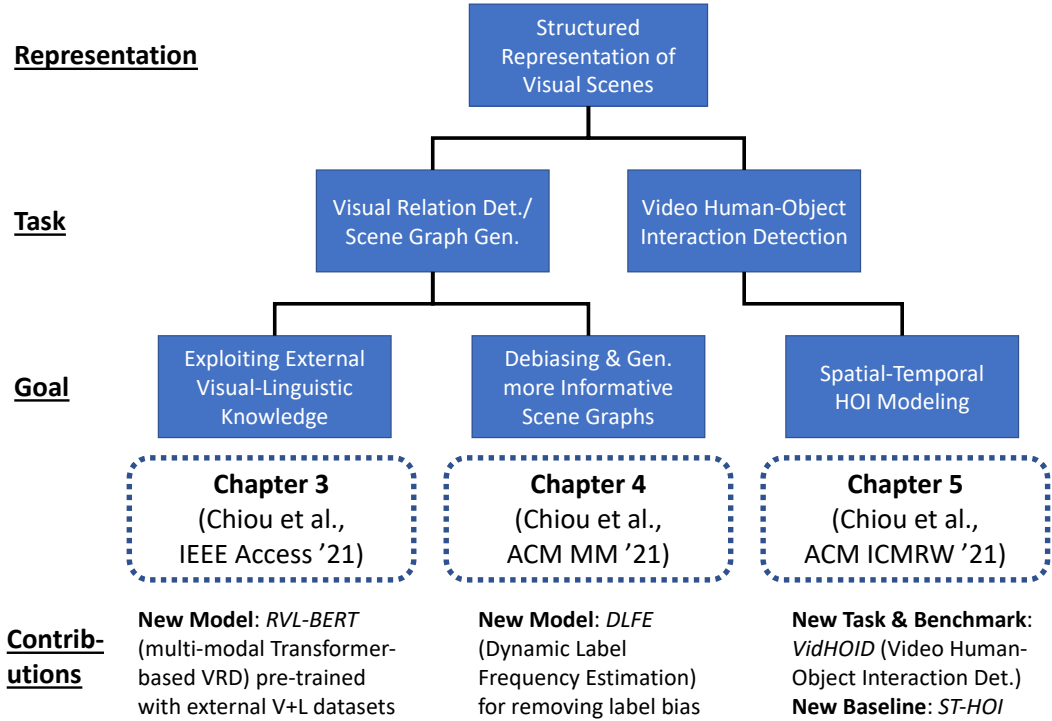

Figure 1.1: Overview of the thesis.

## 1.1 Thesis Overview

Figure 1.1 presents an overview of this thesis's scope and contributions, with respect to the aforementioned research challenges. We lay the common ground of our studies by presenting a comprehensive background on learning structured representations of visual scenes in Section 2. Then in the beginning of each of the following chapter, we discuss in-depth some related works that are especially important to each of the research challenge. In general, we present a collection of enriching and debiasing methods to learn structured visual representations, with VRD and SGG as the applications, in Chapter 3 and 4, respectively. We then present a pioneering study on learning to predict HOI in videos in a human-center and spatial-temporal localization-aware mannger in Chapter 5. Lastly, we present conclusions and future directions of learning structured visual representations in Chapter 6.

### 1.1.1 Problem Statement and Contributions

We briefly summarize the challenges we encounter in learning structured representations of visual scenes and how we address them in this section.

- Inspired by human reasoning mechanisms, it is believed that external visual-linguistic knowledge is beneficial for reasoning visual relationships of objects in images, which is however rarely considered in existing VRD methods [45]. Given the success of pretraining multi-modal Transformers on large-scale datasets [85, 110], we propose a novel model named *Relational Visual-Linguistic Bidirectional Encoder Representations from Transformers* (RVL-BERT), which performs relational reasoning with both vision-and-language knowledge learned via self-supervised pre-training with multimodal representations. RVL-BERT also utilizes an effective spatial module and a novel mask attention module to explicitly capture spatial information among salient objects (Chapter 3).

- Increasing efforts have been paid to the long tail problem in SGG [16, 148, 177, 164, 167, 53]; however, the imbalance in the fraction of missing labels of different classes, or labeling bias [148] including *reporting bias* [115] and *bounded rationality* [139], exacerbating the long tail is rarely considered and cannot be solved by the existing debiasing methods. Notably, while we show in Chapter 3 that (external) knowledge can be helpful, "bad" dataset knowledge such as the labeling bias could be detrimental. We show that, due to the missing labels, SGG can be viewed as a *Learning from Positive and Unlabeled data* (PU learning) [30, 36, 6] problem, where the labeling bias can be removed by recovering the unbiased probabilities from the biased ones by utilizing *label frequencies*, *i.e.,* the per-class fraction of labeled, positive examples in all the positive examples. We then propose a debiasing method based on unbiased probability recovery to alleviate the long tail and generate prominently more informative scene graphs (Chapter 4).

- While detecting non-temporal HOIs (*e.g.*, *sitting on* a chair) from static images is feasible, it is unlikely even for humans to infer temporal-related HOIs (*e.g.*, *opening/closing* a door) from a single video frame. However, conventional HOID methods (*e.g.*, the baseline in [159] operating on only static images have been used to predict temporal-related interactions, which is essentially guessing without temporal contexts, leading to sub-optimal performance. In addition, efforts in other video tasks like VidVRD and VidHOIR might not be applied straightforwardly, due to the lack of attention to humans and the video-level annotations, and spatial-temporal object localization, respectively. We bridge this gap by detecting video-based HOIs with explicit temporal information. We show that a naive temporal-aware variant of a common action detection baseline does not work on video-based HOIs due to a feature-inconsistency issue. We then propose *Spatial-Temporal HOI Detection* (ST-HOI) utilizing temporal information such as human and object trajectories, correctly-localized visual features, and spatial-temporal masking pose features. We also construct a new video HOI benchmark dubbed VidHOI where our proposed approach serves as a solid baseline (Chapter 5).

### 1.1.2 Publications

All the works presented in this thesis have been published in the following peer-reviewed journals or conference proceedings (and are all open-sourced):

- **M.-J. Chiou**, R. Zimmermann, and J. Feng, "Visual Relationship Detection with Visual-Linguistic Knowledge from Multimodal Representations", *IEEE Access*, vol. 9, pp. 50441–50451, 2021.[1]

- **M.-J. Chiou**, H. Ding, H. Yan, C. Wang, R. Zimmermann, and J. Feng, "Recovering the Unbiased Scene Graphs from the Biased Ones", in *Proceedings of the 29th ACM International Conference on Multimedia (ACMMM)*, 2021, pp. 1581–1590.[2]

- **M.-J. Chiou**, C.-Y. Liao, L.-W. Wang, R. Zimmermann, and J. Feng, "ST-HOI: A Spatial-Temporal Baseline for Human-Object Interaction Detection in Videos", in *Proceedings of the 2021 Workshop on Intelligent Cross-Data Analysis and Retrieval (ICDAR)*, 2021, pp. 9–17.[3]

---

[1] https://github.com/coldmanck/RVL-BERT
[2] https://github.com/coldmanck/recovering-unbiased-scene-graphs
[3] https://github.com/coldmanck/VidHOI

The other works done during the course of this Ph.D. have also been published in the following peer-reviewed conference proceedings (and the first one is open-sourced):

- **M.-J. Chiou**, Z. Liu, Y. Yin, A.-A. Liu, and R. Zimmermann, "Zero-Shot Multi-View Indoor Localization via Graph Location Networks", in *Proceedings of the 28th ACM International Conference on Multimedia (ACMMM)*, 2020, pp. 3431–3440.[4]

- Y. Yin, **M.-J. Chiou**, Z. Liu, H. Shrivastava, R. R. Shah, and R. Zimmermann, "Multi-Level Fusion based Class-aware Attention Model for Weakly Labeled Audio Tagging", in *Proceedings of the 27th ACM International Conference on Multimedia (ACMMM)*, 2019, pp. 1304–1312.

---

[4] https://github.com/coldmanck/zero-shot-indoor-localization-release

# Chapter 2

# Background and Related Work

Visual scenes are naturally structured with salient objects in human visual system [35, 129]. This fact motivates straightforwardly recent efforts in learning to detect pairwise visual relationships among salient objects [128, 109] (Section 2.1). To represent visual scenes in a structured manner with the detected visual relations, researchers arrange the visual relations of an image into a scene graph [71] in which nodes are objects and edges are predicates (Section 2.2). On the other hand, in video domain, visual relationships with humans as the subject (*i.e.*, human-object interactions) are usually of central interest to a variety of real-word tasks, drawing increasing attention to video human object interaction detection (Section 2.3).

## 2.1 Visual Relationship Detection

Visual relationships in the form of (`subject`, `predicate`, `object`) describe the interactions between salient objects in images [109]. Formally, given a set of object classes $\mathcal{C}$ and predicate classes $\mathcal{R}$, a visual relationship instance is a triplet $(c_s, r, c_o)$ along with a tuple $(b_s, b_o)$, where $c_s, c_o \in \mathcal{C}, r \in \mathcal{R}$, $b_s, b_o$ correspond to the 2-d bounding box coordinates of the subject and the object, respectively. Given a set of objects $O = \{o_1, ..., o_n\}$ in an image where each object $o_i$ is associated with a class $c_i \in \mathcal{C}$, and $E \subseteq O \times \mathcal{R} \times O$ are all the pairwise visual relationships, *Visual Relationship Detection* (VRD) [128, 170, 109, 88, 202, 29, 193, 118, 187, 194, 183, 65, 72, 61, 45, 192, 13, 32, 112] aims to retrieve all of the visual relations $E$ in input images. Figure 2.1 (left) shows an illustration of VRD. Understanding visual relationships between objects is proved beneficial to vision tasks such as action detection [173, 161] and image retrieval [71, 130].

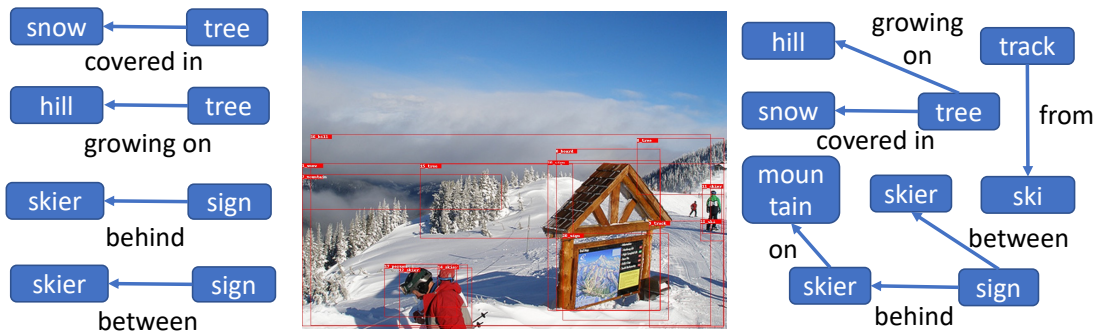

Figure 2.1: An illustrative comparison of VRD (left) and SGG (right). (Left) VRD models aim to predict visual relationships between (independent) object pairs. For instance, while the `tree` involving in the first and second object pair are essentially the same object, the two pairs are independent examples and are not jointly learned during training. (Right) In contrast, SGG methods form a structured representation comprising the salient objects, where the contexts are taken into account. This fact is reflected by that the two pairs involving `tree` are now combined into a local graph so that the contexts (*e.g.*, `hill` and `snow`) can be exploited simultaneously.

Three experimental settings are defined in the pioneering VRD paper [109]:

- *Predicate Detection*: given an input image and a set of localized objects (with object categories identified), machines are tasked to predict a ranked set of predicates between all possible object pairs.

- *Phrase Detection*: given an image, machines have to predict visual relationship triplets (`subject,predicate,object`), each with a single bounding box covering both `subject` and `object`.

- *Relationship Detection*: given an image, the task is to output a set of (`subject,predicate,object`) triplets, each associated with two bounding boxes of the `subject` and `object`, respectively.

Among the earliest works, Sadeghi et al. propose to formulate visual relationship to different combinations of object categories and relationship predicates, treating detection task as a classification problem [128]. Specifically, they represent visual relationships as *visual phrases* in the form of (subject, predicate, object). They then learn appearance vectors for each category and train a multi-class detection system to classify the visual phrases.

While the above approach proposed by Sadeghi et al. [128] works in a restricted size of dataset with the limited number of object categories and predicates, it is not capable of scaling to a large-scale dataset. If the given dataset consists of $N$ objects and $K$ predicates, then for the total $N^2K$ different combinations it needs $O(N^2K)$ detectors. Lu et al. propose to use $O(N + K)$ detectors to detect objects and predicates individually and fuse them into a higher-level representation for classification [109]. In their pipeline, a R-CNN detector [40] is first used to detect object proposals in an given image. For each object pairs, a likelihood score is then generated using a visual appearance module (utilizing purely visual feature) and a language module (utilizing language prior). These scores are then used to select and output a set of visual relationships.

Following Lu et al. [109], Liang et al. formulate visual relationship and attribute detection problem as a sequential decision-making process [96]. First of all, they define a directed semantic action graph where each node denotes an object, a relationship or an attribute. They then propose a *Deep Variation-structured Reinforcement Learning* (VRL) framework to detects relationships and attributes, introducing a variation-structured traversal scheme in which the agent select the optimal (attribute, predicate and object) actions in each step. Dai et al. propose *Deep Relational Network* (DR-Net) to infer the subtle relationships between object categories and relationship predicates [29]. Specifically, they detect candidate salient region proposals using pretrained object detector, and then extract spatial features, appearance features and image features of the pairs of regions. Spatial features and appearance features are then combined using a fusion layer and fed into DR-Net. Next, DR-Net is used to generate visual phrases (`subject,predicate,object`). As a special form of RNNs, it takes in a fixed set of inputs, *i.e.*, the fused input features and image features for subject and object, and output the predicted category probabilities for each element in the triplet.

More recently, *linguistic knowledge* has been incorporated as guidance signals for the VRD systems [109, 29, 45]. For instance, Lu et al. [109] proposed to detect objects and predicates individually with language priors and fuse them into a higher-level representation for classification. Dai et al. [29] also exploited statistical dependency between object categories and predicates to infer their subtle relationships. Going one step further, Gu et al. [45] proposed a dedicated module utilizing bi-directional

Gated Recurrent Unit to encode *external* language knowledge and a Dynamic Memory Network [82] to pick out the most relevant facts. However, none of the aforementioned works consider external *visual* knowledge, which arguably is also beneficial to visual relationship recognition. In Chapter 3, we propose to exploit both the visual and linguistic knowledge by pre-training bi-modal Transformers [156, 143] on large-scale multimodal datasets including 3.3 million image-alt text from Conceptual Captions dataset [135], Books Corpus [201] and the *Wikipedia*. The prior knowledge in the pre-trained model is shown to benefit the VRD performance in this thesis (Chapter 3).

## 2.2 Scene Graph Generation

*Scene graphs* [71] are graph-structured representations of images consisting of objects as nodes and predicates as directed edges, where each object pair along with their directed edges are visual relationships in the form of (`subject`,`predicate`,`object`). Formally, given an image, a scene graph $G = (O, E)$ comprises all the salient objects $O$ as nodes and pairwise visual relations $E$ as directed edges. *Scene Graph Generation* (SGG) [90, 191, 179, 89, 57, 15, 195, 119, 188, 189, 148, 102, 164, 177, 122, 167, 152, 74] is similar to VRD in that it also aims to generate pairwise visual relationships $E$ among salient objects $O$ in images. However, SGG models put an emphasis on representing a scene with a graph structure $G$ (on which neighboring objects and relationships are jointly learned) instead of a set of disconnected visual relation triplets. With scene graphs, we aim to generate more holistic, structured representations of visual scenes. Figure 2.1 shows an illustrative comparison of VRD and SGG.

There are also three experimental settings defined for evaluating SGG models [174]: *Predicate Classification* (PredCls), *Scene Graph Classification* (SGCls) and *Scene Graph Detection* (SGDet). PredCls is essentially the Predicate Detection task of VRD where it asks for predicate prediction given ground truth object bounding boxes and labels. SGCls tasks machines to classify the objects and predict a set of ranked visual relation triplets given localized objects (without class labels). SGDet is similar to the Relationship Detection task of VRD in that only the raw images are given; however, it requires a scene graph output instead of a set of visual relation triplets.

Among one of the earliest pioneers in SGG, Johnson et al. [71] were aware of the limitation of object name-based image retrieval (`person, horse`), and they proposed to utilize scene graphs, which consists of visual relationships (`person, riding, horse`) and attributes (`horse is black`), to search for images. Specifically, they proposed to match query scene graphs with unannotated test images with a Conditional Random Field (CRF) to model the distribution over all possible groundings, on which maximum a posteriori (MAP) is performed to obtain the most plausible grounding.

Xu et al. [174] are among the first to note that, by doing *local* learning, VRD models ignore surrounding contexts which could be important for making predictions. For example, by knowing *keyboards*, *mouses*, and *desk*, a SGG model can predict a big screen as *monitor* instead of *television*. They thus propose to do *joint* SGG learning with an *iterative message passing* scheme which represents node and edges as hidden states of two GRUs [25], node GRU and edge GRU, respectively. This is followed by an update mechanism, called *message pooling*, which comprises a node and an edge *message pooling* module: for each node and edge, the module take in the two neighboring edges' and nodes' hidden states to update its hidden states, respectively. They have shown through experiments that the proposed feature refinement scheme boost the performance of node (object) and edge (visual relation) predictions. Concurrently, while Li et al. [90] also propose to jointly learn SGG, they additionally draw inspiration from the correlation between the tasks between SGG, object detection and sub-region image captioning, and they propose *Multi-level Scene Description Network* (MSDN) to train visual understanding models in a multi-task manner. By conducting feature refining process on three levels, *i.e.*, object proposal, phrase proposal and caption proposal, and propagating messaging between them, they observed substantial performance gain on all of the three tasks.

More recent efforts [179, 191, 15, 149, 179, 45] follow [174, 90] to define a SGG system with three main modules: proposal generation module, object classification module, and relationship prediction module.

- **Proposal generation.** A pre-trained object detection model (*e.g.*, Faster R-CNN [123]) is usually adopted for generating proposals.

- **Contextualized object classification.** *Contexts* are arguably the most important ingredients in SGG. To be specific, by looking at other objects/predicates in the same image it would be easier to make predictions. For instance, `kitchen table` and `stove` are indicative of more informative `chopping board` instead of a more general `wooden board`. Instead of using the predictions of object detection models directly (like in VRD), the generated proposals along with the logits are refined into *object contexts* [174, 179, 191, 15, 149] followed by decoding into object labels. These object contexts are then refined via message passing algorithms (*e.g.*, Tree LSTM [59, 146], Gated Graph Neural Networks [93]) on a fully-connected [174, 179, 15], chained [191] or tree-structured [149] graph. Classifiers are then trained on the refined graph hidden states.

- **Contextualized relationship prediction.** Similar to object contexts, contexts are taken into account for relationship prediction. By knowing some very likely relationship triplets (`human, ride, horse`) nearby, machines could predict more confidently `ride` for other visually blurred or partially occluded (`human, ?, horse`) pairs. In general, the object contexts along with bounding box features are taken in by graph models to compute *relation contexts*, which are then utilized for training a visual relationship classifier [191, 148, 15].

Not until the recent works [149, 15] proposed the less biased *mean recall* metrics did the research communities in SGG pay attention to the class imbalance problem [45, 34, 102, 148, 177, 164]. Many recent approaches can be viewed as model-agnostic debiasing methods [148, 177, 164]. Tang et al. [148] borrow the counterfactual idea from causal inference to remove the context-specific bias. Yan et al. [177] propose to perform re-weighting with class relatedness-aware weights. Wang et al. [164] transfer the less-biased knowledge from the secondary learner to the main one with knowledge distillation. However, we note that there is a critical difference between the long tail of SGG datasets and that of other datasets. The common understanding to long-tailed datasets is that, naturally in "real-world" applications, a larger number of classes are the *rare* classes accounting for a small portion of examples [106, 27, 197]. However, we argue that SGG datasets like Visual Genome (VG) dataset [80] might not be good indicators to the real-world visual relationship distribution, because these datasets miss an unneglectable amount of visual relationship annotations due to the prohibitive, quadratic (or even cubic) labeling cost [148]. By not exhaustively labeling visual relation instances, a labeled dataset might misses some predicate classes more than others, caused by labeling biases including

- *Reporting bias* [115]: people tend to under-report all the information and report only those that are more obvious, *e.g.*, it is easier to label (`car`, **`on`**, `road`) than (`car`, **`parked on`**, `road`); and

- *Bounded rationality* [139]: the rationality is limited when people making decision, *e.g.*, it is more apparent for people to be sure of (`person`, **`on`**, `bed`), rather than (`person`, **`sleep in`**, `bed`).

In the both cases, annotators need to take into account the details to label the visual relations extensively; however, the fact that the SGG datasets were not (able to) collected in this way causes the long tail. In this thesis, we dive deep into this issue and propose a model-agnostic approach to remove the labeling biases when training a SGG system.

## 2.3 Video Human-Object Interaction Detection

In contrast to static-images, videos are usually utilized to record human-centered activities such as actions or interactions with objects or other humans. Real-word applications of human-centered visual understanding include pedestrian detection [33, 196, 105] and unmanned store systems [46, 104, 107]. Since humans are the focus, understanding their interactions with other objects can be helpful. While there are a variety of related works in human object interaction detection (HOID) and video understanding, we note that they are not well-suited to video human object interaction detection (VidHOID). First, HOID [141, 11] can be viewed as a special case of VRD, where the category of subjects is always `human`; however, HOID approaches only take static-images as input and ignore the sequential structure of video frames. Second, video visual relationship detection (VidVRD) [133, 121, 154, 131, 103, 14] extend VRD to video domain by detecting visual relation triplets with corresponding bounding box trajectories; nonetheless, VidVRD does not put emphasis on HOIs and thus the existing approaches ignore human-specific features such as human pose (keypoints) and body mesh. Third, video human-object interaction retrieval (VidHOIR) [120, 145, 28] deal with the *sub-activity detection* task (as defined in the CAD-120 dataset [79]) which aims to predict a single human sub-activity for each video segment. This means the actors are assumed to only interact with one object every time; however, this is overly simplified, *e.g.*, a `person` can for sure `hold` a `phone` while also `talk to` another `person`. Moreover, the evaluation metrics of VidHOIR models exclude humans' and objects' spatial localization, showing that VidHOIR is not suitable for structured representations of visual scenes. Different from these three tasks, we propose *video human-object interaction detection* (VidHOID) aiming to perform spatial-temporal localized, multi-human-object interaction detection exploiting human-centered features, which, to the best of our knowledge, remains unexplored. Please refer to Figure 2.2 for an illustrative comparison between these three tasks and our proposed VidHOID. We present a detailed literature review on the three above-mentioned related topics to shed light on the direction of the thesis.

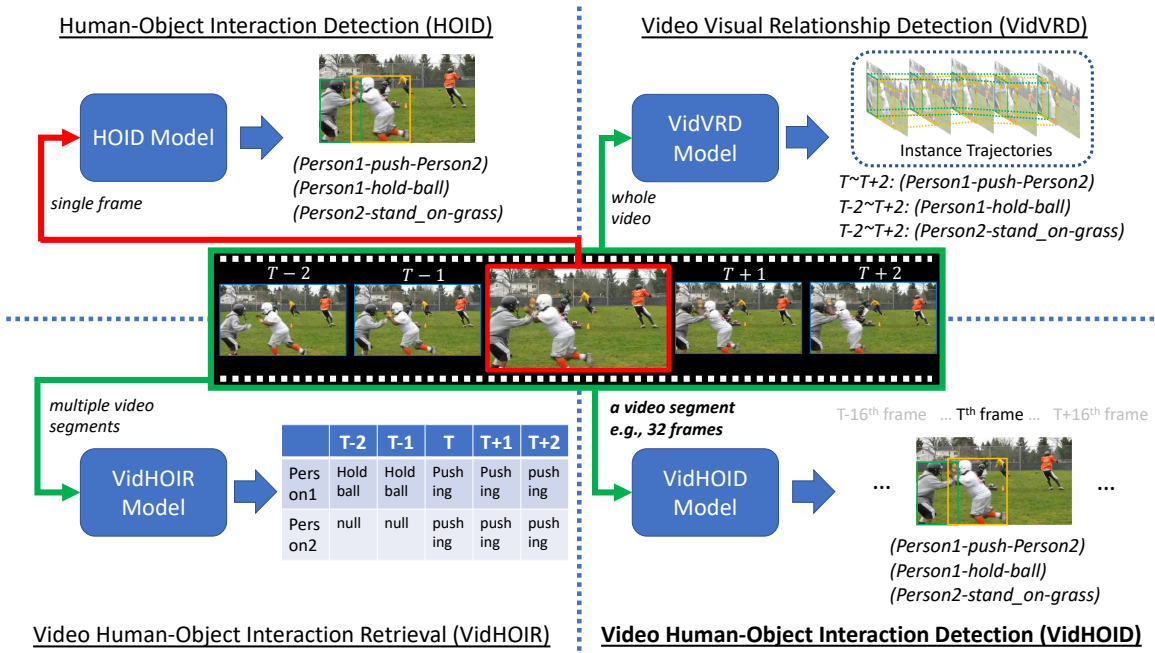

Figure 2.2: An illustrative comparison of human-object interaction (HOI) detection tasks including *human-object interaction detection* (HOID), *video visual relation detection* (VidVRD), *video human-object interaction retrieval* (VidHOIR) and *video human-object interaction detection* (VidHOID). HOID takes in single image (frame) and outputs HOIs with associated bounding boxes. VidVRD takes in a whole video and outputs visual relation instance along with object trajectories. VidHOIR takes in multiple video segments and outputs a sub-activity label for each of them. VidHOID takes in a single video segment (*e.g.*, 32 frames) centered at the *keyframe* and outputs HOIs associated with bounding boxes of the frame.

### 2.3.1 Human-Object Interaction Detection (HOID)

Human-object interaction (HOI) can be seen as a special case of visual relationship: while both are in the tuple form of (`subject`, `predicate`, `object`), the former must have a `human`/`person` as the subject category and a verb as the predicate class. Formally, given a set of object classes $\mathcal{C}$ and verb classes $\mathcal{R}$, a HOI label is a triplet $(h, r, c_o)$ where $c_o \in \mathcal{C}, r \in \mathcal{R}$, and $h, c_o$ are associated with bounding boxes of the human ($b_h$) and the interacting object $b_o$, respectively. Similar to VRD, HOID [48, 11, 39, 120, 92, 162, 159, 49, 91, 163, 97, 155, 199] aims to detect all pairwise HOIs in images. Gupta et al. [48] is among the first to work on HOID when they introduce a task *visual semantic role labeling* (which is essentially HOID) and propose one of the first HOID dataset, V-COCO. They also propose a *instance-based* baseline method which uses pre-trained detector to detect humans and predict the most probable action associated, followed by regressing bounding boxes of the *roles* (objects) involved in the actions. This effort is followed by Chao et al. [11] who propose another HOI dataset, HICO-DET, along with a multi-stream model including a RoI stream taking in the bounding boxes and image features and a pairwise interaction stream encoding the spatial features between human-object pairs.

Following the instance-based scheme, Gkioxari et al. [41] propose InteractNet which extends the Fast R-CNN [40] with an additional human-centric branch predicting possible actions and target object location based on an actor's appearance (*e.g.*, clothing, poses). Specifically, for each action candidate, the model predicts a *density* over possible locations using a Gaussian function whose mean is a 4-d box offset, similar to the bounding box encoding used in Fast R-CNN. Gao et al. [39] further extend this idea to target objects as well: the appearance of an instance provides cues for the part of the image that models should attend to. For instance, whether a person is carrying or riding a bicycle: 1) given the bicycle, machines attend to the pose of the person and 2) given the person, machines attend to the person's hands. They thus propose a three-stream *instance-centric attention network* (iCAN) comprising a human stream (for detecting interactions based on human appearance), a object stream (for detecting interactions based on object appearance) and a pairwise stream (for encoding spatial information between the human and object). On the other hand, Qi et al. [120] introduce *Graph Parsing Neural Networks* (GPNN) which learn an HOI graph in an iterative message passing manner (similar to [174, 90] in VRD). Specifically, the graph structure is defined by a soft adjacency matrix computed by the *link function*, on which node and edge features are refined with *message functions* (for receive incoming messages) and *update functions* (for update hidden states). Finally, GPNN outputs a parse graph with the *readout functions*.

Although there has been enormous progress in HOID, we note that most of the existing HOID models are designed for performing on static-images and are thus not suitable for VidHOID. While one can naively adapt HOID models for videos by operating on each video frame followed by feature fusion, this way ignores the sequential nature of the video frames. In this thesis, we formulate VidHOID as a HOID task *in video domain* and urge for the use of temporal information.

### 2.3.2 Video Visual Relation Detection (VidVRD)

First proposed by Shang et al. [133], VidVRD [133, 121, 154, 131, 103, 14] aims to bridge the gap between image-based VRD methods and VRD in videos by predicting visual relation triplets along with the trajectories of the subject and object. Formally, given a set of object classes $\mathcal{C}$ and predicate classes $\mathcal{R}$, VidVRD aims to detect all video visual relationship instances in a video, where each instance is a triplet $(c_s, r, c_o) \in \mathcal{C} \times \mathcal{R} \times \mathcal{C}$ where $c_s, c_o \in \mathcal{C}, r \in \mathcal{R}$, along with a tuple $(\mathcal{T}_s, \mathcal{T}_o)$ where $\mathcal{T}_s, \mathcal{T}_o$ are the trajectories of the subject and the object (continuous 2-d bounding boxes across video frames). A video visual relation instance is considered correctly retrieved if i) the relation triplet $(c_s, r, c_o)$ is the same as ground truth and ii) both the trajectories of the subject and the object overlap with the ground truth ones with *volumetric Intersection over Union* (vIoU) higher than a certain threshold (0.5 by default).

Shang et al. introduced the first VidVRD dataset, VidVRD-ImageNet [133],

while the same authors later proposed another larger-scale and more extensively annotated dataset, VidOR [131]. In their first work [133], they decompose VidVRD into a three-stage pipeline: object trajectory proposal, relation modeling and relation instance association. Specifically, a video is divided into a set of overlapping short clips on which object trajectories are generated. Similar to the decomposition of triplet class into separate object and predicate classes done in VRD [109], they propose to learn and predict each element separately. For each pair of object trajectory proposal, i) trajectory features $\mathcal{T}_s$ and $\mathcal{T}_o$ extracted with iDT features [160], and ii) a relativity feature vector including relative position, size and motion, are concatenated and used for joint training an "unified" cross-entropy loss. Finally, to associate visual relation instances, they merge the most probable candidates with the identical relation triplet label (`subject,predicate,object`) with sufficient trajectory overlapping. Following [133], Sun et al. [144] also adopt the similar three-step model for relation instance recognition, while they show that the combination of i) relative location feature [193], ii) motion feature, and iii) language embedding (word2vec [113] from GoogleNews) works very well. Sun et al. exploit state-of-the-art video object detection models including FGFA [200], Seq-NMS [50] and KCF tracker [56] for trajectory proposal generation.

We note that VidVRD (and the works on it) might be naively adapted to define VidHOID by restricting the subjects in relation triplets as `human`. Nonetheless, most of the existing methods in VidVRD do not take into account the human-centered features and/or utilize a human-centric features, *e.g.*, pose features [159, 49, 92], action density [41] or attention [39]) which are common in HOID models. Moreover, the *relation detection* task of VidVRD requires to predict trajectories accurately along with visual relation triplets; however, video object detection (*a.k.a.*, multiple object tracking) [98] might be overly and unnecessarily hard for both annotation and detection. First, video object detection is still a challenging task as objects might have large movements and noncontinuous trajectories. Second, as shown in THUMOS challenge [44, 63], the action boundaries are inherently fuzzy, resulting in significant disagreement between annotators. Third, as discussed in [4, 44], the hierarchical nature of human actions – a human can be not only stepping down the curb but also crossing the street (and walking to school) – makes defining a vocabulary of action categories on video-level ill-posed. To circumvent these problems, the authors of AVA dataset [44] omit video object detection and define the actions at the finest level (where actions can be determined with 1-second clips). Concretely, they adopt a keyframe-centered annotation and detection strategy, *i.e.*, detecting HOIs for a short video segment (*e.g.*, 1 second) centered at a keyframe (sampled uniformly in, *e.g.*, 1 Hz). On the other hand, the *relation tagging* task of VidVRD is too loose by only predicting a set of visual relation triplets without locating the objects. An alternative is to reduce the 3D localization to 2D by requiring only detecting bounding boxes in the keyframe. In this thesis, we follow the above-mentioned keyframe-centered strategy to define VidHOID and establish a new benchmark with keyframe-based evaluation metrics.

### 2.3.3 Video Human-Object Interaction Retrieval (VidHOIR)

VidHOIR, introduced by Koppula et al. [79] along with their CAD-120 dataset, is defined as predicting a sequence of sub-activities (one for each segment) involving 10 actions (*e.g.*, reaching, moving, placing, opening, etc.) for an actor in a video. The authors proposed a Markov random field-based model which treats objects and sub-activities as nodes and relations (including object-object, object-sub-activity and temporal interactions) as edges. They put together hand-crafted features, including human poses, joint positions, SIFT [108] features and 3D centroids, to represent each of the video segment. Qi et al. [120] extend their GPNN to solve VidHOIR by adapting the link function (with ConvLSTM [172]) to taking in the soft adjacency matrix at the previous time step. Dabral et al. [28], on the other hand, propose to decouple spatial feature extraction from temporal reasoning so that spatial models, such as Capsule Networks [127] and Graph Convolutional Networks [76], can be utilized.

We note that there are a couple of facts that make the VidHOIR definition unsuitable for detecting HOIs in videos. First, in CAD-120, for each video (sub-)segment and at each time step, the actor only perform *one* sub-activity at a time. This does not reflect the fact that a person can, of course, interact with multiple objects simultaneously. For instance, as we mentioned earlier, a `person` can `hold` a `phone` while also `talk to` another `person`. This is partly because the videos in CAD-120 were generated in a strictly-controlled laboratory condition; however, this also means the videos are far from real-world scenarios, and the methods developed with this definition are likely to have a hard time adapting for real-word applications. Furthermore, the current evaluation metrics of VidHOIR do not take into account the spatial-temporal localization of the instances (*i.e.*, humans and objects). This means VidHOIR can not be used for constructing "structured" representations of visual scenes. In this thesis, we instead propose that a VidHOID model should not only be able to predict multiple HOIs among multiple persons and objects, but also localize trajectories of the instances.

# Chapter 3

# Visual Relationship Detection with External Multimodal Knowledge

As discussed in Section 2.1, VRD aims to detect objects and classify triplets of `subject-predicate-object` in a query image. Most of the existing works [128, 170, 109, 88, 202, 29, 193, 187, 183, 65, 72, 61, 45, 192, 32, 112] train VRD models only on the target datasets (*e.g.*, the VRD dataset [109] or the VG dataset [80]). While other computer vision datasets (*e.g.*, captioning datasets [135, 101], language database [201]) are not collected for the purpose of training VRD models, they might still be useful for VRD because of their visual and linguistic knowledge. For instance, VRD models have been shown beneficial to training image captioning models [87, 12], and jointly learning VRD and region captioning boost the performance of both [90]. In this chapter, we study how we can exploit external visual and linguistic knowledge for more performant VRD models.

## 3.1 Introduction

To enhance the performance of VRD systems, some recent works incorporate the external *linguistic* knowledge from pre-trained word vectors [109], structured knowledge bases [45], raw language corpora [187], etc., as priors, which has taken inspiration from human reasoning mechanism. For instance, for a relationship triplet case `person-ride-bike` as shown in Figure 3.1, with linguistic knowledge, the predicate `ride` is more accurate for describing the relationship of `person` and `bike` compared with other relational descriptions like `on` or `above`, which are rather abstract. In addition, we argue that the external *visual* knowledge is also beneficial to lifting detection performance of the VRD models, which is however rarely considered previously. Take the same `person-ride-bike` in Figure 3.1 as an example. If the pixels inside the bounding box of `person` are masked (zeroed) out, humans can still predict them as a person since we have seen many examples and have plenty of visual knowledge regarding such cases. This reasoning process would be helpful for VRD systems since it incorporates relationships of the basic visual elements; however, most previous approaches learn visual knowledge only from target datasets and neglect external visual knowledge in abundant unlabeled data. Inspired by the recent successful visual-linguistic pre-training methods (BERT-like models) [85, 110], we propose to exploit both *linguistic* and *visual* knowledge from Conceptual Captions [135] — a large-scale dataset containing 3.3M images with coarsely-annotated descriptions (alt-text) that were crawled from the web, to achieve boosted VRD performance. We first pre-train our backbone model (multimodal BERT) on Conceptual Captions with different pretext tasks to learn the visual and linguistic knowledge. Specifically, our model mines visual prior information via learning to predict labels for an image's subregions that are randomly masked out.

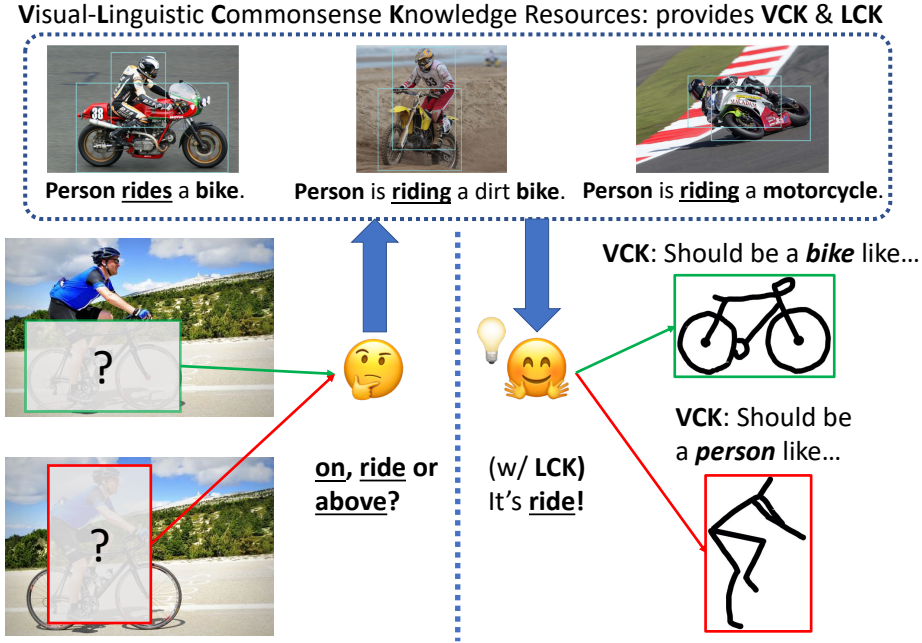

Figure 3.1: Illustration of human reasoning over visual relationships with external visual and linguistic knowledge. With these knowledge, a human is able to "guess" the visually blurred regions and prefer `ride` rather than `on` or `above`. **VCK**: Visual Knowledge. **LCK**: Linguistic Knowledge.

The model also considers linguistic knowledge through learning to predict randomly masked out words of sentences in image captions. The pre-trained weights are then used to initialize the backbone model and trained together with other additional modules (detailed at below) on visual relationship datasets.

Besides visual and linguistic knowledge, spatial features are also important cues for reasoning over object relationships in images. For instance, for `A-on-B`, the bounding box (or it's center point) of `A` is often above that of `B`. However, such spatial information is not explicitly considered in BERT-like visual-linguistic models [143, 110, 85]. We thus design two additional modules to help our model better utilize such information: a mask attention module and a spatial Module. The former predicts soft attention maps of target objects, which are then used to enhance visual features by focusing on target regions while suppressing unrelated areas; the latter augments the final features with bounding boxes coordinates to explicitly take spatial information into account.

We integrate the aforementioned designs into a novel VRD model, named ***R**elational **V**isual-**L**inguistic **B**idirectional **E**ncoder **R**epresentations from **T**ransformers* (RVL-BERT). RVL-BERT makes use of the pre-trained visual-linguistic representations as the source of visual and language knowledge to facilitate the learning and reasoning process on the downstream VRD task. It also incorporates a novel mask attention module to actively focus on the object locations in the input images and a spatial module to capture spatial relationships more accurately.

In this chapter, our contribution is three-fold. Firstly, we are among the first to identify the benefit of visual-linguistic knowledge to visual relationship detection, especially when objects are occluded. Secondly, we propose RVL-BERT – a multi-modal VRD model pre-trained on visaul-linguistic knowledge bases learns to predict visual relationships with the attentions among visual and linguistic elements, with the aid of the spatial and mask attention module. Finally, we show through extensive experiments that the knowledge and the proposed modules effectively improve the model performance, and our RVL-BERT achieves competitive results on two VRD datasets. This chapter including all of the texts, figures, tables, illustrations, equations is based on our published paper [24].

## 3.2 Preliminaries

### 3.2.1 Representation Pre-training

In the past few years, self-supervised learning which utilizes its own unlabeled data for supervision has been widely applied in representation pre-training. BERT, ELMo [117] and GPT-3 [8] are representative language models that perform self-supervised pre-training on various pretext tasks with either Transformer blocks or bidirectional LSTM. More recently, increasing attention has been drawn to multimodal (especially visual and linguistic) pre-training. Based on BERT, Visual-Linguistic BERT (VL-BERT) [143] pre-trains a single stream of cross-modality transformer layers from not only image captioning datasets but also language corpora. It is trained on BooksCorpus [201] and English Wikipedia in addition to Conceptual Captions [135]. We refer interested readers to [143] for more details of VL-BERT.

### 3.2.2 BERT and VL-BERT

In this work, we utilize both visual and linguistic knowledge learned in the pretext tasks. While VL-BERT can be applied to training VRD without much modification, we show experimentally that their model does not perform well due to lack of attention to spatial features. By contrast, we propose to enable knowledge transfer for boosting detection accuracy and use two novel modules to explicitly exploit spatial features.

Let a sequence of $N$ embeddings $x = \{x_1, x_2, ..., x_N\}$ be the features of input sentence words, which are the summation of *token*, *segment* and *position embedding* as defined in BERT [31]. The BERT model takes in $x$ and utilizes a sequence of $n$ multi-layer bidirectional Transformers [156] to learn contextual relations between words. Let the input feature at layer $l$ denoted as $x^l = \{x_1^l, x_2^l, ..., x_N^l\}$. The feature of x at layer $(l+1)$, denoted as $x^{l+1}$, is computed through a Transformer layer which consists of two sub-layers: 1) a multi-head self-attention layer plus a residual

connection

$$\tilde{h}_i^{l+1} = \sum_{m=1}^{M} W_m^{l+1} \left\{ \sum_{j=1}^{N} A_{i,j}^m \cdot V_m^{l+1} x_j^l \right\}, \tag{3.1}$$

$$h_i^{l+1} = \text{LayerNorm}(x_i^l + \tilde{h}_i^{l+1}), \tag{3.2}$$

where $A_{i,j}^m \propto (Q_m^{l+1} x_i^l)^T (K_m^{l+1} x_j^l)$ represents a normalized dot product attention mechanism between the $i$-th and the $j$-th feature at the $m$-th head, and 2) a position-wise fully connected network plus a residual connection

$$\tilde{x}_i^{l+1} = W_2^{l+1} \cdot \text{GELU}((W_1^{l+1} h_i^{l+1}) + b_1^{l+1}) + b_2^{l+1}, \tag{3.3}$$

$$x_i^{l+1} = \text{LayerNorm}(h_i^{l+1} + \tilde{x}_i^{l+1}), \tag{3.4}$$

where GELU is an activation function named Gaussian Error Linear Unit [54]. Note that **Q** (Query), **K** (Key), **V** (Value) are learnable embeddings for the attention mechanism, and **W** and **b** are learnable weights and biases respectively.

Based on BERT, VL-BERT [143] adds $O$ more multi-layer Transformers to take in additional $k$ visual features. The input embedding becomes $x = \{x_1, ..., x_N, x_{N+1}, ..., x_{N+O}\}$, which is computed by the summation of not only the token, segment and position embeddings but also an additional *visual feature embedding* which is generated from the bounding box of each corresponding word. The model is then pre-trained on two types of pretext tasks to learn the visual-linguistic knowledge: 1) *masked language modeling with visual clues* that predicts a randomly masked word in a sentence with image features, and 2) *masked RoI classification with linguistic clues* that predicts the category of a randomly masked region of interest (RoI) with linguistic information.

## 3.3 Methodology

### 3.3.1 Model Overview

Figure 3.2 shows the overall architecture of our proposed RVL-BERT. For the backbone BERT model, we adopt a 12-layer Transformer and initialize it with the pre-trained weights of VL-BERT for visual and linguistic knowledge. Note that while our model is inspired by VL-BERT, it differs in several important aspects:

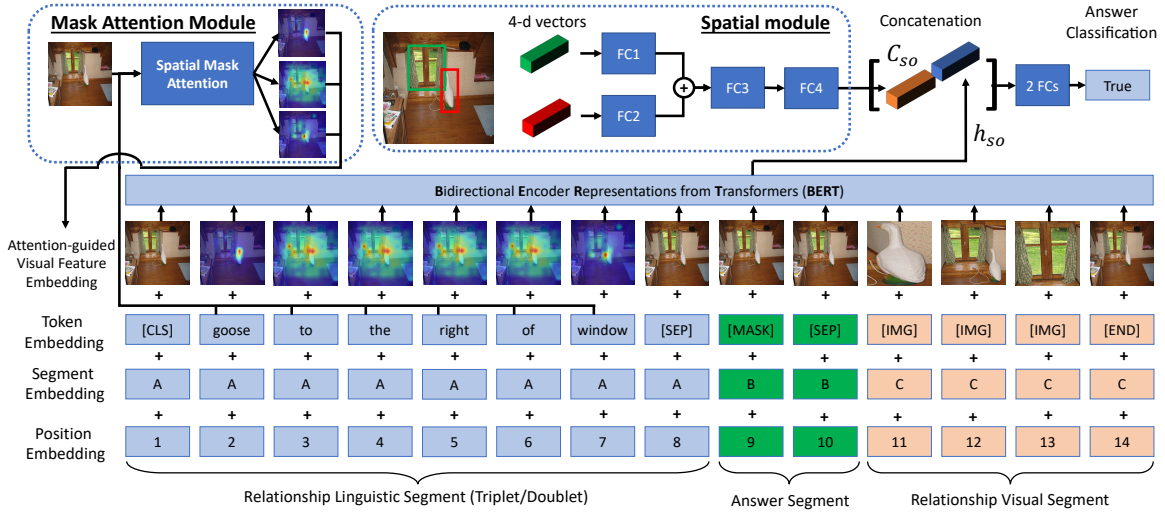

Figure 3.2: Architecture illustration of proposed RVL-BERT for SpatialSense dataset [180]. It can be easily adapted for VRD dataset [109] by replacing triplets `subject-predicate-object` with doublets `subject-object` and performing predicate classification instead of binary classification on the output feature of "[MASK]".

1) RVL-BERT explicitly arranges query object pairs in sequences of `subject-predicate-object` (instead of sentences in the original design) and receives an extra answer segment for relationship prediction. 2) Our model is equipped with a novel mask attention module that learns attention-guided visual feature embeddings for the model to attend to target object-related area. 3) A simple yet effective spatial module is added to capture spatial representation of subjects and objects, which are of importance in spatial relationship detection.

Let $N$, $A$ and $O$ denote the number of elements for the relationship linguistic segment, the answer segment, and the relationship visual segment, respectively. Our model consists of $N + A + O$ multi-layer Transformers, which takes in a sequence of linguistic and visual elements, including the output from the mask attention module, and learns the context of each element from all of its surrounding elements. For instance, as shown in Figure 3.2, to learn the representation of the linguistic element `goose`, the model looks at not only the other linguistic elements (*e.g.*, `to the right of` and `window`) but also all visual elements (*e.g.*, `goose`, `window`). Along with the multi-layer Transformers, the spatial module extracts the location information of subjects and objects using their bounding box coordinates. Finally, the output representation of the element in the answer segment, $h_{so}$, is augmented with the output of the spatial module $C_{so}$, followed by classification with a 2-layer fully connected network.

The input to the model can be divided into three groups by the type of segment, or four groups by the type of embedding. We explain our model below from the segment-view and the embedding-view, respectively.

#### 3.3.1.1 Input Segments

For each input example, RVL-BERT receives a relationship linguistic segment, an answer segment, and a relationship visual segment as input.

- **Relationship linguistic segment** (light blue elements in Figure 3.2) is the linguistic information in a triplet form `subject-predicate-object`, like the input form of SpatialSense dataset [180], or a doublet form `subject-object` like the input in VRD dataset [109]). Note that each term in the triplet or doublet may have more than one element, such as `to the right of`. This segment starts with a special element "[CLS]" that stands for classification[1] and ends with a "[SEP]" that keeps different segments separated.

- **Answer segment** (green elements in Figure 3.2) is designed for learning a representation of the whole input and has only special elements like "[MASK]" that is for visual relationship prediction and the same "[SEP]" as in the relationship linguistic segment.

- **Relationship visual segment** (tangerine color elements in Figure 3.2) is the visual information of a relationship instance, also taking the form of triplets or doublets but with each component term corresponding to only one element even if its number of words of the corresponding label is greater than one.

#### 3.3.1.2 Input Embeddings

There are four types of input embeddings: token embedding $t$, segment embedding $s$, position embedding $p$, and (attention-guided) visual feature embedding $v$. Among them, the attention-guided visual feature embedding is newly introduced while the others follow the original design of VL-BERT. We denote the input of RVL-BERT as $x = \{x_1, ..., x_N, x_{N+1}, ..., x_{N+A}, x_{N+A+1}..., x_{N+A+O}\}$, $\forall x_i : x_i = t_i + v_i + s_i + p_i$ where $t_i \in t$, $v_i \in v$, $s_i \in s$, $p_i \in p$.

---

[1] We follow the original VL-BERT to start a sentence with the "[CLS]" token, but we do not use it for classification purposes.

- **Token Embedding**. We transform each of the input words into a $d$-dimensional feature vector using WordPiece embeddings [169] comprising 30,000 distinct words. In this sense, our model is flexible since it can take in any object label with any combination of words available in WordPiece. Note that for those object/predicate names with more than one word, the exact same number of embeddings is used. For the $i$-th object/predicate name in an input image, we denote the token embedding as $t = \{t_1, ..., t_N, t_{N+1}, ..., t_{N+A}, t_{N+A+1}..., t_{N+A+O}\}$, $t_i \in \mathbb{R}^d$, where $d$ is the dimension of the embedding. We utilize WordPiece embeddings for relationship triplets/doublets $\{t_2, ..., t_{N-1}\}$, and use special predefined tokens "[CLS]", "[SEP]", "[MASK]" and "[IMG]" for the other elements.

- **Segment Embedding**. We use three types of learnable segment embeddings $s = \{s_1, ..., s_{N+A+O}\}$, $s_i \in \mathbb{R}^d$ to inform the model that there are three different segments: "$A$" for relationship linguistic segment, "$B$" for answer segment and "$C$" for relationship visual segment.

- **Position Embedding**. Similar to segment embeddings, learnable position embeddings $p = \{p_1, ..., p_{N+A+O}\}$, $p_i \in \mathbb{R}^d$ are used to indicate the order of elements in the input sequence. Compared to the original VL-BERT where the position embeddings of the relationship visual segment are the same for each RoI, we use distinct embeddings as our RoIs are distinct and ordered.

- **Visual Feature Embedding**. These embeddings are to inform the model of the internal visual knowledge of each input word. Given an input image and a set of RoIs, a CNN backbone is utilized to extract the feature map, which is prior to the output layer, followed by RoI Align [51] to produce fixed-size feature vectors $z = \{z_0, z_1, ..., z_K\}$, $z_i \in \mathbb{R}^d$ for $K$ RoIs, where $z_0$ denotes the feature of the whole image. For triplet inputs, we additionally generate $K(K-1)$ features for all possible union bounding boxes: $u = \{u_1, ..., u_{K(K-1)}\}$, $u_i \in \mathbb{R}^d$. We denote the input visual feature embedding as $v = \{v_1, ..., v_N, v_{N+1}, ..., v_{N+A}, v_{N+A+1}..., v_{N+A+O}\}$, $v_i \in \mathbb{R}^d$. We let subject and object be $s$ and $o$, with $s, o \in \{1, ..., K\}, s \neq o$, and let the union bounding box of $s$ and $o$ be $so \in \{1, ..., K(K-1)\}$.

For the relationship visual segment $\{v_{N+A}, ..., v_{N+A+O-1}\}$ (excluding the final special element), we use $z_s$ and $z_o$ as the features of subject $s$ and object $o$ in doublet inputs, and add another $u_{so}$ in between in case of triplet inputs. For the special elements other than "[IMG]", we follow VL-BERT to use the full image feature $z_0$. However, for the relationship linguistic segment $\{v_2, ..., v_{N-1}\}$ (excluding the first and final special elements), it is unreasonable to follow the original design to use the same, whole-image visual feature for all elements, since each object/predicate name in the relationship linguistic segment should correspond to different parts of the image. To better capture distinct visual information for elements in relationship linguistic segment, we propose a *mask attention module* to learn to generate attention-guided visual feature embeddings that attend to important (related) regions, which is detailed at below.

### 3.3.2 Mask Attention Module

An illustration of the mask attention module is shown in Figure 3.3. Denote the visual feature (the feature map before average pooling) used by the mask attention module as $v_s \in \mathbb{R}^{d_c \times d_w \times d_h}$, where $d_c, d_w, d_h$ stand for the dimension of the channel, width, and height, respectively. To generate the feature for an object $s$ (*e.g.*, `goose` in Figure 3.3), the mask attention module takes in and projects the visual feature $v_s$ and the word embedding [2] $w_s$ into the same dimension using a standard CNN and a replication process, respectively

$$\tilde{v}_s = \sigma(W_1^T v_s + b_1), \tag{3.5}$$

$$w_s = \text{Replication}(w_s), \tag{3.6}$$

where Replication($\cdot$) replicates the input vector of size $d$ into the feature map of dimension $d \times d_w \times d_h$. The above is followed by element-wise addition to fuse the features, two convolutional layers as well as a re-scaling process to generate the

---

[2]Note that for object labels with more than one word, the embeddings of each word are element-wise summed in advance.

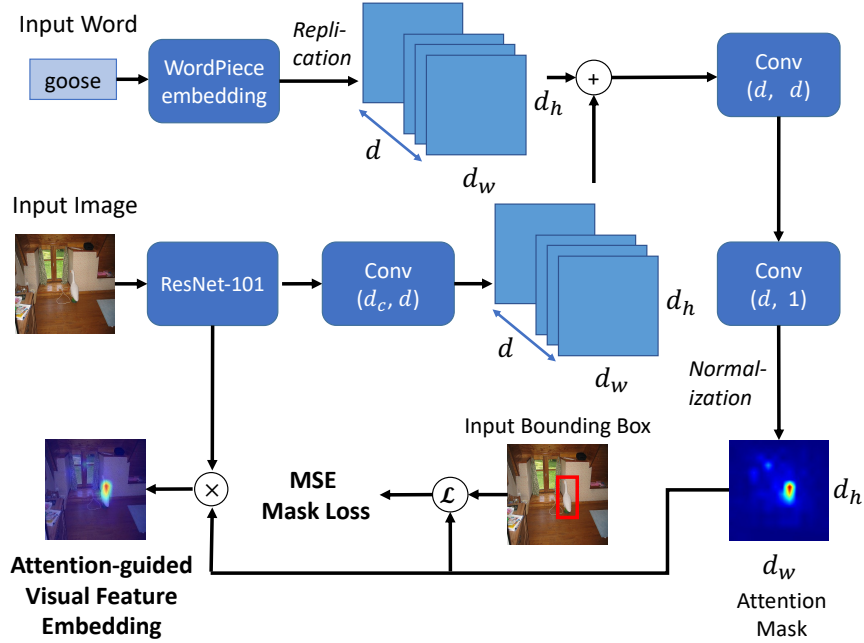

Figure 3.3: The pipeline of Mask Attention Module. Given an input image and the corresponding word embedding(s), the module generates an attention mask (heatmap) and outputs an attention-guided visual embedding.

attention mask $m_s$

$$\tilde{m}_s = \sigma(W_2^T(\tilde{v}_s + w_s) + b_2), \tag{3.7}$$

$$m_s = \text{Norm}(W_3^T \tilde{m}_s + m_3), \tag{3.8}$$

where the min-max $\text{Norm}(\cdot)$ applied to each element is defined by

$$\text{Norm}(x_i) = \frac{x_i - \min(x)}{\max(x) - \min(x)}. \tag{3.9}$$

Note that in the above equations all of the $W$'s and $b$'s are learnable weights and biases of the convolutional layers, respectively. The attention-guided visual feature $v_s^{att}$ is then obtained by performing Hadamard product between the visual feature and the attention mask: $v_s^{att} = v_s \circ m_s$. Finally, $v_s^{att}$ is pooled into $v_s^{att} \in \mathbb{R}^d$ to be used in $\{v_2, ..., v_{N-1}\}$.

To learn to predict the attention masks, we train the module against the Mean Squared Error (MSE loss) between the mask $m_s$ and the resized ground truth mask

$b_s$ consisting of all ones inside the bounding box and outside all zeros:

$$\mathcal{L}_{mask} = \frac{1}{d_w d_h} \sum_{i=1}^{d_w} \sum_{j=1}^{d_h} (m_s^{ij} - b_s^{ij})^2, \tag{3.10}$$

where $d_w$, $d_h$ denote the width and length of the attention mask.

### 3.3.3 Spatial Module

The spatial module aims to augment the output representation with spatial knowledge by paying attention to bounding box coordinates. See the top part of Figure 3.2 for its pipeline.

Let $(x_i^0, y_i^0)$, $(x_i^1, y_i^1)$ denote the top-left and bottom-right coordinates of a bounding box of an object $i$ of an input image, and let $w$, $h$ be the width and height of the image. The 4-dimensional normalized coordinate of an object $i$ is defined by $C_i = (x_i^0/w, y_i^0/h, x_i^1/w, y_i^1/h)$. The spatial module takes in coordinate vectors of a subject $s$ and an object $o$, and encodes them using linear layers followed by element-wise addition fusion and a two-layer, fully-connected layer

$$\tilde{C}_{so} = \sigma(W_4 C_s + b_4) + \sigma(W_5 C_o + b_5), \tag{3.11}$$

$$C_{so} = W_7 \, \sigma(W_6 \tilde{C}_{so} + b_6) + b_7. \tag{3.12}$$

The output feature $C_{so}$ is then concatenated with the multimodal feature $h_{so}$ to produce $f_{so}$ for answer classification:

$$f_{so} = [C_{so}; h_{so}]. \tag{3.13}$$

## 3.4 Experiments

### 3.4.1 Datasets

We first ablate our proposed model on VRD dataset [109], which is the most widely used benchmark. For comparison with previous methods, we also evaluate on SpatialSense [180] dataset. Compared with Visual Genome (VG) dataset [80], SpatialSense suffers less from the dataset language bias problem, which is considered a distractor for performance evaluation — in VG, the visual relationship can be "guessed" even without looking at the input image [191, 15]. Despite the strong dataset bias, we still provide the experimental results on VG for reference.

#### 3.4.1.1 VRD

The VRD dataset consists of 5,000 images with 37,993 visual relationships. We follow [109] to divide the dataset into a training set of 4,000 images and a test set of 1,000 images, while only 3,780 and 955 images are annotated with relations, respectively. For all possible pairs of objects in an image, our model predicts by choosing the best-scoring predicate and records the scores, which are then used to rank all predictions in the ascending order. Since the visual relationship annotations in this dataset are far from exhaustive, we cannot use precision or average precision as they will penalize correct detections without corresponding ground truth. Traditionally, Recall@K is adopt to bypass this problem and we follow this practice throughout our experiments. For VRD, the task named *Predicate Detection/Classification* measures the accuracy of predicate prediction given ground truth classes and bounding boxes of subjects and objects independent of the object detection accuracy. Following [109, 193], we use **Recall@K**, or the fraction of ground truth relations that are recalled in the top $K$ candidates. $K$ is usually set as 50 or 100 in the literature.

#### 3.4.1.2 SpatialSense

SpatialSense is a relatively new visual relationship dataset focusing on especially spatial relations. Different from Visual Genome [80], SpatialSense is dedicated to reducing dataset bias, via a novel data annotation approach called Adversarial Crowdsourcing which prompts annotators to choose relation instances that are hard to guess by only looking at object names and bounding box coordinates. SpatialSense defines nine spatial relationships `above`, `behind`, `in`, `in front of`, `next to`, `on`, `to the left of`, `to the right of`, and `under`, and contains 17,498 visual relationships in 11,569 images. The task on SpatialSense is binary classification on given visual relationship triplets of images, namely judging if a triplet `subject-predicate-object` holds for the input image. Since in SpatialSense the number of examples of "True" equals that of "False", the **classification accuracy** can be used as a fair measure. We follow the original split in [180] to divide them into 13,876 and 3,622 relations for training and test purposes, respectively.

#### 3.4.1.3 Visual Genome

Visual Genome (VG) is one of the most widely used datasets for visual relationship detection and scene graph generation. We follow [174] to use a subset of 89,169 images with 150 object categories and 50 predicate classes. We evaluate our model under the setting of Predicate Classification (PredCls) where the ground truth object labels and boxes are given and we use the same **Recall@K** as for VRD dataset.

### 3.4.2 Implementation

For the backbone model, we use BERT$_{\text{BASE}}$[3] that is pre-trained on three datasets including Conceptual Captions [135], BooksCorpus [201] and English Wikipedia. For extracting visual embedding features, we adopt Fast R-CNN [40] (detection branch of Faster R-CNN [123]). We randomly initialize the final two fully connected layers and the newly proposed modules (*i.e.*, mask attention module and spatial module). During training, we find our model empirically gives the best performance when freezing the parameters of the backbone model and training on the newly introduced modules. We thus get a lightweight model compared to the original VL-BERT as the number of trainable parameters is reduced by around 96%, *i.e.*, down from 161.5M to 6.9M and from 160.9M to 6.4M when trained on the SpatialSense dataset and the VRD dataset, respectively. ReLU is used as the nonlinear activation function $\sigma$. We use $d = 768$ for all types of input embeddings, $d_c = 2048$ for the dimension of channel of the input feature map and $d_w = d_h = 14$ for the attention mask in the mask attention module. The training loss is the sum of the softmax cross-entropy loss for answer classification and the MSE loss for the mask attention module. The experiments were conducted on a single NVIDIA Quadro RTX 6000 GPU in an end-to-end manner using Adam [75] optimizer with initial learning rate $1 \times 10^{-4}$ after linear warm-up over the first 500 steps, weight decay $1 \times 10^{-4}$ and exponential decay rate 0.9 and 0.999 for the first- and the second-moment estimates, respectively. We trained our model for 60 and 45 epochs for VRD and SpatialSense dataset, respectively, as there are more images in the training split of SpatialSense. For experiments on the VRD dataset, we followed the training practice in [61] to train with an additional "no relationship" predicate and for each image we sample 32

---

[3]There are two variants of BERT: BERT$_{\text{BASE}}$ that has 12-layer Transformers and BERT$_{\text{LARGE}}$ that has 24-layer Transformers.

Table 3.1: Ablation results for different losses of mask attention module and ways of feature combination. **.3**, **.5** and **.7** denote different $\alpha$ values in $f_{so} = \alpha C_{so} + (1-\alpha)h_{so}$.

| MAM Loss | | Feature Combination | | | | Recall@50 |
|---|---|---|---|---|---|---|
| BCE | MSE | .3 | .5 | .7 | concat | |
| ✓ | | | | | ✓ | 53.50 |
| | ✓ | | | | ✓ | **55.55** |
| | ✓ | ✓ | | | | 55.46 |
| | ✓ | | ✓ | | | 54.74 |
| | ✓ | | | ✓ | | 55.19 |

relationships with the ratio of ground truth relations to negative relations being $1:3$.

### 3.4.3 Ablation Study Results

#### 3.4.3.1 Training Objective for Mask Attention Module

We first compare performance difference between training the mask attention module (MAM) against **MSE** loss or binary cross entropy (**BCE**) loss. The first two rows of Table 3.1 show that MSE outperforms BCE by relative 3.8% on Recall@50. We also observe that training with BCE is relatively unstable as it is prone to gradient explosion under the same setting.

#### 3.4.3.2 Feature Combination

We also experiment with different ways of feature combination, namely, element-wise addition and concatenation of the features. To perform the experiments, we modify Eqn. 3.13 as $f_{so} = \alpha C_{so} + (1 - \alpha)h_{so}$, and we experiment with different $\alpha$ values (**.3**, **.5** and **.7**). The last five rows of Table 3.1 show that concatenation performs slightly better than addition under all $\alpha$ values.

The setting in the second row of Table 3.1 empirically gives the best performance, and thus we stick to this setting for the following experiments.

#### 3.4.3.3 Module Effectiveness

We ablate the training strategy and the modules in our model to study their effectiveness. **VL** indicates that the RVL-BERT utilizes the external multimodal knowledge learned in the pretext tasks via weight initialization. **Spatial** (**S**) means the spatial module, while **Mask Att.** (**M**) stands for the mask attention module. Table 3.2 shows that each module effectively helps boost the performance. The visual-linguistic knowledge lifts the **Basic** model by 12% (or absolute 5%) of Recall@50 on VRD dataset, while the spatial module further boosts the model by more than 23% (or absolute 10%). As the effect of the mask attention module is not apparent on the VRD dataset (0.2% improvement), we also experiment on the SpatialSense dataset (Overall Accuracy) and find the mask attention module provide a relative 1% boost of accuracy.

Table 3.2: Ablation results of each module on VRD dataset (Recall@50) and SpatialSense dataset (Overall Acc.) **VL**: Visual-Linguistic Knowledge. **S**: Spatial. **M**: Mask Att.

| Model | VL | Spatial | Mask Att. | R@50 | Acc. |
| --- | --- | --- | --- | --- | --- |
| Basic | | | | 40.22 | 55.4 |
| +VL | ✓ | | | 45.06 | 61.8 |
| +VL+S | ✓ | ✓ | | 55.45 | 71.6 |
| +VL+S+M | ✓ | ✓ | ✓ | **55.55** | **72.3** |

### 3.4.4 Quantitative Results

#### 3.4.4.1 VRD Dataset

We conduct experiments on VRD dataset to compare our method with the following, existing approaches:

- **Visual Phrase** [128] represents visual relationships as visual phrases and learns appearance vectors for each category for classification.

- **Joint CNN** [109] classifies the objects and predicates using only visual features from bounding boxes.

- **VTransE** [193] projects objects and predicates into a low-dimensional space and models visual relationships as a vector translation.

- **PPR-FCN** [194] uses fully convolutional layers to perform relationship detection.

Table 3.3: Performance comparison with existing models on VRD dataset. Results of the existing methods are extracted from [109] and respective papers.

| Model | Recall@50 | Recall@100 |
| --- | --- | --- |
| Visual Phrase [128] | 0.97 | 1.91 |
| Joint CNN [109] | 1.47 | 2.03 |
| VTransE [193] | 44.76 | 44.76 |
| PPR-FCN [194] | 47.43 | 47.43 |
| Language Priors [109] | 47.87 | 47.87 |
| Zoom-Net [183] | 50.69 | 50.69 |
| TFR [65] | 52.30 | 52.30 |
| Weakly (+ Language) [118] | 52.60 | 52.60 |
| LK Distillation [187] | 55.16 | 55.16 |
| Jung et al. [72] | 55.16 | 55.16 |
| UVTransE [61] | 55.46 | 55.46 |
| MF-URLN [192] | 58.20 | 58.20 |
| HGAT [112] | **59.54** | **59.54** |
| **RVL-BERT** | 55.55 | 55.55 |

- **Language Priors** [109] utilizes individual detectors for objects and predicates and combines the results for classification.

- **Zoom-Net** [183] introduces new RoI Pooling cells to perform message passing between local objects and global predicate features.

- **TFR** [65] performs a factorization process on the training data and derives relational priors to be used in VRD.

- **Weakly** [118] adopts a weakly-supervised clustering model to learn relations from image-level labels.

- **LK Distillation** [187] introduced external knowledge with a teacher-student knowledge distillation framework.

- **Jung et al.** [72] propose a new spatial vector with element-wise feature combination to improve the performance.

- **UVTransE** [61] extends the idea of vector translation in VTransE with the contextual information of the bounding boxes.

- **MF-URLN** [192] adopts a multi-modal feature model including visual features, spatial features and linguistic features and train the system with a "relationship-ness" loss to exploit unlabeled data.

- **HGAT** [112] proposes a graph-based approaches to jointly learn with both object and relation contexts.

Table 3.3 shows the performance comparison on the VRD dataset.[4] It can be seen that our **RVL-BERT** achieves competitive Recall@50/100 (53.07/55.55) compared to most of the existing methods, while lags behind the latest state-of-the-art, such as MF-URLN and HGAT.

For MF-URLN, we notice that there are two critical design choices significantly increasing its performance: i) the linguistic features extracted from external (Wikipedia) and the internal statistics embedding transformed by Naive Bayes with Laplace smoothing (R@50: 54.66→58.22; +3.56), and ii) the "determinate confidence subnetwork" which predicts if a candidate object pair is selected to labeled (R@50 under Relation Detection mode: 17.36→23.89; +6.53). While RVL-BERT has implicitly learned linguistic knowledge from external datasets, their external Word2Vec embedding and the internal statistics embedding might be compact and powerful additions to our model as well.

For HGAT, we find that they also have two important modules accounted for better performance: i) triplet-level reasoning which borrows the idea from SGG methods [191, 15, 149] (54.55→59.54; +4.99) and ii) the graph attention mechanism similar to Graph Attention Networks [157] (54.89→59.54; +4.65). While external linguistic knowledge ("semantic consistency"; similar to the idea of Chen et al. [15] is also utilized in HGAT, it contributes a relatively limited performance boost (58.42→59.54; +1.12). As our RVL-BERT is essentially an object pair-wise laerner, it can possibly be stacked upon the graph reasoning networks like the one in HGAT or other SGG efforts [191, 15, 149].

---

[4]Note that for the results other than **Visual Phrases** and **Joint CNN**, Recall@50 is equivalent to Recall@100 (also observed in [109, 187]) because the number of ground truth subject-object pairs is less than 50.

Table 3.4: Classification accuracy comparison on the test split of SpatialSense dataset. Bold font represents the highest accuracy; underline means the second highest. Results of existing methods are extracted from [180]. †Note that DSRR [32] is a concurrent work that was published in August 2020.

| Model | Overall | above | behind | in | in front of | next to | on | left of | right of | under |
|---|---|---|---|---|---|---|---|---|---|---|
| L-baseline [180] | 60.1 | 60.4 | 62.0 | 54.4 | 55.1 | 56.8 | 63.2 | 51.7 | 54.1 | 70.3 |
| PPR-FCN [194] | 66.3 | 61.5 | 65.2 | 70.4 | 64.2 | 53.4 | 72.0 | <u>69.1</u> | 71.9 | 59.3 |
| ViP-CNN [88] | 67.2 | 55.6 | 68.1 | 66.0 | 62.7 | 62.3 | 72.5 | **69.7** | 73.3 | 66.6 |
| Weakly [118] | 67.5 | 59.0 | 67.1 | 69.8 | 57.8 | **65.7** | 75.6 | 56.7 | 69.2 | 66.2 |
| S-baseline [180] | 68.8 | 58.0 | 66.9 | <u>70.7</u> | 63.1 | 62.0 | 76.0 | 66.3 | <u>74.7</u> | 67.9 |
| VTransE [193] | 69.4 | 61.5 | 69.7 | 67.8 | 64.9 | 57.7 | 76.2 | 64.6 | 68.5 | 76.9 |
| L+S-baseline [180] | 71.1 | 61.1 | 67.5 | 69.2 | 66.2 | 64.8 | 77.9 | **69.7** | <u>74.7</u> | <u>77.2</u> |
| DR-Net [29] | 71.3 | **62.8** | **72.2** | 69.8 | 66.9 | 59.9 | <u>79.4</u> | 63.5 | 66.4 | 75.9 |
| DSRR† [32] | **72.7** | 61.5 | <u>71.3</u> | 71.3 | 67.8 | <u>65.1</u> | **79.8** | 67.4 | **75.3** | **78.6** |
| **RVL-BERT** | <u>72.3</u> | <u>62.5</u> | 70.3 | **71.9** | **70.2** | <u>65.1</u> | 78.5 | 68.0 | 74.0 | 75.5 |
| Human Perf. [180] | 94.6 | 90.0 | 96.3 | 95.0 | 95.8 | 94.5 | 95.7 | 88.8 | 93.2 | 94.1 |

#### 3.4.4.2 SpatialSense Dataset

We compare our model with various recent methods, including some methods that have been compared in the VRD experiments.

- **L-baseline**, **S-baseline** and **L+S-baseline** are baselines in [180] taking in simple language and/or spatial features and classifying with fully-connected layers.

- **ViP-CNN** [88] utilizes a phrase-guided message passing structure to model relationship triplets.

- **DR-Net** [29] exploits statistical dependency between object classes and predicates.

- **DSRR** [32] is a concurrent work[5] that exploits depth information for relationship detection with an additional depth estimation model.

- The **Human Performance** result is extracted from [180] for reference.

Table 3.4 shows that our full model outperforms almost all existing approaches in terms of the overall accuracy and obtains the highest (like `in` and `in front`

---
[5]Originally published in August 2020.

Table 3.5: Performance comparison with existing models on VG dataset. Results of the existing methods are extracted from the respective papers. Models are evaluated in PredCls mode.

| Model | Recall@50 | Recall@100 |
|---|---|---|
| Joint CNN [193] | 27.9 | 35.0 |
| Graph R-CNN [179] | 54.2 | 59.1 |
| Message Passing [174] | 59.3 | 61.3 |
| Freq [191] | 59.9 | 64.1 |
| Freq+Overlap [191] | 60.6 | 62.2 |
| SMN [191] | 65.2 | 67.1 |
| KERN [15] | 65.8 | 67.7 |
| VCTree [149] | **66.4** | **68.1** |
| **RVL-BERT** | 62.9 | 66.6 |

`of`) or second-highest accuracy for several relationships. While the concurrent work DSRR achieves a slightly higher overall recall, we expect our model to gain another performance boost with the additional depth information introduced in their work.

#### 3.4.4.3 Visual Genome Dataset

We also experiment on Visual Genome (VG) dataset [80]. We follow [174, 191, 148] to adopt the most widely-used dataset split which consists of 108K images and includes the most frequent 150 object classes and 50 predicates. When evaluating visual relationship detection/scene graph generation on VG, there are three common evaluation modes including (1) Predicate Classification (PredCls): ground truth bounding boxes and object labels are given, (2) Scene Graph Classification (SGCls): only ground truth boxes given, and (3) Scene Graph Detection (SGDet): nothing other than input images is given. We experiment with PredCls, which is a similar setting to what we perform on VRD dataset [109] and SpatialSense dataset [180].

Comparison results on VG are presented in Table 3.5, where our proposed RVL-BERT achieves competitive results. Notably, as our model only learn from object pairwise features, it lags behind graph reason-based approaches including SMN, KERN and VCTree. This can be verified with the ablation studies conducted in [191] where SMN's vanilla version (without context learning) achieves R@50/100 of 63.7/65.6, which is at the same level of ours 62.9/66.6.

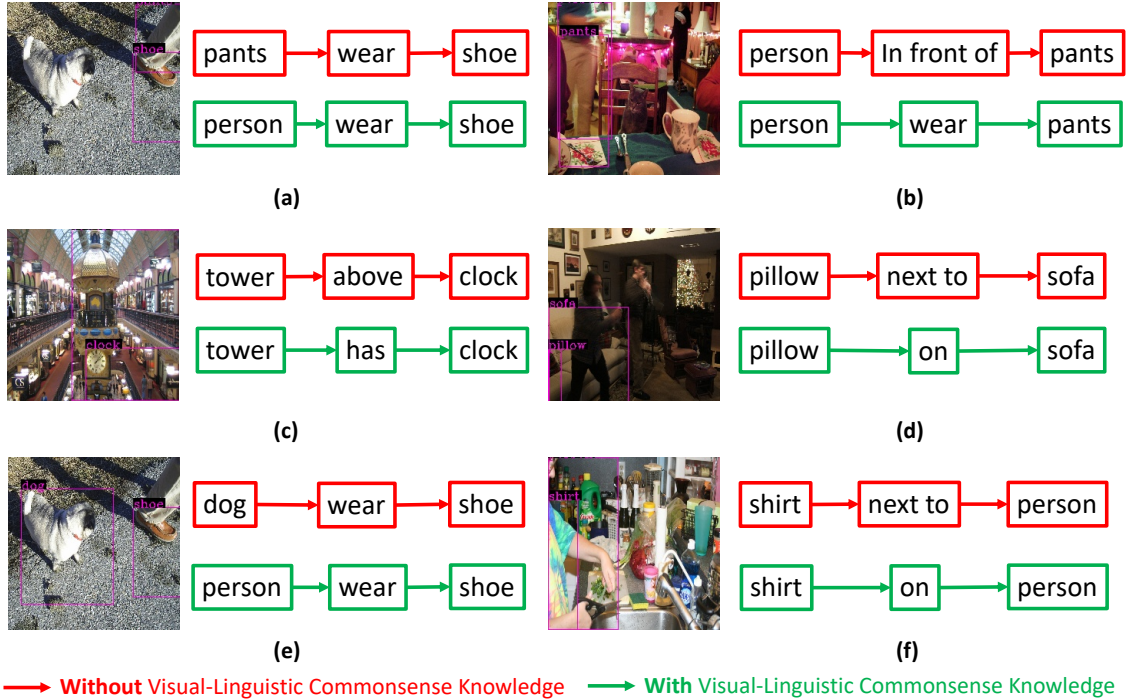

Figure 3.4: Qualitative comparisons between predicting visual relationship with or without visual-linguistic knowledge. Red boxes and arrows denotes predicting with the model without the knowledge, while the green boxes and arrows means predicting with the knowledge. This visualizations are performed during testing on the VRD dataset [109].

In addition, We also note that as mentioned in section 3.4.1, visual relationship detection can be biased as predicates could be "guessed" accurately given explicit correlations between object labels and predicates. Both SMN [191] and KERN [15] exploit this property and use the frequency bias and object co-occurrence, respectively. However, the usage of bias could reversely undermine the capability of generalization which has been demonstrated by comparing mean recall in recent works (e.g., [148]).

### 3.4.5 Qualitative Results

#### 3.4.5.1 Visual-Linguistic Knowledge

Figure 3.4 shows qualitative comparisons between predicting visual relationships with and without the visual-linguistic knowledge in our model. Especially:

1. The example (a) in the figure shows that, with *linguistic* knowledge, a `person`

is more likely to `wear a shoe`, rather than `pants` to `wear a shoe`. That is, the conditional probability $p(\texttt{wear}|(\texttt{person-shoe}))$ becomes higher, while $p(\texttt{wear}|(\texttt{pants-shoe}))$ becomes lower and does not show up in top 100 confident triplets after observing the linguistic fact.

2. The same to 1 also applies to the example (b) (where `person-wear-pants` is more appropriate than `person-in front of-pants`) and the example (c) (where `tower-has-clock` is semantically better than `tower-above-clock`).

3. On the other hand, as the `person` in the example (e) is visually occluded, the model without *visual* knowledge prefers to `dog-wear-shoe` rather than `person-wear-shoe`; however, our model with the visual knowledge knows that the occluded part is likely to be a person and is able to make correct predictions.

4. The same to 3 also applies to the example (d) (where both `pillow` and `sofa` are not clear) and (f) (where `person` is obscure).

These examples demonstrate the effectiveness of our training strategy of exploiting rich visual and linguistic knowledge by pre-training on weakly-labeled visual-linguistic datasets.

We also refer to Appendix A.3 of [143] for a visualization of attention maps (after pre-training) as our model shares the same pre-training procedure as theirs. Their visualization results show that some attention heads are trained to attend text tokens to the associated object regions, which means the pre-training procedure promotes the alignment of visual and linguistic contents. This is effectively similar to our Mask Attention Module which aims to align text tokens to the respective object regions.

### 3.4.5.2 Mask Attention Module

The mask attention module aims to teach the model to learn and predict the attention maps emphasizing the locations of the given object labels. To study its effectiveness, we visualize the attention maps that are generated by the mask attention module during testing on the both datasets in Figure 3.5. The first two rows show two examples from SpatialSense, while the last two rows show three examples from the VRD dataset. Since the input embeddings of the model for the SpatialSense dataset in the form of triplets `subject-predicate-object`, and the VRD dataset in the form of doublets `subject-object` are different, three and two attention maps are generated for each example, respectively.

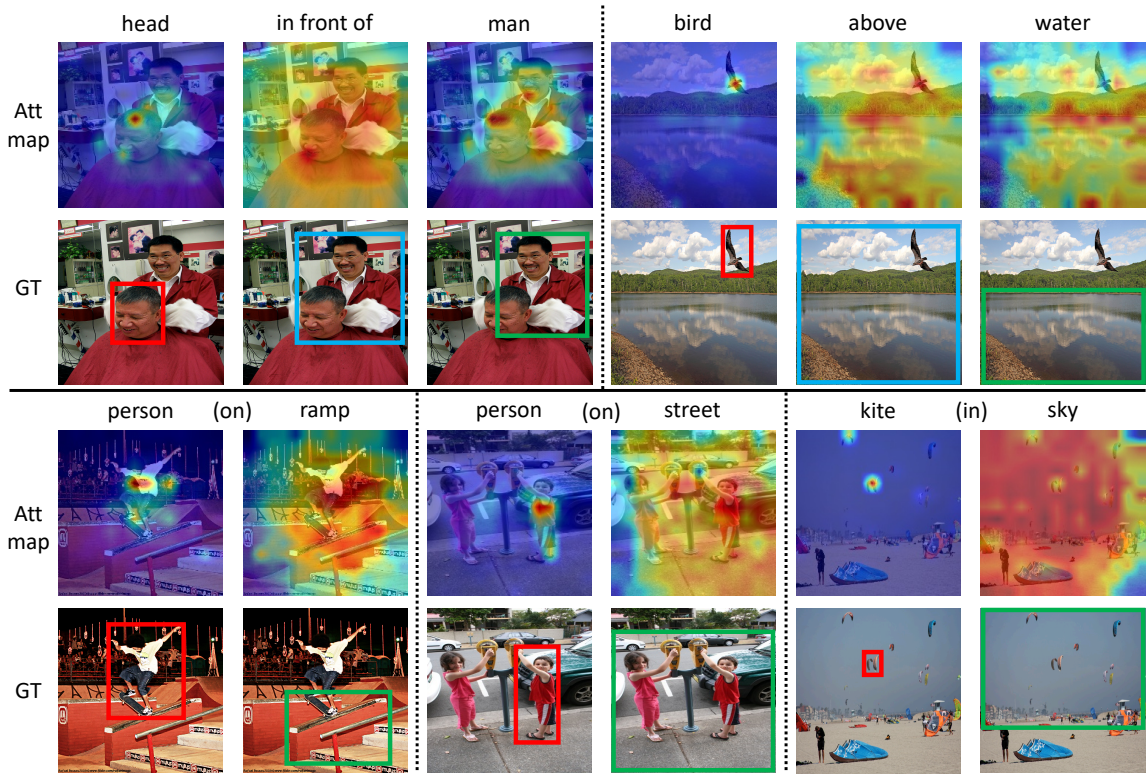

Figure 3.5: Attention map visualization of SpatialSense (the first two rows) and VRD dataset (the last two rows). For each example, the first row shows predicted attention maps while the second shows ground truth bounding boxes.

For the both datasets, the model is actively attending to the region that contains the target object. Especially for the triplet data from SpatialSense, the model is also looking at the union bounding boxes which include cover both subjects and objects. For example, for the top-left example `head-in front of-man` in Figure 3.5, mask attention first looks at the person's `head` who is getting a haircut, followed by attending to the joint region of `head` and the barber (`man`), then finally focus on the barber `man`.

We also observe that, for objects (classes) with larger size, mask attention tends to look more widely at the whole image. For instance, `water` of the top-right example and `sky` of the bottom-right example attend to almost the whole image. We conjecture that this is due to the larger size of objects or regions making it harder to learn to focus on the specific target areas. In addition, We also find the mask attention module learns better with triplet inputs than doublets inputs and this is assumedly because the additional examples of union boxes provide more contexts and facilitate the learning process.

## 3.5 Summary

In this chapter, we proposed a novel visual relationship detection system named RVL-BERT, which exploits visual knowledge in addition to linguistic knowledge learned during self-supervised pre-training on large-scale external datasets. A novel mask attention module is designed to help the model learn to capture the distinct spatial information and a spatial module is utilized to emphasize the bounding box coordinates. We have shown that the effectiveness of the proposed modules and its competitive performance through ablation studies, quantitative and qualitative experiments on two challenging visual relationship detection datasets.

# Chapter 4

# Unbiased Scene Graph Generation

In Chapter 3 we have worked on VRD with external knowledge and we have successfully established a VRD model for representing visual scenes in a structured manner. However, we note that there are still a couple of issues that should be solved for learning better structured representations.

First, as discussed in Section 2.1 and 2.2, VRD models [128, 109, 88, 202, 193, 118, 187] usually learn visual relation triplets *independently*, ignoring the important object and relation contexts in the same image. SGG methods [191, 179, 57, 15, 16, 45, 13, 34, 189, 74, 148, 102, 164, 177, 77, 53, 165, 190, 134, 185, 122, 167, 62, 152, 188, 24, 19], in contrast, connect relation pair proposals into a graph structure on which the hidden states (of object and relation proposals) are refined with surrounding contexts. Thanks to the stronger performance, the use of SGG methods is becoming more prevalent among recent efforts in learning visual relation-based structured representations. SGG has also been shown to be helpful for image captioning [182, 181, 87], visual question answering [150, 136], indoor scene understanding [3, 20] and thus has been drawing increasing attention. Therefore, we shift the focus to SGG and utilize graph-based message passing approaches [191, 149] in this chapter. Notably, we found that there are already efforts showing the benefit of prior/external knowledge to SGG [45, 189, 134], similar to our proposed approach in Chapter 3.

Second, long-tailed problem [197] which causes a trained model to favor a few frequent classes over a larger number of infrequent classes is prevalent in SGG [148]. This issue is especially critical in SGG as most of the tail predicate classes (*e.g.*, `parked on`, `flying in`) are more descriptive than the head classes (*e.g.*, `on`, `in`). The task of generating unbiased and informative scene graphs is called unbiased

SGG [148, 177, 164, 186, 86]. While in Chapter 3 the prior knowledge (along with the proposed model modules) has been shown to benefit VRD performance, we suspect that it is the "bad" knowledge from the datasets (both internal datasets, *e.g.*, Visual Genome [80], and external datasets, *e.g.*, Conceptual Captions) that causes the long-tailed problem in SGG. Specifically, the labeling biases including *bounded rationality* [139] and *reporting bias* [115] should be removed/offset. We thus propose to study how we can debias SGG models by offsetting the labeling biases a model learned during training on a biased dataset.

## 4.1 Introduction

The long tail problem is common and challenging in SGG [148]: since certain predicates (*i.e.,* head classes) occur far more frequently than others (*i.e.,* tail classes) in the most widely-used VG dataset [80], a model that trained with this unbalanced dataset would favor predicting heads against tails. For instance, the number of training examples of `on` is ~830× higher than that of `painted on` in the VG dataset, and (given ground truth objects) a classical SGG model MOTIFS [191] achieves 74.3 Recall@20 for `on`, in sharp contrast to 0.0 for `painted on`. However, the fact that the head classes are less descriptive than the tail classes makes the generated scene graphs coarse-grained and less informative, which is not ideal.

Most of the existing efforts in long-tailed SGG [16, 148, 177, 164, 167, 53] deal with the skewed class distribution directly. However, unlike common long-tailed classification tasks where the long tails are mostly caused by the unbalanced class prior distributions, the long tail of SGG with the VG dataset is significantly affected by the imbalance of missing labels, which remains unsolved. The missing label problem arises as it is unrealistic to annotate the overwhelming number of possible visual relations (*i.e.,* $KN(N-1)$ possibilities given $K$ predicate classes and $N$ objects in an image). Training SGG models by treating all unlabeled pairs as negative examples (which is the default setting for most of the existing SGG works) introduces *missing label bias* in predictions, *i.e.,* predicted probabilities could be under-estimated. What is worse, labeling biases [115, 139, 148] which are prevalent in the VG dataset cause an imbalance in the missing labels of different predicates. That is, the conspicuous classes (*e.g.,* `on`, `in`) are more likely to be annotated than the inconspicuous ones (*e.g.*, `parked on`, `covered in`). Generally, conspicuous classes are more extensively annotated and have higher *label frequencies*, *i.e.*, the fraction of labeled, positive examples in all examples of individual classes. The unbalanced label frequency distribution means that the predicted probability of an inconspicuous class could be under-estimated more than that of a conspicuous one, causing a long tail. To produce meaningful scene graphs, the inconspicuous but informative predicates need to be properly predicted. To the best of our knowledge, none of the existing SGG debiasing methods [16, 148, 177, 164, 167, 53] effectively solve this labeling bias problem.

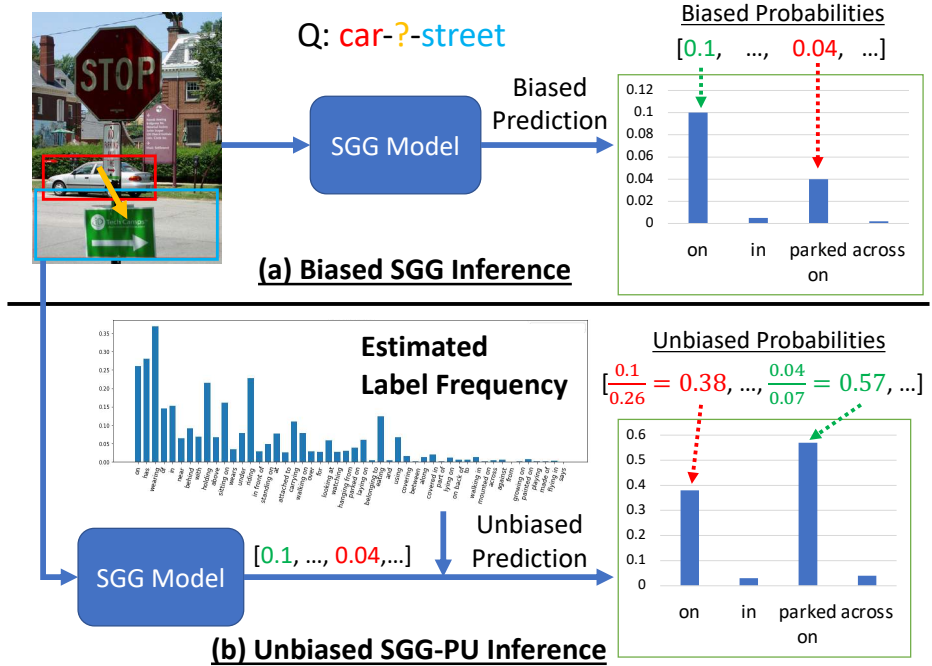

Figure 4.1: An illustrative comparison between the traditional, biased inference and the unbiased PU inference for SGG. (a) Traditionally SGG models are not trained in the PU setting and thus output biased probabilities in favor of conspicuous classes (*e.g.*, `on`). (b) We remove the labeling bias from the biased probabilities by discounting the difference in the chance of being labeled, *i.e.*, label frequency, so that inconspicuous classes (*e.g.*, `parked on`) are properly predicted.

We propose to tackle the labeling bias problem by removing the effect of unbalanced label frequency distribution. That is, we aim to recover the unbiased version of per-class predicted probabilities such that they are independent of the per-class missing label bias. To do this, we first show that learning a SGG model

CHAPTER 4. UNBIASED SCENE GRAPH GENERATION

with the VG dataset can viewed as a *Learning from Positive and Unlabeled data* (PU learning) [30, 36, 6] problem, where a target PU dataset contains only positive examples and unlabeled data. For clarity, we define that a *biased model* is trained on a PU dataset by treating the unlabaled data as negatives and outputs *biased probabilities*, while an *unbiased model* is trained on a fully-annotated dataset and outputs *unbiased probabilities.* Under the PU learning setting, the per-class unbiased probabilities are proportional to the biased ones with the per-class label frequencies as the proportionality constants [36]. Motivated by this fact, we propose to recover the unbiased visual relationship probabilities from the biased ones by dividing by the estimated per-class label frequencies so that the imbalance (*i.e.,* labeling bias) can be offset. Especially, the inconspicuous predicates with their probabilities being under-estimated more could then be predicted with higher confidences so that the scene graphs are more informative. An illustrative comparison of the traditional, biased method and our unbiased one is shown in Fig. 4.1.

A traditional estimator of label frequencies is the per-class average of biased probabilities on a training/validation set predicted by a biased model [36]. While this estimator can work in the easier SGG settings where ground truth bounding boxes are given, *i.e.*, PredCls and SGCls, it is found unable to provide estimates for some classes in the hardest SGG setting where no additional information other than images is provided, *i.e.* SGDet. The reason is that there are no *valid* examples (*i.e.*, predicted object pairs that match ground truth boxes and object labels) can be used for label frequency estimation. For instance, by forwarding a trained MOTIFS [191] model on VG training set, 9 out of 50 predicates do not have even a single valid example, making it impossible to estimate. We propose to take advantage of the training-time data augmentation such as random flipping to increase the number of valid examples. That is, instead of performing post-training estimations, we propose Dynamic Label Frequency Estimation (DLFE) utilizing augmented training examples by maintaining a moving average of the per-batch biased probability during training. The significant increase in the number of valid examples, especially in SGDet, enables accurate label frequency estimation for unbiased probability recovery.

Our contribution in this work is three-fold. First, we are among the first to tackle the long tail problem in SGG from the perspective of labeling bias, which we remove by recovering the per-class unbiased probability from the biased one with a PU based approach. Second, to obtain accurate label frequency estimates for recovering unbiased probabilities in SGG, we propose DLFE which takes advantage of training-time data augmentation and averages over multiple training iterations to introduce more valid examples. Third, we show that DLFE provides more reliable label frequency estimates than a naive variant of the traditional estimator, and we demonstrate that SGG models with DLFE effectively alleviates the long tail and achieve state-of-the-art debiasing performance with remarkably more informative scene graphs. We will release the source code to facilitate research towards an unbiased SGG methodology. This chapter including all of the texts, figures, tables, illustrations, equations is based on our published paper [18].

## 4.2 Related Work

### 4.2.1 Unbiased Scene Graph Generation

Unbiased scene graph generation [148, 177, 164, 186, 86] aims to generate more informative and descriptive scene graphs and has gained increased attention over the past few years. Among the first to discuss on unbiased SGG, Tang et al. [148] point out that labeling biases, *i.e.*, bounded rationality [139] and reporting bias [115], cause SGG models to predict easier (`person`,`beside`,`table` rather than `person`,`eating on`,`table`) or more apparent (`person`,`on`,`bile` instead of `person`,`ride on`,`bike`) predicates, respectively. They then propose to remove the biases, by performing *total direct effect* (TDE) inference with biased SGG models. Specifically, they subtract the counterfactual bias (*e.g.*, pseudo-likelihood with wiped-out instances) from the observed outcome (*e.g.*, likelihood with original instances). Following this effort, Yan et al. [177] propose to train unbiased SGG models by re-weighting with class relatedness-aware weights. Wang et al. [164] transfer the less-biased knowledge from the secondary learner to the main one with knowledge distillation. This work is similar to TDE in that we also tackle the report bias and the bounded rationality; however, we are different in that we view the bias(es) as a result of unbalanced label frequency distribution (from the perspective of positive-unlabeled learning), which we can accurately estimate using dataset-level statistics and we remove via a simple post-processing technique.

### 4.2.2 Positive Unlabeled (PU) learning

While the traditional classification setting aims to learn classifiers with both positive and negative data, *Learning from Positive and Unlabeled data*, or *Positive Unlabeled* (PU) learning, is a variant of the traditional setting where a PU dataset contains only positive and unlabeled examples [30, 36, 6]. That is, an unlabeled example can either be truly a negative, or belongs to one or more classes. Learning a biased classifier assuming all unlabeled examples are negative (which is the default setting for most of the existing SGG works) could introduce missing label bias, producing unbalanced predictions. Common PU learning methods can be roughly divided into two categories [6]: (a) training an unbiased model, and (b) inferencing a biased model in a PU manner. We adopt the latter approach in this work due to its convenience and favorable flexibility.

We note that while Chen et al. [13] also deal with SGG in the PU setting, they do not dive deep into the long tail problem in scene graphs as we do in this work. They propose a three-stage approach which generates pseudo-labels for the unlabeled examples with a biased trained model, followed by training a less biased model with the additional "positive" examples. However, their approach is time and resource consuming since it requires re-generating pseudo labels if different SGG models are used. Unlike [13], our approach not only can be easily adapted for any SGG model with minimal modification, but is superior in terms of debiasing performance.

## 4.3 Methodology

Scene graph generation aims to generate a graph $G$ comprising bounding boxes $B$, object labels $O$, and visual relationships $S$, given an input image $I$. The SGG task $P(G|I)$ is usually decomposed into three components for joint training [191]:

$$P(G|I) = P(B|I)P(O|B,I)P(S|O,B,I), \qquad (4.1)$$

where $P(B|I)$ denotes proposal generation, $P(O|B,I)$ means object classification and $P(S|O,B,I)$ is relationship prediction. We propose to biasedly train a SGG model while we dynamically estimate the label frequencies during training. The estimated label frequencies are then used to recover the unbiased probabilities during inference.

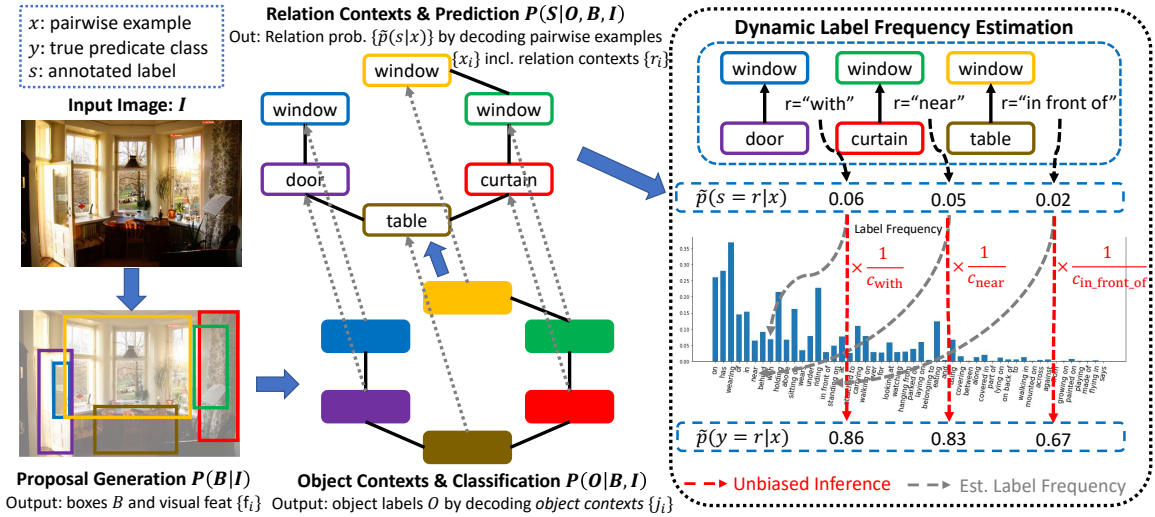

Figure 4.2: An illustration of training and inferencing a SGG model in a PU manner with Dynamic Label Frequency Estimation (DLFE). Given an input image, proposals and their features are extracted by an object detector. Object classification is performed via message passing on a (*e.g.,* chained [191]) graph followed by *object contexts* decoding. Object contexts together with bounding boxes and features are then fed into another graph to refine into *relation contexts*, followed by decoding into the *biased probabilities* $\tilde{p}(s|x)$. DLFE dynamically estimates the label frequencies $c$ with the moving averages of biased probabilities during training. Finally, the unbiased probability of class $r$ is recovered with $\tilde{p}(y=r|x) = \frac{1}{c_r}\tilde{p}(s=r|x)$ during inference.

We describe our choice of proposal generation, object classification and relationship prediction in Section 4.3.1. We then explain how we recover the unbiased probabilities from the biased ones from a PU perspective in Section 4.3.2, followed by presenting our Dynamic Label Frequency Estimation (DLFE) in Section 4.3.3. Figure 4.2 shows an illustration of DLFE applied to SGG models like [191, 149].

## 4.3.1 Model Components

### 4.3.1.1 Proposal Generation $P(B|I)$

Given an image $I$, we adopt a pre-trained object detector [123] to extract $N$ object proposals $B = \{b_i | i = 1, ..., N\}$, together with their visual representation $\{f_i | i = 1, ..., N\}$ and $N(N-1)$ union bounding box representations $\{f_{i,j} | i, j = 1, ..., N\}$ pooled from the output feature map. The visual representations also come

with predicted class probabilities: $\{p_i | i = 1, ..., N\}$ and $\{p_{i,j} | i, j = 1, ..., N\}$.

#### 4.3.1.2 Object Classification $P(O|B, I)$

For object classification, a graphical representation is constructed which takes in object features $\{f_i\}$ and class probabilities $\{p_i\}$ and outputs *object context* $\{j_i\}$ refined with message passing algorithms. We experiment our methods with either chained-structured graphs [191] with bi-directional LSTM [59], or tree-structured graphs [149] with TreeLSTM [146]. The output object contexts are then fed into a linear layer followed with a Softmax layer to decode into predicted object labels $O = \{o_i | i = 1, ..., N\}$.

#### 4.3.1.3 Relationship Prediction $P(S|O, B, I)$

Similar to that of object classification, another graphical representation of the same type is established to propagate contexts between features. The module takes in both the object labels $O$ and the object contexts $\{j_i\}$ and outputs refined *relation contexts* $\{r_i\}$. For each object pair $\{(i, j) | i, j = 1, ..., N, i \neq j\}$, their relation contexts $(r_i, r_j)$, bounding boxes $(b_i, b_j)$, union bounding boxes $b_{ij}$ and features $f_{i,j}$ are gathered into a pairwise feature $x_{ij}$ for decoding into a probability vector over the $K$ classes with MLPs followed by a Softmax layer.

### 4.3.2 Recovering the Unbiased Scene Graphs

Learning SGG from a dataset with missing labels can be viewed as a PU learning problem, which is different from the traditional classification in that (a) no negative examples are available, and (b) unlabeled examples can either be truly negatives or belong to any class. Learning classifiers from a PU dataset by treating all unlabeled data as negatives could introduce strong *missing label bias* [36], *i.e.,* predicted probabilities could be under-estimated, and labeling biases [115, 139], *i.e.,* predicted probability of an inconspicuous class could be under-estimated more than that of a conspicuous one. We propose to avoid the both biases by recovering the unbiased probabilities, marginalizing the effect of uneven label frequencies.

Given $K$ predicate classes, we denote the visual relation examples taken in by the relationship prediction module of a SGG model by a set of tuple $(x, y, s)$, with $x$ an example (*i.e.,* pairwise object features), $y \in \{0, ..., K\}$ the true predicate class (0 means the background class) and $s \in \{0, ..., K\}$ the relation label (0 means unannotated). The class $y$ cannot be observed from the dataset: while we can derive $y = s$ if the example is labeled ($s \neq 0$), $y$ can be any number ranging from 0 to $K$ for an unlabeled example ($s = 0$).

For clarity, we now regard $x$, $y$ and $s$ as random variables. For a target class $r \in \{1, ..., K\}$, a biased SGG model is trained to predict the *biased probability* $P(s = r|x)$, which can be derived as follows:

$$P(s = r|x) = P(s = r, y = r|x) \quad \text{(by PU definition)} \tag{4.2}$$
$$= P(y = r|x)P(s = r|y = r, x), \tag{4.3}$$

where $P(s = r|y = r, x)$ is the probability of example $x$ being selected to be labeled and is called *propensity score* [6]. Dividing each side by $P(s = r|y = r, x)$ we obtain the *unbiased probability* $P(y = r|x)$:

$$P(y = r|x) = \frac{P(s = r|x)}{P(s = r|y = r, x)}. \tag{4.4}$$

However, as discussed in PU learning literature [36, 6], generally we have no idea whether an unlabeled example being truly a negative or a false negative; we must make some assumptions about either the label mechanism or the class distributions of data (or both) to enable learning with these PU data. One of such label mechanism assumption is *Selected Completely At Random* (SCAR) [6]: non-background examples are selected for labeling entirely at random regardless of $x$, *i.e.,* the set of labeled examples is uniformly (i.i.d.) drawn from the set of positive examples, which is not realistic but still reasonable[1]. This means we bypass the dependence on each $x$ so that $P(s = r|y = r, x) = P(s = r|y = r)$, and Eq. 4.4 can be re-written as

$$P(y = r|x) = \frac{P(s = r|x)}{P(s = r|y = r)}, \tag{4.5}$$

where $P(s = r|y = r)$ is the *label frequency* of class $r$, or $c_r$, which is the fraction of labeled examples in all the examples of class $r$. Notably, discounting the effect of per-class label frequencies in this way also removes the labeling bias. Since label frequencies are usually not provided by annotators, an estimation is required.

---
[1]There exists other label mechanism assumptions, such as *Selected At Random* (SAR) and *Probabilistic Gap PU* (PGPU), and data assumptions such as *separability* and *smoothness*. Other PU learning methods, *e.g.*, the *two-step techniques* assuming both the separability and smoothness, might also be utilized. Note that we do not discuss whether other assumptions are more suitable for the VG dataset; instead, we *empirically* show that SCAR-enabled PU learning method helps unbiased SGG trained on the VG dataset. Please refer to [6] for more details on PU learning and [5, 7] for deeper discussions on SCAR and SAR.

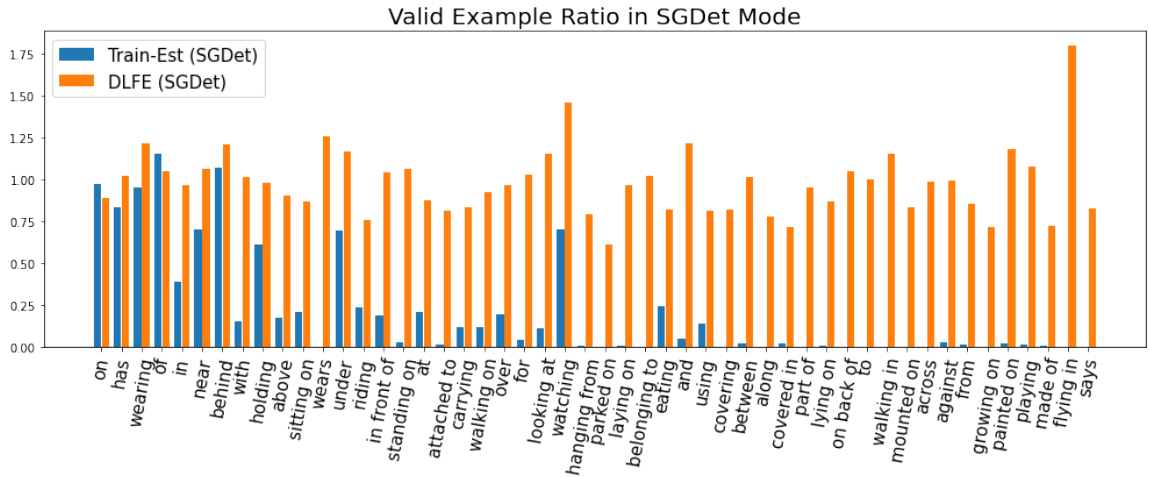

Figure 4.3: The per-class ratio of *valid* examples in all examples of VG150 training set [174] (SGDet), by inferencing a trained MOTIFS (Train-Est) or dynamically inferencing a training MOTIFS with augmented data (DLFE). Numbers of DLFE are averaged over all epochs. All numbers can exceed 1 as a ground truth pair can match multiple proposal pairs.

### 4.3.3 Dynamic Label Frequency Estimation

One of the most common estimators of label frequencies, named Train-Est, is the per-class average of biased probabilities $\tilde{p}(s|x)$ predicted by a biased model [36]. Denote an annotated example in a training or validation set with $(x, y, s)$, where $x$ is the pairwise example, $s$ is the relation label and $y$ is the true class. Note $y = r$ as we are only considering the annotated ones. Referring to [36], the biased probability $p(s|x)$ of class $r$ can be derived as follows:

$$\begin{aligned}
p(s = r|x) &= p(s = r|x, y = r)p(y = r|x) \\
&\quad + p(s = r|x, y \neq r)p(y \neq r|x) \\
&= p(s = r|x, y = r) \times 1 + 0 \times 0 \quad \text{(since y=r)} \\
&= p(s = r|y = r), \quad \text{(the SCAR assumption)}
\end{aligned}$$

where $p(s = r|y = r)$ is the label frequency of class $r$. Thus, we can obtain a reasonable label frequency estimate via averaging the per-class biased probability

with a training or validation set:

$$c_r = P(s = r | y = r) \approx \frac{1}{N_r} \sum_{(x, y=r) \in D} \tilde{p}(s = r | x), \tag{4.6}$$

where $D$ denotes a training or validation set and $N_r$ is the cardinality of $\{(x, y = r)\}$.

However, we find this way of estimation inconvenient and unsuitable for SGG. To understand why, recall that PredCls, SGCls and SGDet are the three SGG training and evaluation settings, and note that re-estimation of label frequencies is required for each setting since the expected biased probabilities could vary depending on the task difficulty[2]. Firstly, the post-training estimation required before inferencing in each SGG setting is inconvenient and unfavorable. Secondly, the absence of ground truth bounding boxes in SGDet mode results in lack of *valid* examples for label frequency estimation. For a proposal pair to be valid, its both objects must match ground truth boxes (with IoU $\geq 0.5$) and object labels simultaneously. By using Train-Est with MOTIFS [191], as revealed in in Fig. 4.3 (the blue bars), 9 out of 50 predicates do not have even a valid example, *i.e.*, $\{(x, y = r)\}$ in Eqn. 4.6 is empty, making it impossible to compute. In addition, more valid examples are missing for inconspicuous classes: as the examples of those classes are concentrated in a much smaller number of images, not matching a bounding box could invalidate lots of examples. A naive remedy is using a default value for those missing estimates; however, as we show in section 4.4.3 the performance is sub-optimal.

To alleviate this problem, we propose to take advantage of the training-time data augmentation to get back more valid examples for tail classes. Concretely, during training we augment input data by horizontal flipping with a probability of 0.5, and meanwhile we estimate label frequencies with per-batch biased probabilities. By doing this, the number of valid examples of tail classes could become more normal (and higher) than that of Train-Est, since averaging over augmented examples and multiple training iterations (with varying object label predictions) essentially introduces more samples, which in turn increases the number of valid examples.

Based on this idea, we propose Dynamic Label Frequency Estimation (DLFE) where the main steps are shown in Algorithm 1. In detail, we maintain per-class moving averages of the biased probabilities (Eqn. 4.6) throughout the training. The

---

[2]Using label frequencies estimated in other mode is found to degrade the performance.

---

**Algorithm 1:** DLFE during training time
    **Input**   : Training dataset $D^t$ and momentum $\alpha$
    **Output:** Biased model $g(\cdot)$ and estimated label frequency $c$
1: **for** *each mini batch* $\mathcal{B} = \{(x_i, s_i)\} \in D^t$: **do**
2:     Forward model to obtain the biased probabilities $g(x)$;
    // in-batch average of biased probabilities
3:     **for** *each predicate class* $r \in \{1, ..., K\}$: **do**
4:         $\mathcal{B}' \leftarrow \{(x_i, s_i) \in \mathcal{B} | s_i = r\}$;
5:         $c'_r \leftarrow \frac{1}{|\mathcal{B}'|} \sum_{(x,s) \in \mathcal{B}'} g(s|x)$;
        // Update the exponential moving average
6:         $\tilde{c}_r \leftarrow \alpha \times c'_r + (1 - \alpha) \times \tilde{c}_r$;
7:     **end**
8: **end**
    // Save for inference use
9: $c \leftarrow \tilde{c}$;

---

estimated label frequencies $\tilde{c}$ are dynamically updated by the per-batch averages $c'$ with a momentum $\alpha$ so that the estimates that are more recent matter more:

$$\tilde{c} \leftarrow \alpha \times c' + (1 - \alpha) \times \tilde{c}. \tag{4.7}$$

Note that for each mini-batch we update the estimated $\tilde{c}_r$ of class $r$ only if at least one valid example of $r$ presents in the current batch. The estimated values gradually stabilize along with the converging model, and we save the final estimates $c \in \mathbb{R}^K$ (as a vector of length $K$). During inference, the estimated label frequencies are utilized to recover the unbiased probability distribution $\tilde{p}(y|x)$ from the biased one $\tilde{p}(s|x)$ by

$$\tilde{p}(y|x) = \frac{1}{c} \odot \tilde{p}(s|x), \tag{4.8}$$

where $\odot$ denotes the Hadamard (element-wise) product. The average per-epoch number of valid examples of this way is shown in Fig. 4.3 (the tangerine bars), where the inconspicuous classes get remarkably more (4× or more) valid examples. This not only enables accurate estimations for all the classes but makes the estimations easier as no additional, post-training estimation is required.

## 4.4 Experiments

### 4.4.1 Evaluation Settings

We follow the recent efforts in SGG [191, 15] to experiment on a subset of the VG dataset [80] named VG150 [174], which contains 62,723 images for training, 5,000 for validation and 26,446 for testing. As discussed earlier, we train and evaluate in the three SGG settings: PredCls, SGCls and SGDet. We evaluate models with, or without *graph constraint*: whether only a single relation with the highest confidence is predicted for each object pair. Non-graph constraint is denoted as "ng". For evaluation, we adopt recall-based metrics which measures the fraction of ground truth visual relations appearing in top-$K$ confident predictions, where $K$ is 20, 50, or 100. However, as the plain recall could be dominated by a biased model predicting mostly head classes, we follow [15, 149, 177, 164] to average over per-class recall and focus on the less biased per-class/mean recall (mR@$K$) and non-graph constraint per-class/mean recall (ng-mR@$K$). We note that the ng per-class/mean recall should be the fairest measure for debiasing methods since it 1) treats each class equally and 2) reflects the fact that more than one visual relations could exist for an object pair. We follow the long-tailed recognition research [106] to divide the distribution into three parts, including head (many-shot; top-15 frequent predicates), middle (medium-shot; mid-20) and tail (few-shot; last-15) and compute their ng-mRs. Note that by DLFE in this section, we mean the dynamic label frequency estimation along with our unbiased scene graph recovery approach.

### 4.4.2 Implementation Details

As DLFE is a model-agnostic strategy, we experiment with two popular SGG backbones: MOTIFS [191] and VCTree [149]. Following [148, 164], we adopt a pre-trained and frozen Faster R-CNN [123] with ResNeXt-101-FPN [171, 99] as the object detector, which achieves 28.14 mAP on VG's testing set [148]. All the hyperparameters, including the momentum $\alpha = 0.1$, are tuned with the validation set. All models are trained using SGD optimizer with the initial learning rate of 0.01 after the first 500 iterations of warm-up. Random flipping is applied to all the training examples. The learning rate is decayed, for a maximum of twice, by the factor of 10 once the validation performance plateaus twice consecutively. Training can early stop when the maximum decay step (two) is reached before the maximum 50,000 iterations. The final checkpoint is used for evaluation. The batch size for all experiments is 48 (images). For SGDet setting, we sample 80 proposals from each image and apply per-class NMS [125]. Beside ground truth visual relations, we follow [148] to sample up to $1,024$ pairs with background-to-ground truth ratio being 3:1.

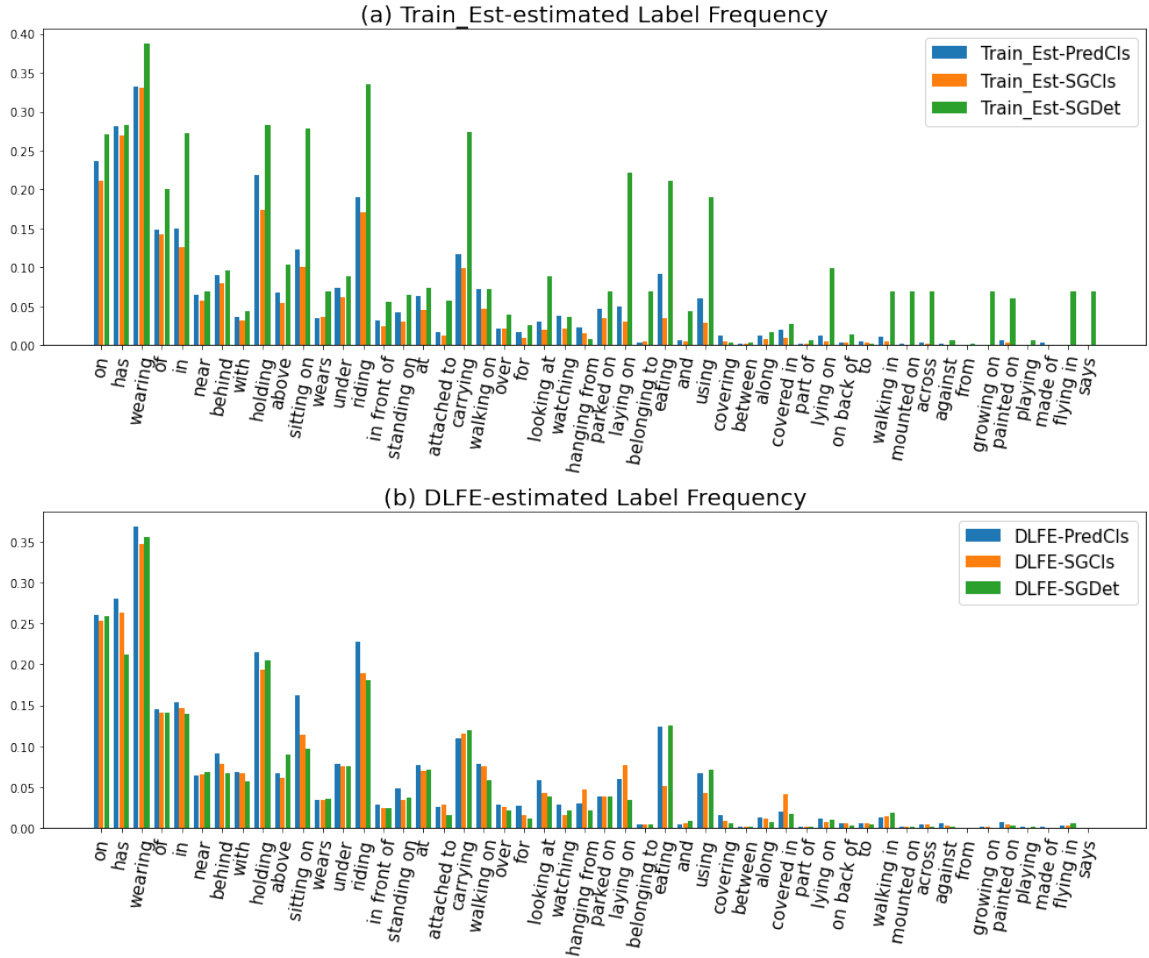

Figure 4.4: The label frequencies estimated by (a) Train-Est or (b) DLFE with MOTIFS [191]. The classes with higher label frequency are more *conspicuous* than those with a lower one. Predicates sorted by class frequency in descending order.

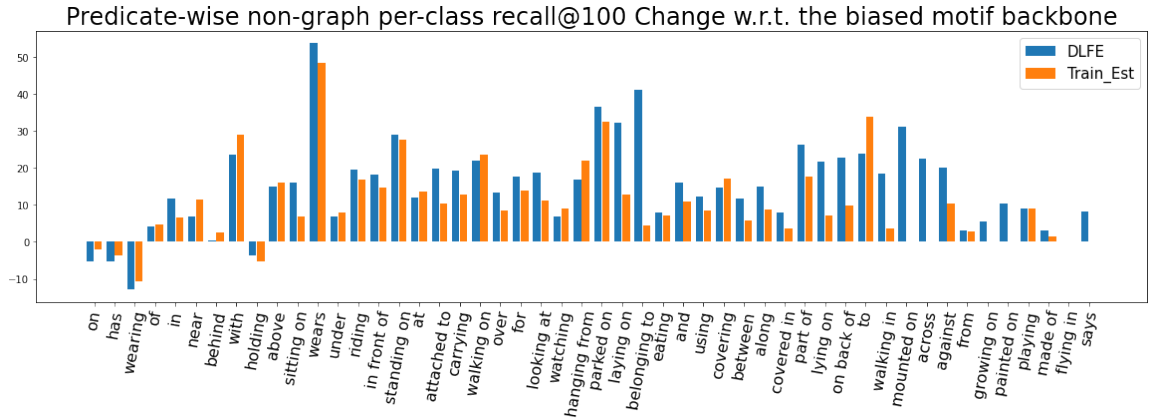

Figure 4.5: The absolute ng per-class R@100 changes when recovering MOTIFS's [191] unbiased probabilities with the label frequencies estimated by Train-Est or DLFE, in SGDet mode.

### 4.4.3 Comparing DLFE to Train-Est

We aim to answer the question: *whether DLFE is more effective in estimating label frequency than Train-Est*, by comparing 1) the consistencies of the estimated label frequencies and 2) their debiasing performance. As discussed earlier that label frequencies of the predicates lacking a valid example cannot be estimated by Train-Est, we thus naively assign the median of the other estimated label frequencies to those missing values.

A comparison of the estimated label frequencies is presented in Fig. 4.4. It is clear from (a) that, even for the classes that with at least one valid example, the Train-Est estimated values tend to be abnormally high in SGDet setting. Note that while there might be differences in estimated values for different SGG settings, with a same backbone they should still be relatively similar. In contrast, (b) shows DLFE-estimated values are more consistent across the three settings. We also compare their debiasing performance in SGDet with the absolute ng per-class R@100 change in Fig. 4.5. Apparently, Train-Est barely improves the per-class recalls especially for tail classes that lacks enough valid examples, while DLFE achieves higher and more consistent improvement across all the predicates.

We also present results in non-graph constraint mean recall (ng-mR@$K$) in Table 4.1. Across both the MOTIFS [191] and VCTree [149] backbone, our DLFE achieves significantly higher ng-mRs in both PredCls and SGDet setting and is on par with Train-Est in SGCls setting.

|  | PredCls | | | SGCls | | | SGDet | | |
| --- | --- | --- | --- | --- | --- | --- | --- | --- | --- |
| Model | 20 | 50 | 100 | 20 | 50 | 100 | 20 | 50 | 100 |
| MOTIFS [191, 148] | 19.9 | 32.8 | 44.7 | 11.3 | 19.0 | 25.0 | 7.5 | 12.5 | 16.9 |
| MOTIFS-Train-Est [36] | 24.4 | 38.9 | 50.5 | 17.1 | **26.1** | **32.8** | 8.9 | 14.1 | 18.9 |
| **MOTIFS-DLFE** | **30.0** | **45.8** | **57.7** | **17.6** | 25.6 | 32.0 | **11.7** | **18.1** | **23.0** |
| VCTree [149, 148] | 21.4 | 35.6 | 47.8 | 12.4 | 19.1 | 25.5 | 7.5 | 12.5 | 16.7 |
| VCTree-Train-Est [36] | 25.0 | 39.1 | 52.4 | 21.0 | **32.2** | **39.4** | 8.1 | 13.0 | 17.1 |
| **VCTree-DLFE** | **29.1** | **44.6** | **56.8** | **21.6** | 31.4 | 38.8 | **11.7** | **17.5** | **22.5** |

Table 4.1: Comparison of non-graph constraint mean recalls (ng-mR@$K$, $K \in \{20, 50, 100\}$) between Train-Est and our DLFE, in PredCls, SGCls and SGDet.

These results verify the claim that, apart from being more convenient (requiring no post-training estimation), DLFE is more effective than naive Train-Est for providing more reliable estimates.

### 4.4.4 Comparing to other Debiasing Methods

While we list the results of different SGG backbone model for reference, we mainly compare our approach with the model-agnostic debiasing methods including *Focal Loss* [100], *Resampling* [9], *Reweighting*, *L2+{u,c}KD* [164], *TDE* [148], *PCPL* [177] and *STL* [13].

- L2+{u,c}KD is a two-learner knowledge distillation framework for reducing dataset biases.

- TDE is an inference-time debiasing method which ensembles counterfactual thinking by removing context-specific bias.

- PCPL learns the relatedness scores among predicates which are used as the weights in reweighting, and is the current state-of-the-art in terms of mR.

- STL generates soft pseudo-labels for unlabeled data used for joint training.

We re-implement PCPL and STL since their backbone is not directly comparable, and Reweighting since, to the best of our knowledge, there does not exist its reported performance for VCTree. We report our reproduced results of TDE with the authors' codebase [147].

CHAPTER 4.  UNBIASED SCENE GRAPH GENERATION|  | PredCls | | | SGCls | | | SGDet | | |
|---|---|---|---|---|---|---|---|---|---|
| Model | 20 | 50 | 100 | 20 | 50 | 100 | 20 | 50 | 100 |
| KERN [15] | - | 36.3 | 49.0 | - | 19.8 | 26.2 | - | 11.7 | 16.0 |
| GB-Net-$\beta^{\diamond}$ [189] | - | 44.5 | 58.7 | - | 25.6 | 32.1 | - | 11.7 | 16.6 |
| MOTIFS$^{\dagger}$ [191, 164] | 19.9 | 32.8 | 44.7 | 11.3 | 19.0 | 25.0 | 7.5 | 12.5 | 16.9 |
| MOTIFS-Reweight$^{\ddagger}$ | 20.5 | 33.5 | 44.4 | 12.6 | 19.1 | 24.3 | 8.0 | 12.9 | 16.8 |
| MOTIFS-L2+uKD$^{\ddagger}$ [164] | - | 36.9 | 50.9 | - | 22.7 | 30.1 | - | 14.0 | 19.5 |
| MOTIFS-L2+cKD$^{\ddagger}$ [164] | - | 37.2 | 50.8 | - | 22.1 | 29.6 | - | 14.2 | 19.8 |
| MOTIFS-TDE$^{\dagger}$ [148] | 18.7 | 29.0 | 38.2 | 10.7 | 16.1 | 21.1 | 7.4 | 11.2 | 14.9 |
| MOTIFS-PCPL$^{\dagger}$ [177] | 25.6 | 38.5 | 49.3 | 13.1 | 19.9 | 25.6 | 9.8 | 14.8 | 19.6 |
| MOTIFS-STL$^{\dagger}$ [13] | 15.7 | 29.4 | 43.2 | 10.3 | 18.4 | 27.2 | 6.4 | 10.6 | 15.0 |
| **MOTIFS-DLFE** | **30.0** | **45.8** | **57.7** | **17.6** | **25.6** | **32.0** | **11.7** | **18.1** | **23.0** |
| VCTree$^{\dagger}$ [149, 164] | 21.4 | 35.6 | 47.8 | 14.3 | 23.3 | 31.4 | 7.5 | 12.5 | 16.7 |
| VCTree-Reweight$^{\ddagger}$ | 20.6 | 32.5 | 41.6 | 14.1 | 21.3 | 27.8 | 8.0 | 12.1 | 15.9 |
| VCTree-L2+uKD$^{\ddagger}$ [164] | - | 37.7 | 51.7 | - | 26.8 | 35.2 | - | 13.8 | 19.1 |
| VCTree-L2+cKD$^{\ddagger}$ [164] | - | 38.4 | 52.4 | - | 26.8 | 35.8 | - | 13.9 | 19.0 |
| VCTree-TDE$^{\dagger}$ [148] | 20.9 | 32.4 | 41.5 | 12.4 | 19.1 | 25.5 | 7.8 | 11.5 | 15.2 |
| VCTree-PCPL$^{\dagger}$ [177] | 25.1 | 38.5 | 49.3 | 17.2 | 25.9 | 32.7 | 9.9 | 15.1 | 19.9 |
| VCTree-STL$^{\dagger}$ [13] | 16.8 | 31.8 | 45.1 | 12.7 | 22.0 | 32.7 | 6.0 | 10.0 | 14.1 |
| **VCTree-DLFE** | **29.1** | **44.6** | **56.8** | **21.6** | **31.4** | **38.8** | **11.7** | **17.5** | **22.5** |

Table 4.2: Performance comparison in ng-mR@$K$ ($K \in \{20, 50, 100\}$) on VG150 [80, 174]. Models in the first section are with VGG16 backbone [140]. $\dagger$ models implemented or reproduced ourselves with ResNeXt-101-FPN [99] backbone. $\ddagger$ models also with the same ResNeXt-101-FPN backbone while their performance are reported by the respective papers. $\diamond$ model using external knowledge bases.

The results in ng-mR@$K$ are presented in Table 4.2, where DLFE significantly improves ng-mRs for both MOTIFS and VCTree and outperforms the existing debiasing methods. Notably, TDE, which was proposed to alleviate the long tail by removing the context-specific bias, is shown to even *adversely* affect the ability of predicting multiple relations for per object pair. This means that TDE undermines the ability of ranking possible predicates for object pairs, which is suspectedly because TDE is overly strong in suppressing the predicates other than the top ones. This result also shows that removing the labeling bias is more beneficial for debiasing SGG models.

While the graph-constraint mR metric does not reflect the fact that multiple relations could exist between an object pair, due to its popularity we still present the results in Table 4.3. Debiasing MOTIFS with our proposed DLFE still significantly improves its mR, achieving state-of-the-art mR across all the three setting. Large performance boost are also seen in VCTree with DLFE, and new SOTAs are attained for PredCls (mR@20), SGCls and SGDet.

|  | PredCls | | | SGCls | | | SGDet | | |
|---|---|---|---|---|---|---|---|---|---|
| Model | 20 | 50 | 100 | 20 | 50 | 100 | 20 | 50 | 100 |
| IMP+ [174, 15] | - | 9.8 | 10.5 | - | 5.8 | 6.0 | - | 3.8 | 4.8 |
| FREQ [191, 149] | 8.3 | 13.0 | 16.0 | 5.1 | 7.2 | 8.5 | 4.5 | 6.1 | 7.1 |
| MOTIFS [191, 149] | 10.8 | 14.0 | 15.3 | 6.3 | 7.7 | 8.2 | 4.2 | 5.7 | 6.6 |
| KERN [15] | - | 17.7 | 19.2 | - | 9.4 | 10.0 | - | 6.4 | 7.3 |
| VCTree [149] | 14.0 | 17.9 | 19.4 | 8.2 | 10.1 | 10.8 | 5.2 | 6.9 | 8.0 |
| GPS-Net [102] | 17.4 | 21.3 | 22.8 | 10.0 | 11.8 | 12.6 | 6.9 | 8.7 | 9.8 |
| GB-Net-$\beta^{\diamond}$ [189] | - | 22.1 | 24.0 | - | 12.7 | 13.4 | - | 7.1 | 8.5 |
| MOTIFS[†] [191, 148] | 13.0 | 16.5 | 17.8 | 7.2 | 8.9 | 9.4 | 5.3 | 7.3 | 8.6 |
| MOTIFS-Focal[‡] [100, 148] | 10.9 | 13.9 | 15.0 | 6.3 | 7.7 | 8.3 | 3.9 | 5.3 | 6.6 |
| MOTIFS-Resample[‡] [9, 148] | 14.7 | 18.5 | 20.0 | 9.1 | 11.0 | 11.8 | 5.9 | 8.2 | 9.7 |
| MOTIFS-Reweight[†] | 14.3 | 17.3 | 18.6 | 9.5 | 11.2 | 11.7 | 6.7 | 9.2 | 10.9 |
| MOTIFS-L2+uKD[‡] [164] | 14.2 | 18.6 | 20.3 | 8.6 | 10.9 | 11.8 | 5.7 | 7.9 | 9.5 |
| MOTIFS-L2+cKD[‡] [164] | 14.4 | 18.5 | 20.2 | 8.7 | 10.7 | 11.4 | 5.8 | 8.1 | 9.6 |
| MOTIFS-TDE[†] [148] | 17.4 | 24.2 | 27.9 | 9.9 | 13.1 | 14.9 | 6.7 | 9.2 | 11.1 |
| MOTIFS-PCPL[†] [177] | 19.3 | 24.3 | 26.1 | 9.9 | 12.0 | 12.7 | 8.0 | 10.7 | 12.6 |
| MOTIFS-STL[†] [13] | 13.3 | 20.1 | 22.3 | 8.5 | 12.8 | 14.1 | 5.4 | 7.6 | 9.1 |
| **MOTIFS-DLFE** | **22.1** | **26.9** | **28.8** | **12.8** | **15.2** | **15.9** | **8.6** | **11.7** | **13.8** |
| VCTree[†] [149, 148] | 14.1 | 17.7 | 19.1 | 9.1 | 11.3 | 12.0 | 5.2 | 7.1 | 8.3 |
| VCTree-Reweight[†] | 16.3 | 19.4 | 20.4 | 10.6 | 12.5 | 13.1 | 6.6 | 8.7 | 10.1 |
| VCTree-L2+uKD[‡] [164] | 14.2 | 18.2 | 19.9 | 9.9 | 12.4 | 13.4 | 5.7 | 7.7 | 9.2 |
| VCTree-L2+cKD[‡] [164] | 14.4 | 18.4 | 20.0 | 9.7 | 12.4 | 13.1 | 5.7 | 7.7 | 9.1 |
| VCTree-TDE[†] [148] | 19.2 | **26.2** | **29.6** | 11.2 | 15.2 | 17.5 | 6.8 | 9.5 | 11.4 |
| VCTree-PCPL[†] [177] | 18.7 | 22.8 | 24.5 | 12.7 | 15.2 | 16.1 | 8.1 | 10.8 | 12.6 |
| VCTree-STL[†] [13] | 14.3 | 21.4 | 23.5 | 10.5 | 14.6 | 16.6 | 5.1 | 7.1 | 8.4 |
| **VCTree-DLFE** | **20.8** | 25.3 | 27.1 | **15.8** | **18.9** | **20.0** | **8.6** | **11.8** | **13.8** |

Table 4.3: Performance comparison of SGG models in graph-constraint mR@$K$ ($K \in \{20, 50, 100\}$) on VG150 [80, 174] testing set. Models in the first section are with VGG16 while the others are with ResNeXt-101-FPN. †, ‡ and ◇ are with the same meanings as in Table 4.2.

While results in graph constraint recalls (R@$K$) and mean recalls (mR@$K$) are less reflective of how unbiased a SGG model is, we still provide them for reference. We note that the plain recall (R@$K$) is biased as it favors the head classes and

|  | Predicate Classification | | | | | |
|---|---|---|---|---|---|---|
| Model | R@20 | R@50 | R@100 | mR@20 | mR@50 | mR@100 |
| IMP+ [174, 15] | 52.7 | 59.3 | 61.3 | - | 9.8 | 10.5 |
| FREQ [191, 149] | 53.6 | 60.6 | 62.2 | 8.3 | 13.0 | 16.0 |
| UVTransE [62] | - | 61.2 | 64.3 | - | - | - |
| MOTIFS [191, 149] | 58.5 | 65.2 | 67.1 | 10.8 | 14.0 | 15.3 |
| KERN [15] | - | 65.8 | 67.6 | - | 17.7 | 19.2 |
| NODIS [188] | 58.9 | 66.0 | 67.9 | - | - | - |
| HCNet [122] | 59.6 | 66.4 | 68.8 | - | - | - |
| VCTree [149] | 60.1 | 66.4 | 68.1 | 14.0 | 17.9 | 19.4 |
| GPS-Net [102] | 60.7 | 66.9 | 68.8 | 17.4 | 21.3 | 22.8 |
| GB-Net-$\beta^{\diamond}$ [189] | - | 66.6 | 68.2 | - | 22.1 | 24.0 |
| HOSE-Net [167] | - | 66.7 | 69.2 | - | - | - |
| Part-Aware [152] | 61.8 | 67.7 | 69.4 | 15.2 | 19.2 | 20.9 |
| DG-PGNN [74] | - | 69.0 | 72.1 | - | - | - |
| MOTIFS$^{\dagger}$ [191, 148] | 59.0 | 65.5 | 67.2 | 13.0 | 16.5 | 17.8 |
| MOTIFS-Focal$^{\ddagger}$ [100, 148] | 59.2 | 65.8 | 67.7 | 10.9 | 13.9 | 15.0 |
| MOTIFS-Resample$^{\ddagger}$ [9, 148] | 57.6 | 64.6 | 66.7 | 14.7 | 18.5 | 20.0 |
| MOTIFS-Reweight$^{\dagger}$ | 45.4 | 54.7 | 56.5 | 14.3 | 17.3 | 18.6 |
| MOTIFS-L2+uKD$^{\ddagger}$ [164] | 57.4 | 64.1 | 66.0 | 14.2 | 18.6 | 20.3 |
| MOTIFS-L2+cKD$^{\ddagger}$ [164] | 57.7 | 64.6 | 66.4 | 14.4 | 18.5 | 20.2 |
| MOTIFS-TDE$^{\dagger}$ [148] | 32.9 | 45.0 | 50.6 | 17.4 | 24.2 | 27.9 |
| MOTIFS-PCPL$^{\dagger}$ [177] | 48.4 | 54.7 | 56.5 | 19.3 | 24.3 | 26.1 |
| MOTIFS-STL$^{\dagger}$ [13] | 56.5 | 65.0 | 66.9 | 13.3 | 20.1 | 22.3 |
| MOTIFS-DLFE | 46.4 | 52.5 | 54.2 | 22.1 | 26.9 | 28.8 |
| VCTree$^{\dagger}$ [149, 148] | 59.8 | 65.9 | 67.5 | 14.1 | 17.7 | 19.1 |
| VCTree-Reweight$^{\dagger}$ | 53.8 | 60.7 | 62.6 | 16.3 | 19.4 | 20.4 |
| VCTree-L2+uKD$^{\ddagger}$ [164] | 58.5 | 65.0 | 66.7 | 14.2 | 18.2 | 19.9 |
| VCTree-L2+cKD$^{\ddagger}$ [164] | 59.0 | 65.4 | 67.1 | 14.4 | 18.4 | 20.0 |
| VCTree-TDE$^{\dagger}$ [148] | 34.4 | 44.8 | 49.2 | 19.2 | 26.2 | 29.6 |
| VCTree-PCPL$^{\dagger}$ [177] | 50.5 | 56.9 | 58.7 | 18.7 | 22.8 | 24.5 |
| VCTree-STL$^{\dagger}$ [13] | 57.1 | 65.2 | 67.0 | 14.3 | 21.4 | 23.5 |
| VCTree-DLFE | 45.7 | 51.8 | 53.5 | 20.8 | 25.3 | 27.1 |

Table 4.4: Recall and mean recall (with graph constraint) results in PredCls task on VG150. Models in the first section are with VGG backbone [140]. $\dagger$, $\ddagger$ and $\diamond$ are with the same meaning as in Table 4.2.

|  | Scene Graph Classification | | | | | |
| --- | --- | --- | --- | --- | --- | --- |
| Model | R@20 | R@50 | R@100 | mR@20 | mR@50 | mR@100 |
| IMP+ [174, 15] | 31.7 | 34.6 | 35.4 | - | 5.8 | 6.0 |
| FREQ [191, 149] | 29.3 | 32.3 | 32.9 | 5.1 | 7.2 | 8.5 |
| UVTransE [62] | - | 30.9 | 32.2 | - | - | - |
| MOTIFS [191, 149] | 32.9 | 35.8 | 36.5 | 6.3 | 7.7 | 8.2 |
| KERN [15] | - | 36.7 | 37.4 | - | 9.4 | 10.0 |
| NODIS [188] | 36.0 | 39.8 | 40.7 | - | - | - |
| HCNet [122] | 34.2 | 36.6 | 37.3 | - | - | - |
| VCTree [149] | 35.2 | 38.1 | 38.8 | 8.2 | 10.1 | 10.8 |
| GPS-Net [102] | 36.1 | 39.2 | 40.1 | 10.0 | 11.8 | 12.6 |
| GB-Net-$\beta^{\diamond}$ [189] | - | 37.3 | 38.0 | - | 12.7 | 13.4 |
| HOSE-Net [167] | - | 36.3 | 37.4 | - | - | - |
| Part-Aware [152] | 36.5 | 39.4 | 40.2 | 8.7 | 10.9 | 11.6 |
| DG-PGNN [74] | - | 39.3 | 40.1 | - | - | - |
| MOTIFS[†] [191, 148] | 36.4 | 39.5 | 40.3 | 7.2 | 8.9 | 9.4 |
| MOTIFS-Focal[‡] [100, 148] | 36.0 | 39.3 | 40.1 | 6.3 | 7.7 | 8.3 |
| MOTIFS-Resample[‡] [9, 148] | 34.5 | 37.9 | 38.8 | 9.1 | 11.0 | 11.8 |
| MOTIFS-Reweight[†] | 24.2 | 29.5 | 31.5 | 9.5 | 11.2 | 11.7 |
| MOTIFS-L2+uKD[‡] [164] | 35.1 | 38.5 | 39.3 | 8.6 | 10.9 | 11.8 |
| MOTIFS-L2+cKD[‡] [164] | 35.6 | 38.9 | 39.8 | 8.7 | 10.7 | 11.4 |
| MOTIFS-TDE[†] [148] | 21.4 | 27.1 | 29.5 | 9.9 | 13.1 | 14.9 |
| MOTIFS-PCPL[†] [177] | 31.9 | 35.3 | 36.1 | 9.9 | 12.0 | 12.7 |
| MOTIFS-STL[†] [13] | 35.4 | 39.9 | 40.9 | 8.5 | 12.8 | 14.1 |
| MOTIFS-DLFE | 29.0 | 32.3 | 33.1 | 12.8 | 15.2 | 15.9 |
| VCTree[†] [149, 148] | 42.1 | 45.8 | 46.8 | 9.1 | 11.3 | 12.0 |
| VCTree-Reweight[†] | 38.0 | 42.3 | 43.5 | 10.6 | 12.5 | 13.1 |
| VCTree-L2+uKD[‡] [164] | 40.9 | 44.7 | 45.6 | 9.9 | 12.4 | 13.4 |
| VCTree-L2+cKD[‡] [164] | 41.4 | 45.2 | 46.1 | 9.7 | 12.4 | 13.1 |
| VCTree-TDE[†] [148] | 21.7 | 28.8 | 32.0 | 11.2 | 15.2 | 17.5 |
| VCTree-PCPL[†] [177] | 36.5 | 40.6 | 41.7 | 12.7 | 15.2 | 16.1 |
| VCTree-STL[†] [13] | 40.6 | 45.7 | 46.9 | 10.5 | 14.6 | 16.6 |
| VCTree-DLFE | 29.7 | 33.5 | 34.6 | 15.8 | 18.9 | 20.0 |

Table 4.5: Recall and mean recall (with graph constraint) results in SGCls task on VG150. Models in the first section are with VGG backbone [140]. †, ‡ and ◇ are with the same meaning as in Table 4.2.

|  | Scene Graph Detection | | | | | |
|---|---|---|---|---|---|---|
| Model | R@20 | R@50 | R@100 | mR@20 | mR@50 | mR@100 |
| IMP+ [174, 15] | 14.6 | 20.7 | 24.5 | - | 3.8 | 4.8 |
| FREQ [191, 149] | 20.1 | 26.2 | 30.1 | 4.5 | 6.1 | 7.1 |
| UVTransE [62] | - | 25.3 | 28.5 | - | - | - |
| MOTIFS [191, 149] | 21.4 | 27.2 | 30.3 | 4.2 | 5.7 | 6.6 |
| KERN [15] | - | 27.1 | 29.8 | - | 6.4 | 7.3 |
| NODIS [188] | 21.5 | 27.4 | 30.7 | - | - | - |
| HCNet [122] | 22.6 | 28.0 | 31.2 | - | - | - |
| VCTree [149] | 22.0 | 27.9 | 31.3 | 5.2 | 6.9 | 8.0 |
| GPS-Net [102] | 22.6 | 28.4 | 31.7 | 6.9 | 8.7 | 9.8 |
| GB-Net-$\beta^{\diamond}$ [189] | - | 26.3 | 29.9 | - | 7.1 | 8.5 |
| HOSE-Net [167] | - | 28.9 | 33.3 | - | - | - |
| Part-Aware [152] | 23.4 | 29.4 | 32.7 | 5.7 | 7.7 | 8.8 |
| DG-PGNN [74] | - | 31.2 | 32.5 | - | - | - |
| MOTIFS$^{\dagger}$ [191, 148] | 25.8 | 33.1 | 37.6 | 5.3 | 7.3 | 8.6 |
| MOTIFS-Focal$^{\ddagger}$ [100, 148] | 24.7 | 31.7 | 36.7 | 3.9 | 5.3 | 6.6 |
| MOTIFS-Resample$^{\ddagger}$ [9, 148] | 23.2 | 30.5 | 35.4 | 5.9 | 8.2 | 9.7 |
| MOTIFS-Reweight$^{\dagger}$ | 18.3 | 24.4 | 29.3 | 6.7 | 9.2 | 10.9 |
| MOTIFS-L2+uKD$^{\ddagger}$ [164] | 24.8 | 32.2 | 36.8 | 5.7 | 7.9 | 9.5 |
| MOTIFS-L2+cKD$^{\ddagger}$ [164] | 25.2 | 32.5 | 37.1 | 5.8 | 8.1 | 9.6 |
| MOTIFS-TDE$^{\dagger}$ [148] | 12.4 | 17.3 | 20.8 | 6.7 | 9.2 | 11.1 |
| MOTIFS-PCPL$^{\dagger}$ [177] | 21.3 | 27.8 | 31.7 | 8.0 | 10.7 | 12.6 |
| MOTIFS-STL$^{\dagger}$ [13] | 22.5 | 29.9 | 34.9 | 5.4 | 7.6 | 9.1 |
| MOTIFS-DLFE | 18.9 | 25.4 | 29.4 | 8.6 | 11.7 | 13.8 |
| VCTree$^{\dagger}$ [149, 148] | 24.1 | 30.8 | 35.2 | 5.2 | 7.1 | 8.3 |
| VCTree-Reweight$^{\dagger}$ | 20.8 | 27.8 | 32.0 | 6.6 | 8.7 | 10.1 |
| VCTree-L2+uKD$^{\ddagger}$ [164] | 24.4 | 31.6 | 35.9 | 5.7 | 7.7 | 9.2 |
| VCTree-L2+cKD$^{\ddagger}$ [164] | 24.8 | 32.0 | 36.1 | 5.7 | 7.7 | 9.1 |
| VCTree-TDE$^{\dagger}$ [148] | 12.3 | 17.3 | 20.9 | 6.8 | 9.5 | 11.4 |
| VCTree-PCPL$^{\dagger}$ [177] | 20.5 | 26.6 | 30.3 | 8.1 | 10.8 | 12.6 |
| VCTree-STL$^{\dagger}$ [13] | 21.6 | 28.8 | 33.6 | 5.1 | 7.1 | 8.4 |
| VCTree-DLFE | 16.8 | 22.7 | 26.3 | 8.6 | 11.8 | 13.8 |

Table 4.6: Recall and mean recall (with graph constraint) results in SGDet task on VG150. Models in the first section are with VGG backbone [140]. $\dagger$, $\ddagger$ and $\diamond$ are with the same meaning as in Table 4.2.

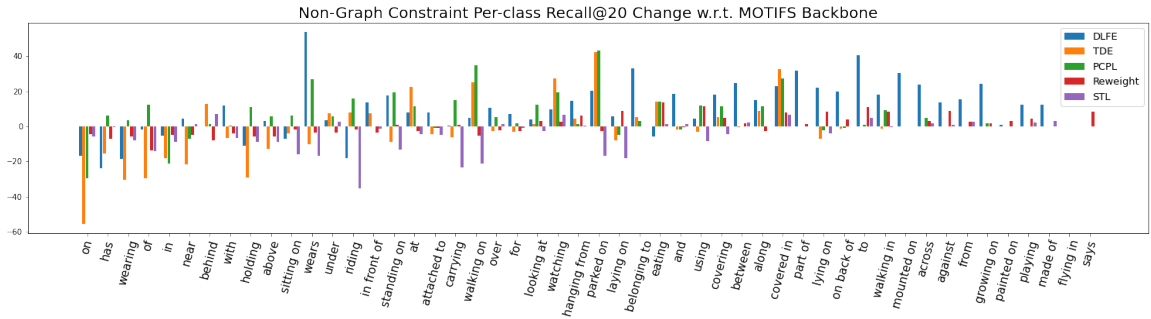

Figure 4.6: Non-graph constraint per-class Recall@20 (PredCls) change w.r.t. MOTIFS baseline. DLFE significantly improves the mid-to-tail recalls (where the other debiasing methods struggle) without compromising much head classes performance.

does not reflect that multiple relations could exist in an object pair. We present a comprehensive comparison of recently published (2020-now) SGG models/debiasing methods[3] in graph-constraint recalls and mean recalls, in PredCls mode in Table 4.4, in SGCls mode in Table 4.5 and in SGDet mode in Table 4.6. While there are some performance drops in more conventional recalls (R@$K$) for the debiasing methods, it is because the predicates are being classified into the more descriptive ones (which do not have been annotated as ground truth). Mean recall (mR@$K$) is less biased than the plain recall as it treats all classes equally; however, it still does not consider the multi-relation issue. Remarkably, it is clear from all the three tables that our DLFE still achieves state-of-the-art mR@$K$ comparing the other debiasing methods with the same backbone, and either MOTIFS-DLFE or VCTree-DLFE attains the highest mR scores across all the models and backbones.

To better understand how DLFE affects the performance of each class, we also present non-graph constraint per-class recall@20 changes compared to MOTIFS backbone-only (biased classifier), in Figure 4.6. While all the debiasing methods increase the recall of the less frequent, middle-to-tail classes, only DLFE improves the tail (last-15) classes' performance significantly. The other approach that also visibly improves the tail classes' performance is Reweighting; however, their relatively small improvements demonstrate that naively dealing with the unbalanced class frequencies is less effective than tackling the labeling bias.

---

[3]Note that we do not compare with [77] as 1) their reported numbers are rather selective and incomplete and 2) their method were not compared with other debiasing methods (*e.g.,* [148]) fairly, *i.e.,* with the same backbone.

|  | Predicate Classification (PredCls) | | | | | |
|---|---|---|---|---|---|---|
|  | Head Recalls | | Middle Recalls | | Tail Recalls | |
| Model | 50 | 100 | 50 | 100 | 50 | 100 |
| MOTIFS[†] [191, 148] | 42.9/65.9 | 45.4/**78.6** | 9.0/30.0 | 10.3/45.4 | 0.0/3.3 | 0.0/9.7 |
| MOTIFS-Reweight[†] | 38.5/57.4 | 40.6/69.2 | 12.6/30.7 | 13.9/43.0 | 2.4/13.3 | 2.9/21.5 |
| MOTIFS-TDE[†] [148] | 40.8/48.3 | 46.3/60.8 | 29.8/34.9 | 34.9/46.1 | 0.0/1.8 | 0.1/5.3 |
| MOTIFS-PCPL[†] [177] | **47.0**/66.5 | **49.5**/77.6 | **25.3**/41.8 | **27.9**/**55.2** | 0.2/6.0 | 0.2/13.2 |
| MOTIFS-STL[†] [13] | 43.4/56.4 | 46.3/70.0 | 13.5/24.1 | 16.5/39.8 | 5.5/9.6 | 6.0/21.2 |
| **MOTIFS-DLFE** | 40.7/61.9 | 42.7/72.4 | **25.3**/**42.8** | **27.9**/54.2 | **15.1**/**31.8** | **15.7**/**44.6** |
| VCTree[†] [149, 148] | 43.8/**67.5** | 46.2/**79.8** | 11.4/34.3 | 13.0/50.0 | 0.0/5.5 | 0.1/12.7 |
| VCTree-Reweight[†] | 43.1/61.6 | 45.4/73.4 | 14.6/28.3 | 15.2/38.3 | 2.3/9.0 | 2.3/14.3 |
| VCTree-TDE[†] [148] | 45.1/54.8 | **50.0**/67.5 | **31.6**/37.9 | **36.2**/49.1 | 0.3/2.5 | 0.5/5.4 |
| VCTree-PCPL[†] [177] | **45.4**/64.5 | 47.8/75.9 | 23.0/**42.6** | 25.4/**54.2** | 0.1/6.9 | 0.1/16.1 |
| VCTree-STL[†] [13] | 43.8/57.6 | 46.5/71.1 | 14.3/26.1 | 17.1/41.8 | **8.4**/13.8 | 9.1/23.5 |
| **VCTree-DLFE** | 37.3/57.5 | 39.6/68.3 | 19.0/36.0 | 20.8/48.2 | 8.1/**26.5** | **9.2**/**38.1** |

Table 4.7: Head, middle and tail (without/with graph constraint) recalls (R@$K$/ng-R@$K$) in the PredCls task on VG150. † and ‡ are with the same meaning as in Table 4.2.

|  | Scene Graph Classification (SGCls) | | | | | |
|---|---|---|---|---|---|---|
|  | Head Recalls | | Middle Recalls | | Tail Recalls | |
| Model | 50 | 100 | 50 | 100 | 50 | 100 |
| MOTIFS[†] [191, 148] | 25.1/**39.1** | 26.3/**45.6** | 3.5/16.6 | 3.9/24.3 | 0.0/2.1 | 0.0/5.4 |
| MOTIFS-Reweight[†] | 25.1/35.7 | 26.3/42.1 | 8.0/16.5 | 8.3/21.9 | 1.6/6.0 | 1.7/9.7 |
| MOTIFS-TDE[†] [148] | 23.5/28.4 | 26.1/35.2 | **15.2**/18.5 | **17.8**/24.3 | 0.0/0.6 | 0.0/2.7 |
| MOTIFS-PCPL[†] [177] | **27.1**/38.7 | **28.3**/44.9 | 9.5/18.5 | 10.3/25.2 | 0.1/3.1 | 0.2/6.6 |
| MOTIFS-STL[†] [13] | 26.4/34.8 | 27.8/42.1 | 8.8/16.2 | 10.2/25.2 | 4.4/5.2 | 5.7/15.1 |
| **MOTIFS-DLFE** | 24.6/36.6 | 25.6/42.0 | 14.2/**23.5** | 15.1/**29.5** | **8.3**/**17.7** | **8.4**/**24.8** |
| VCTree[†] [149, 148] | 29.9/**46.8** | 31.3/**54.7** | 5.8/21.2 | 6.4/31.5 | 0.0/2.6 | 0.0/8.0 |
| VCTree-Reweight[†] | 28.4/41.2 | 29.7/49.3 | 8.5/18.0 | 9.0/25.2 | 1.9/5.9 | 1.9/9.7 |
| VCTree-TDE[†] [148] | 26.2/33.6 | 30.0/42.7 | 18.2/21.2 | **21.2**/28.2 | 0.1/1.9 | 0.1/4.6 |
| VCTree-PCPL[†] [177] | **32.0**/46.2 | **33.5**/53.4 | 13.9/26.3 | 15.2/34.7 | 0.1/5.1 | 0.1/9.4 |
| VCTree-STL[†] [13] | 30.7/40.8 | 32.3/49.7 | 10.3/19.1 | 12.2/30.5 | 4.1/6.9 | 6.6/18.7 |
| **VCTree-DLFE** | 26.9/41.2 | 28.3/48.0 | **18.5**/**29.9** | 19.8/**37.6** | **11.3**/**23.8** | **12.0**/**31.3** |

Table 4.8: Head, middle and tail (without/with graph constraint) recalls (R@$K$/ng-R@$K$) in the SGCls task on VG150. † and ‡ are with the same meaning as in Table 4.2.

|  | Scene Graph Detection (SGDet) | | | | | |
|---|---|---|---|---|---|---|
|  | Head Recalls | | Middle Recalls | | Tail Recalls | |
| Model | 50 | 100 | 50 | 100 | 50 | 100 |
| MOTIFS[†] [191, 148] | 20.7/**28.7** | 24.1/**36.2** | 2.8/9.2 | 3.3/14.2 | 0.0/0.7 | 0.0/1.3 |
| MOTIFS-Reweight[†] | 20.4/25.2 | 23.9/32.2 | 6.4/10.9 | 7.9/14.6 | 1.8/3.1 | 1.9/4.5 |
| MOTIFS-TDE[†] [148] | 17.4/20.4 | 20.9/26.1 | 9.9/12.7 | 12.1/16.6 | 0.0/0.1 | 0.0/1.4 |
| MOTIFS-PCPL[†] [177] | **22.5**/28.5 | **26.0**/35.7 | 9.8/14.5 | 12.0/19.8 | 0.1/1.6 | 0.1/3.3 |
| MOTIFS-STL[†] [13] | 19.2/23.8 | 22.9/31.2 | 4.3/7.9 | 5.4/12.2 | 0.2/1.0 | 0.4/2.6 |
| **MOTIFS-DLFE** | 20.0/27.9 | 23.4/34.1 | **10.9**/**16.9** | **13.3**/**21.9** | **5.3**/**11.1** | **6.7**/**15.1** |
| VCTree[†] [149, 148] | 20.1/**28.2** | 23.3/**35.6** | 2.6/9.6 | 3.3/14.0 | 0.0/0.6 | 0.0/1.3 |
| VCTree-Reweight[†] | 19.6/24.9 | 22.7/31.3 | 6.1/9.8 | 7.2/13.7 | 1.2/2.3 | 1.3/3.4 |
| VCTree-TDE[†] [148] | 17.8/21.7 | 21.4/27.3 | 10.3/12.4 | 12.5/16.5 | 0.0/0.2 | 0.0/1.4 |
| VCTree-PCPL[†] [177] | **22.0**/27.9 | **25.4**/34.7 | 10.3/15.7 | 12.4/20.7 | 0.1/1.5 | 0.1/4.0 |
| VCTree-STL[†] [13] | 18.5/22.8 | 21.8/29.7 | 3.8/7.4 | 4.7/11.5 | 0.1/0.7 | 0.1/1.8 |
| **VCTree-DLFE** | 18.2/25.9 | 21.2/31.9 | **11.3**/**16.8** | **13.1**/**21.3** | **6.0**/**10.2** | **7.2**/**14.7** |

Table 4.9: Head, middle and tail (without/with graph constraint) recalls (R@$K$/ng-R@$K$) in the SGDet task on VG150. † and ‡ are with the same meaning as in Table 4.2.

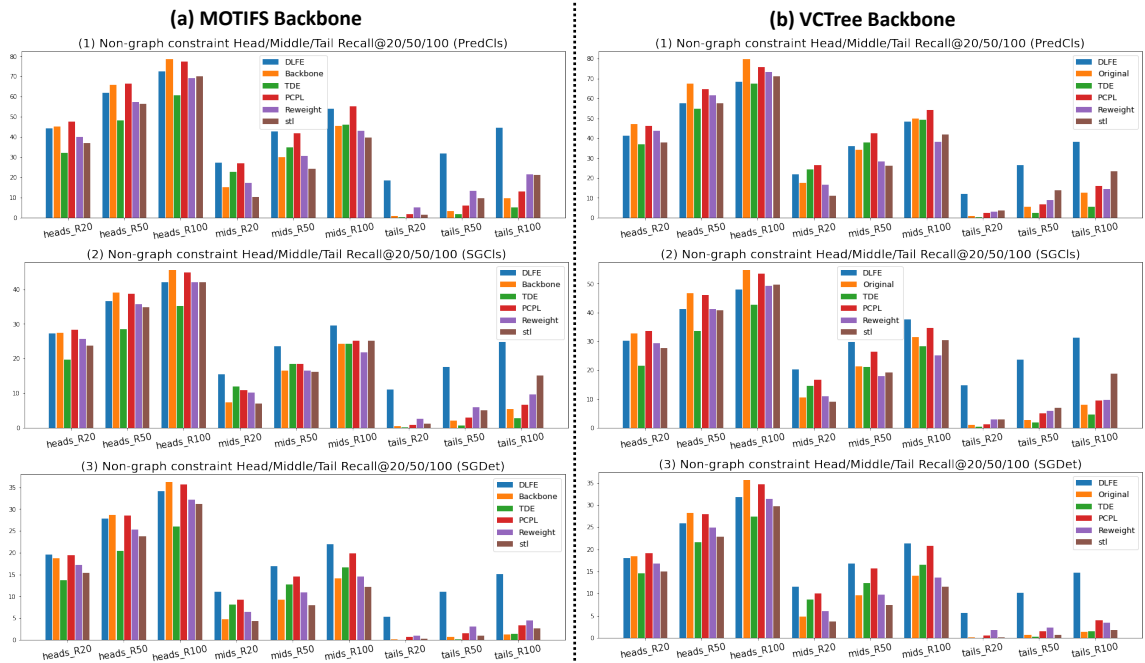

Figure 4.7: Bar plots of head (many shot), middle (medium shot) and tail (few shot) classes based on either MOTIFS (left) and VCTree (right) backbone, evaluated on VG150. From top to down is results in PredCls, SGCls and SGDet, respectively.

We also compare our proposed DLFE with other debiasing methods, *i.e.*, Reweighting, TDE [148], PCPL [177] and STL [13], on the recalls of different part of predicate distribution: (i) head (many-shot), (ii) middle (medium-shot) and (iii) tail (few-shot) recall. The bar plots of head, middle and tail recalls (non-graph constraint) of MOTIFS and VCTree backbones are presented in Figure 4.7. In all the three SGG tasks, both of MOTIFS-DLFE and VCTree-DLFE remarkably outperform other debiasing methods by a large margin regarding the tail recall, while being on par regarding the head and middle recalls.

The recalls with/without graph constraint in numbers for PredCls, SGCls and SGDet task is presented in Table 4.7, Table 4.8 and Table 4.9, respectively. Again, DLFE improves middle and tail recalls more significantly, with the cost of head recall similar to that of other debiasing approaches.

### 4.4.5 Qualitative Results

The scene graphs of three testing images are visualized in Figure 4.8, where the scene graphs on the left side are generated by MOTIFS and those on the right are by MOTIFS-DLFE. (a) is an apparent example that, while `wheel-on-car`, `car-on-street`, `hair-on-man` predicted by MOTIFS are reasonable, `wheel-mounted on-car`, `car-parked on-street`, `hair-belonging to-man` predicted by MOTIFS-DLFE match the ground truth and are also more descriptive (while being inconspicuous). Similarly, `tree-growing on-hill` in Example (b) and `woman-standing on-beach` in (c) are also correct and more descriptive; however, due to the missing label issue in the VG dataset, `tree-growing on-hill` can not be correctly recalled (shown as tangerine color). In addition, there are some seemingly incorrect annotations such as `tree-standing on-tree` in Example (a), where the subject actually indicates a smaller, branch part of a tree. For this object pair, MOTIFS-DLFE predicts `growing on` which, ironically, seems to be more reasonable than the ground truth label.

To understand how DLFE changes the probability distribution, we visualize the biased (MOTIFS) and unbiased (MOTIFS-DLFE) probabilities, given a subject-object pair, in Figure 4.9 and 4.10. Prediction confidences are shown in Figure 4.9 to be calibrated towards minor but expressive predicates like (a) `car-parked on-street`, (b) `wheel-of-train`, (d) `people-sitting on-bench` (while `sitting on` is not in the ground truth). Notably in (c), `fork` is actually not on the plate but was mis-predicted by MOTIFS due to the strong bias (*i.e.,* many `fork-in/on-plate` examples in VG dataset), while MOTIFS-DLFE correctly predicts `near`. Moreover, (b) shows that the confidences of MOTIFS-DLFE for predicates other than the GT `of`, such as `mounted on` and `part of`, have increased remarkably, presumably because they are also reasonable choices.

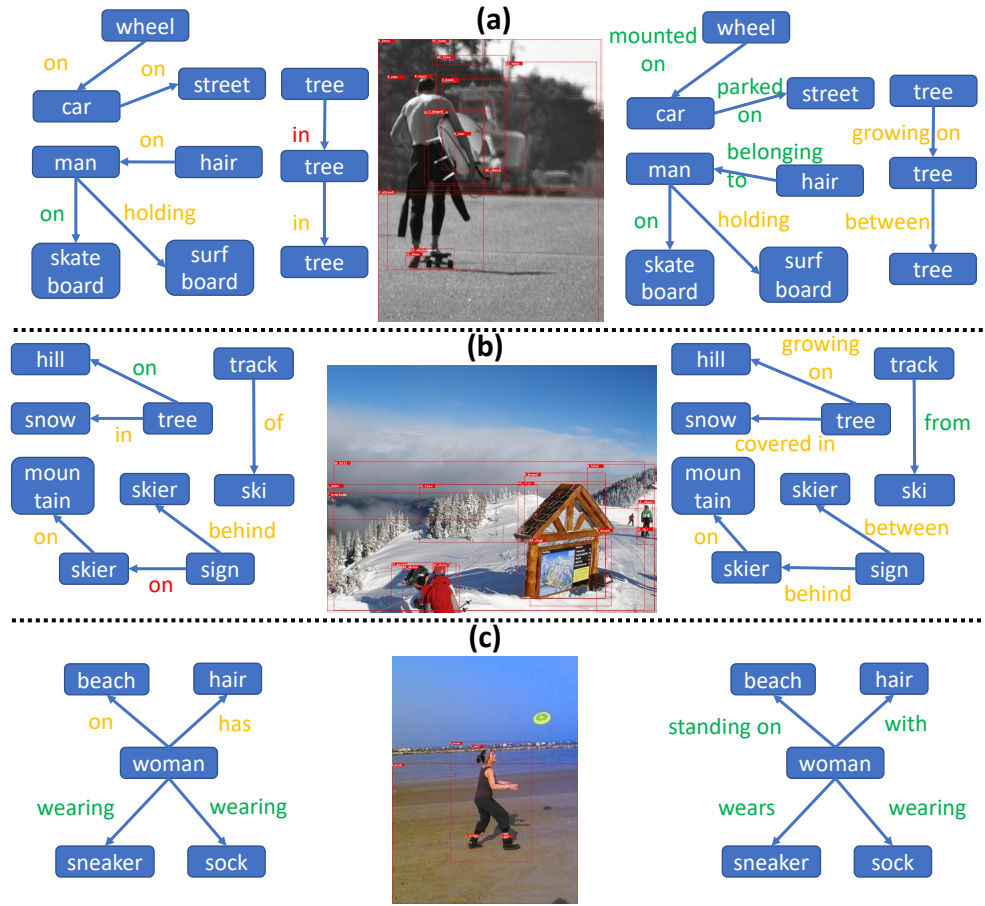

Figure 4.8: Scene graphs generated by MOTIFS (left) and MOTIFS-DLFE (right) in PredCls. Only the top-1 prediction is shown for each object pair. A prediction can be correct (matches GT), incorrect (does not match GT and weird) or acceptable (does not match GT but still reasonable).

Referring to Figure 4.10, for `table-?-window` in example (a), instead of the prediction of `near` by MOTIFS, MOTIFS-DLFE predicts a more descriptive `in front of` which matches the ground truth. The same applies to `bird-?-pole` in example (b) where MOTIFS-DLFE's `standing on` is better than MOTIFS's `on`. (c) is an interesting example that, while the top prediction `on` by MOTIFS-DLFE is the same as that of MOTIFS, more descriptive predicates such as `mounted on`, `attached to` are assigned with the higher confidence scores by MOTIFS-DLFE. This fact should make it easier for, *e.g.*, non-graph constraint mean recall (ng-mR@$K$), to *recall* these fine-grained predicates Finally, `car-?-street` is another example that MOTIFS-DLFE produces more descriptive `parked on` rather than `on`; however, due to the missing labels `parked on` is not in the ground truth. This demonstrates the effectiveness of DLFE for balanced SGG.

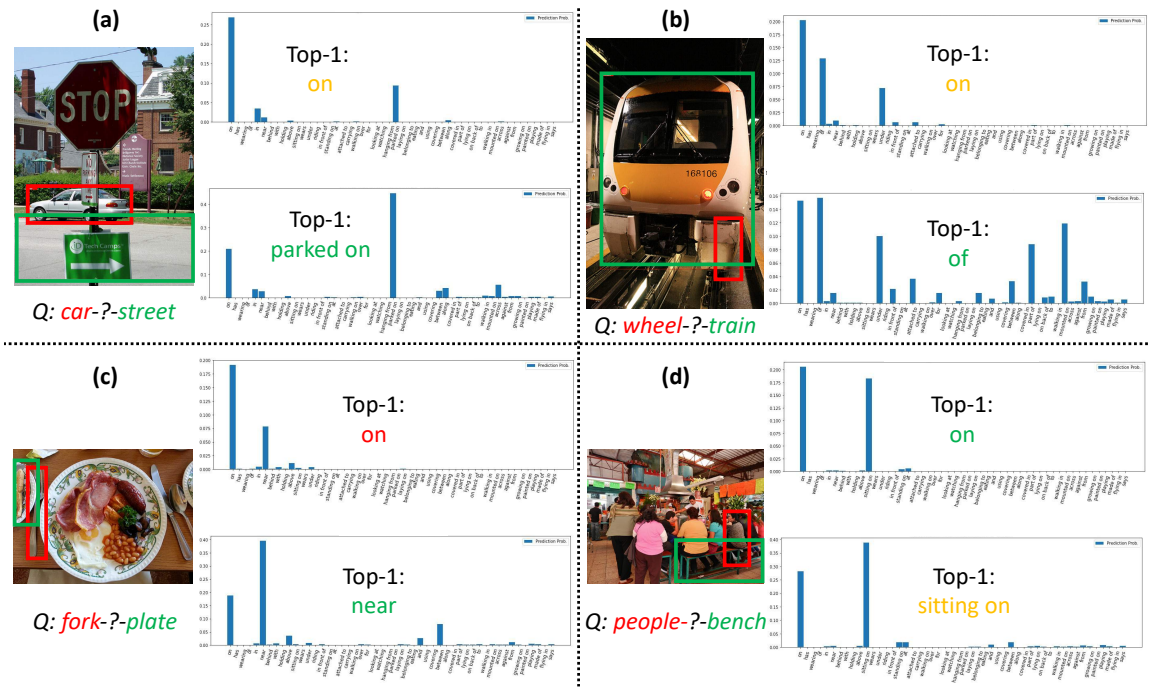

Figure 4.9: Class probability distributions (normalized to sum 1) by MOTIFS (top of each example) and MOTIFS-DLFE (bottom). The top-1 predictions can be correct (GT), incorrect (Non-GT and weird) or acceptable (Non-GT but reasonable).

## 4.5 Discussion

We note that our approach is a debiasing method based on PU learning aiming for the long tail problem especially caused by missing labels. PU learning itself is universal and can be applied to a variety of classification problems, while ours is a variant to better estimate the label frequencies for some tasks (like SGG) lacking enough examples of some classes. We also note that while we shifts the focus to SGG in this chapter, we do not modify underlying SGG architectures such as MOTIFS/VCTree. Instead, we enjoy the stronger visual relationship modeling performance brought by SGG methods.

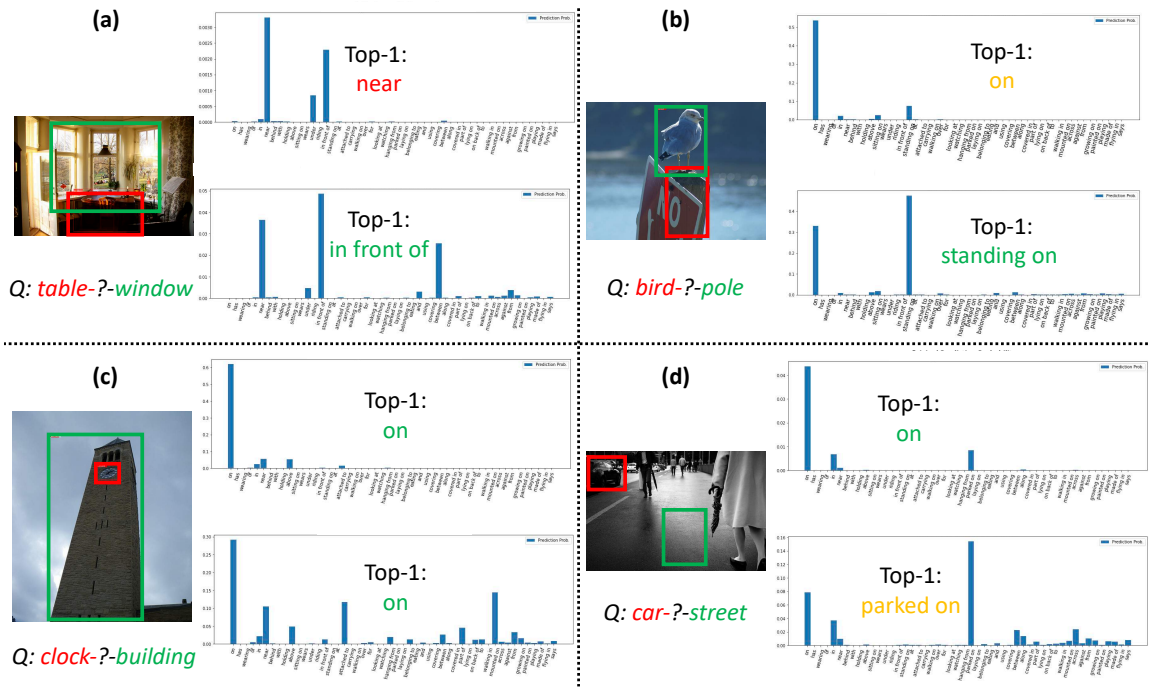

Figure 4.10: Class probability distributions (normalized to sum 1) by MOTIFS (top of each example) and MOTIFS-DLFE (bottom). The top-1 predictions can be correct (GT), incorrect (Non-GT and weird) or acceptable (Non-GT but reasonable).

We also refer the interested readers to Chapter 6.2 for a discussion on a discussion on the limitations and future directions of our proposed method.

## 4.6 Summary

In this chapter, we shift the focus from VRD to SGG to learn structured representations of images in a holistic manner. We deal with the long tail problem in SGG and we argue that the "bad" knowledge (*i.e.*, biased labeling mechanism) is the cause of long-tailed class distribution. To ward off the labeling bias caused by the imbalance in missing labels, we view SGG as a PU learning problem and we remove the per-class missing label bias by recovering the unbiased probabilities from the biased ones. To obtain reliable label frequencies for unbiased probability recovery, we take advantage of the data augmentation during training and perform Dynamic Label Frequency Estimation (DLFE) which maintains the moving averages of per-class biased probability and effectively introduces more valid samples, especially in SGDet training and evaluation mode. Extensive quantitative and qualitative experiments demonstrate that DLFE is more effective in estimating label frequencies than a naive variant of the traditional estimator, and SGG models with DLFE achieve state-of-the-art debiasing performance on the VG dataset, producing well-balanced scene graphs.

# Chapter 5

# Human-Object Interaction Detection in Videos

We have worked on learning structured representations of *static images* in Chapter 3 and 4. However, we note that in larger-scale real-word applications, *video* is not only are popular visual content format (*e.g.*, 500 hours of videos are being uploaded to YouTube every minute in 2020 [83]) but also (as a sequence of images) contains much more information than static images. Moreover, videos, rather than static images, are being utilized to record human-centered activities such as actions or interactions with objects or other humans. Many real-word applications in video understanding, such as pedestrian detection [33, 196, 105] and unmanned store systems [46, 104, 107], are *human-centered* tasks. This means we are more interested in detecting human-object interactions (HOIs) [141, 11, 39, 120, 92, 162, 159, 49, 91, 163, 97, 155, 199] instead of object-object (spatial) relations as we have done in VRD/SGG. Toward learning structured representations in real-word applications, we shift our focus to study how machines can perform human-object interaction detection in videos, or **Vid**eo **H**uman-**O**bject **I**nteraction **D**etection (VidHOID), in this chapter.

## 5.1  Challenges in VidHOID

We have discussed in Section 2.3 that three of the related works including *human-object interaction detection* (HOID), *video visual relation detection* (VidVRD) and *video human-object interaction retrieval* (VidHOIR) are not suitable for detecting HOIs in videos in a human-centered, spatial-temporal localization with atomic action labels. We further discuss the other two challenges we encountered when working

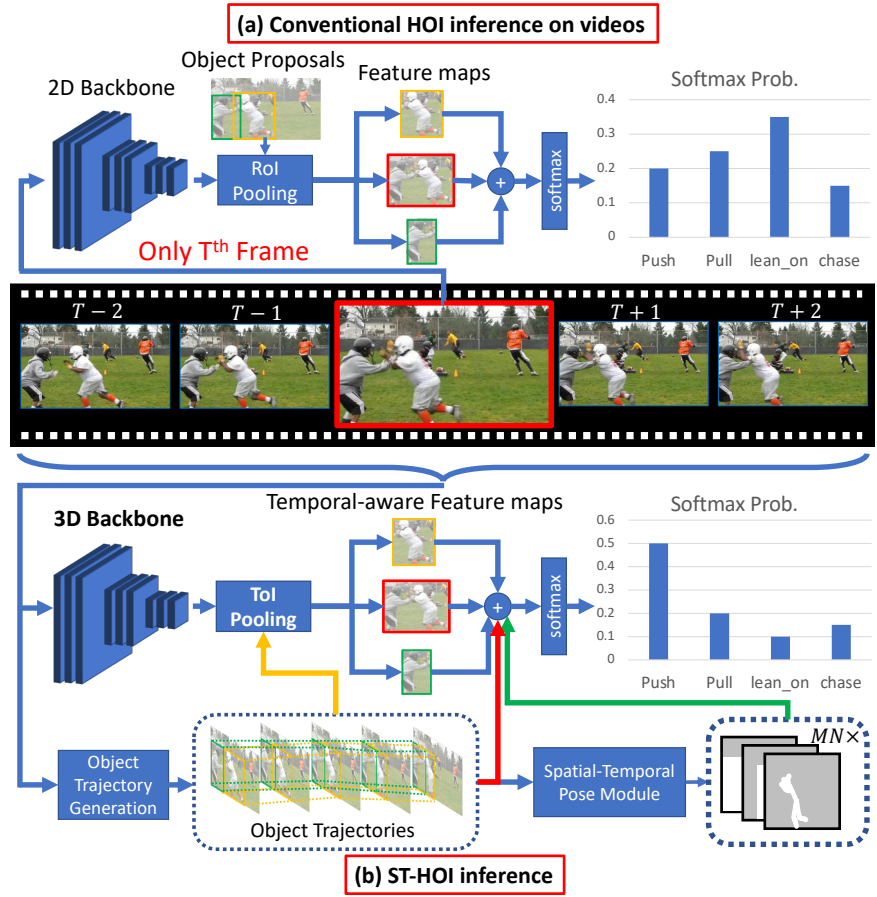

Figure 5.1: An illustrative comparison between conventional HOID methods and our ST-HOI when inferencing on videos. (a) Traditional HOID approaches (*e.g.*, the baseline in [159]) take in only the target frame and predict HOIs based on ROI-pooled visual features. These models are unable to differentiate between push, pull or lean on in this example due to the lack of temporal context. (b) ST-HOI takes in not only the target frame but neighboring frames and exploits temporal context based on trajectories. ST-HOI can thus differentiate temporal-related interactions and prefers push to other interactions in this example.

on VidHOID.

## 5.1.1 Lacking of a Large-scale Benchmark

Although there are abundant studies that have achieved success in detecting HOIs in static images, the fact that few of them [66, 120, 145] consider temporal information (*i.e.*, neighboring frames before/after the target frame) when performed on video data means they are actually "guessing" temporal-related HOIs with only naive co-occurrence statistics. While conventional image-based HOID methods

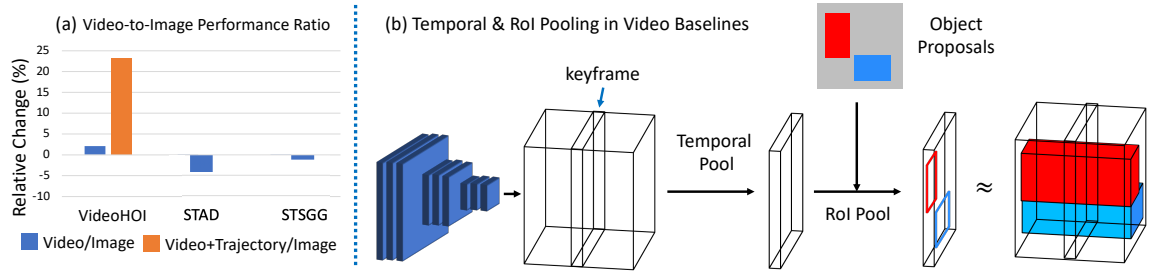

Figure 5.2: (a) Relative performance change (in percentage), on different video tasks by replacing 2D-CNN backbones with 3D ones (blue bars) [153, 10, 38], and on VidHOID by adding trajectory feature (tangerine bar). **VidHOID** (in triplet mAP) is to detect HOI in videos and was performed ourselves on our VidHOI benchmark. **STAD** [44] (in triplet mAP) means Spatial-Temporal Action Detection and was performed on AVA dataset [44]. **STSGG** [69] (PredCls mode; in Recall@20) stands for Spatial-Temporal Scene Graph Generation and was performed on Action Genome [69]. (b) An illustration of temporal-RoI pooling in 3D baselines (*e.g.* [38]). Temporal pooling is usually applied to the output of the penultimate layer of a 3D-CNN (shape of $d \times T \times H \times W$) which average-pools along the time axis into shape of $d \times 1 \times H \times W$, followed by RoI Pooling to obtain feature maps of shape $d \times 1 \times h \times w$. This temporal-RoI pooling, however, is equivalent to pooling the instance-of-interest feature at the same location in the keyframe throughout the video segment, which is erroneous for moving humans and objects.

(*e.g.*, the baseline model in [159]) can be used for inference on videos, they treat input frames as independent and identically distributed (*i.i.d.*) data and make independent predictions for neighboring frames. However, video data are sequential and structured by nature and thus are not *i.i.d.* What is worse is that, without temporal context these methods are unable to differentiate (especially, opposite) temporal interactions, such as `push` versus `pull` a human and `open` versus `close` a door. As shown in Figure 5.1(a), given a video segment, traditional HOID models operate on a single frame at a time and make predictions based on 2D-CNN (*e.g.*, [52]) visual features. These models by nature could not distinguish interactions between two people such as `push`, `pull`, `lean on` and `chase`, which are visually similar in static images. A possible reason causing video-based HOID underexplored is the lack of a suitable video-based benchmark and a feasible setting. To bridge this gap, we first construct a VidHOID benchmark from VidOR [131], dubbed **VidHOI**, where we follow the common protocol in video and HOID tasks to use a keyframe-centered strategy. With VidHOI, we urge the use of video data and propose **VidHOID** as – in both training and inference – performing HOI detection with videos.

### 5.1.2 A Feature Mismatch Problem

To tackle VidHOID, we draw inspiration from solutions to a related task, Spatial-Temporal Action Detection (STAD). STAD bears a resemblance to VidHOID by requiring to localize the human and detect the actions being performed in videos. Note that STAD does not consider the objects that a human is interacting with. STAD is usually tackled by first using a 3D-CNN [153, 10] as the backbone to encode temporal information into feature maps. This is followed by RoI pooling with object proposals to obtain actor features, which are then classified by linear layers. Essentially, this approach is similar to a common HOID baseline illustrated in Figure 5.1(a) and differs only in the use of 3D backbones and the absence of interacting objects.

Based on conventional image-based HOID and STAD methods, a naive yet intuitive idea arises: *can we enjoy the best of both worlds, by replacing 2D backbones with 3D ones and exploiting visual features of interacting objects?* This idea, however, did not work straightforwardly in our preliminary experiment, where we replaced the backbone in the 2D baseline [159] with the 3D one (*e.g.*, SlowFast [38]) to perform VidHOID. The relative change of performance after replacing the backbone is presented in the left most entry in Figure 5.2(a) with a blue bar. In VidHOID experiment, the 3D baseline provides only a limited relative improvement (∼2%), which is far from satisfactory considering the additional temporal context. In fact, this phenomenon has also been observed in two existing works under similar settings [44, 69], where both experiments in STAD and another video task Spatial-Temporal Scene Graph Generation (STSGG) present an even worse, counter-intuitive result: replacing the backbone is actually harmful (also presented as blue bars in Figure 5.2(a)). We probed the underlying reason by analyzing the architecture of these 3D baselines and found that, surprisingly, temporal pooling together with RoI pooling does not work reasonably. As illustrated in Figure 5.2(b), temporal pooling followed by RoI pooling, which is a common practice in conventional STAD methods, is equivalent to cropping features of the same region across the whole video segment without considering the way objects move. It is not unusual for moving humans and objects in neighboring frames to be absent from its location in the target keyframe. Temporal-and-RoI-pooling features at the same location could be getting erroneous features such as other humans/objects or meaningless background. Dealing with this inconsistency, we propose to recover the missing spatial-temporal information in VidHOID by considering human and object trajectories. The performance change of this temporal-augmented 3D baseline on VidHOID is represented by the tangerine bar in Figure 5.2(a), where it achieves ∼23% improvement, in sharp contrast to ∼2% of the original 3D baseline. This experiment reveals the importance of incorporating the "correctly-localized" temporal information.

Keeping the aforementioned ideas in mind, in this work we propose **S**patial-**T**emporal baseline for **H**uman-**O**bject **I**nteraction detection in videos, or **ST-HOI**, which makes accurate HOI prediction with instance-wise spatial-temporal features based on trajectories. As illustrated in Figure 5.1(b), three kinds of such features are exploited in ST-HOI: (a) trajectory features (moving bounding boxes; shown as the red arrow), (b) correctly-localized visual features (shown as the yellow arrow), and (c) spatial-temporal actor poses (shown as the green arrow).

The contribution of our work is three-fold. First, we are among the first to identify the feature inconsistency issue existing in the naive 3D models which we address with simple yet "correct" spatial-temporal feature pooling. Second, we propose a spatial-temporal model which utilizes correctly-localized visual features, per-frame box coordinates and a novel, temporal-aware masking pose module to effectively detect video-based HOIs. Third, we establish the keyframe-based VidHOI benchmark to motivate research in detecting spatial-temporal aware interactions and hopefully inspire VidHOID approaches utilizing the multi-modality data, *i.e.*, video frames, texts (semantic object/relation labels) and audios. This chapter including all of the texts, figures, tables, illustrations, equations is based on our published paper [19].

## 5.2 Related Work

As discussed in Section 2.3, three closely-related tasks including *human-object interaction detection* (HOID), *video visual relation detection* (VidVRD) and *video* *human-object interaction retrieval* (VidHOIR) are not suitable for detecting HOIs in videos in a human-centered, spatial-temporal localization with atomic action labels, and that is why we propose VidHOID in this work. We further discuss two other related video tasks, *spatial-temporal action detection* (STAD) and *spatial-temporal scene graph generation* (STSGG) in this section.

### 5.2.1 Spatial-Temporal Action Detection (STAD)

STAD aims to localize actors and detect the associated actions (without considering interacting objects). One of the most popular benchmark for STAD is AVA [44], where the annotation is done at a sampling frequency of 1 Hz and the performance is measured by framewise mean AP. We followed this annotation and evaluation style when constructing VidHOI, where we converted the original labels into the same format.

As explained in section 5.1, a standard approach to STAD [153, 10] is extracting spatial-temporal feature maps with a 3D-CNN followed by RoI pooling to crop human features, which are then classified by linear layers. As shown in Figure 5.2(a), a naive modification that incorporates RoI-pooled human/object features does not work for VidHOID. In contrast, our ST-HOI tackles VidHOID by incorporating multiple temporal features including trajectories, correctly-localized visual features and spatial-temporal masking pose features.

### 5.2.2 Spatial-Temporal Scene Graph Generation

Spatial-Temporal Scene Graph Generation (STSGG) [69] aims to generate symbolic graphs representing pairwise visual relationships in video frames. A new benchmark, Action Genome, is also proposed in [69] to facilitate researches in STSGG. Ji et al. [69] dealt with STSGG by combining off-the-shelf scene graph generation models with long-term feature bank [168] on top of a 2D- or 3D-CNN, where they found that the 3D-CNN actually undermines the performance. While observing similar results in VidHOI (Figure 5.2(a)), we go one step further to find out the underlying reason is that RoI features across frames were erroneously pooled. We correct this by utilizing object trajectories and applying Tube-of-Interest (ToI) pooling on generated trajectories to obtain correctly-localized position information

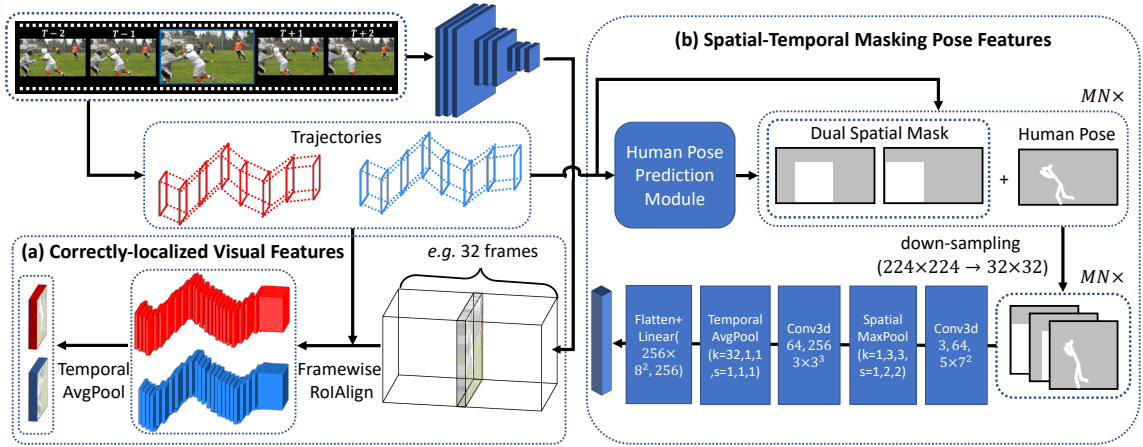

Figure 5.3: An illustration of the two proposed spatial-temporal features. (a) In contrast to performing RoI pooling followed by temporal pooling like [168, 38], we adopt a reverse approach to first frame-wise RoI-pool instance feature maps using trajectories, which are then averaged pool along the time axis to get correctly-localized visual features. (b) With $N$ object trajectories (including $M$ human), for each frame we utilize a trained human pose prediction model (*e.g.*, [37]) to generate 2D actor pose feature and extract a dual spatial mask for all $M \times (N-1)$ valid pair. The pose feature and the mask are concatenated and down-sampled, followed by two 3D convolution layers and spatial-temporal pooling to generate the masking pose features.

and feature maps throughout video segments.

## 5.3 Methodology

### 5.3.1 Overview

We follow STAD approaches [153, 10, 38] to detect VidHOI in a keyframe-centric strategy. Denote $V$ as a video which has $T$ keyframes with sampling frequency of 1 Hz as $\{I_t\}, t = \{1, ..., T\}$, and denote $C$ as the number of pre-defined interaction classes. Given $N$ instance trajectories including $M$ human trajectories ($M \leq N$) in a video segment centered at the target frame, for human $m \in \{1, ..., M\}$ and object $n \in \{1, ..., N\}$ in keyframe $I_t$, we aim to detect pairwise human-object interactions $r_t = \{0, 1\}^C$, where each entry $r_{t,c}, c \in \{1, ..., C\}$ means whether the interaction $c$ exists or not.

Refer to Figure 5.1(b) for an illustration of our ST-HOI. Our model takes in a video segment (sequence of $T$ frames) centered at $I_t$ and utilizes a 3D-CNN as the backbone to extract spatial-temporal feature maps of the whole segment. To rectify the mismatch caused by temporal-RoI pooling, based on $N$ object (including human) trajectories $\{j_i\}, i = \{1, .., N\}, j_i \in \mathbb{R}^{T \times 4}$ we generate temporal-aware features including correctly-localized features and spatial-temporal masking pose features. These features together with trajectories are concatenated and classified by linear layers. Note that we aim at a simple but effective temporal-aware baseline to VidHOID so that we do not utilize tricks in STAD such as non-local block [166] or long-term feature bank [168], and in image-based HOID like interactiveness [92], though we note that these may be used to boost the performance.

### 5.3.2 Correctly-localized Visual Features

We have discussed in previous sections on inappropriately pooled RoI features. We propose to tackle this issue by reversing the order of temporal pooling and RoI-pooling. This approach has recently been proposed in [60] and named as tube-of-interest pooling (ToIPool). Refer to Figure 5.3(a) for an illustration. Denote $v \in \mathbb{R}^{d \times T \times H \times W}$ as the output of the penultimate layer of our 3D-CNN backbone, and denote $v_t \in \mathbb{R}^{d \times H \times W}$ as the $t$-th feature map along the time axis. Recall that we have $N$ trajectories centered at a keyframe. Following the conventional way, we also exploit visual context when predicting an interaction, which is done by utilizing the union bounding box feature of a human and an object. For example, the sky between `human` and `kite` could help to infer the correct interaction `fly`. Recall that $j_i$ represents the trajectory of object $i$, where we further denote $j_{i,t}$ as the 2D bounding box at time $t$. The spatial-temporal instance features $\{\bar{v}_i\}$ are then obtained using ToIPool with RoIAlign [51] by

$$\bar{v}_i = \frac{1}{T} \sum_{t=1}^{T} \text{RoIAlign}(v_t, j_{i,t}), \qquad (5.1)$$

where $\bar{v}_i \in \mathbb{R}^{d \times h \times w}$ and $h$ and $w$ means height and width of the pooled feature maps, respectively. $\bar{v}_i$ is flattened before concatenating with other features.

### 5.3.3 Spatial-Temporal Masking Pose Features

Human poses have been widely utilized in image-based HOID methods [92, 49, 159] to exploit characteristic actor pose to infer some special actions. In addition, some existing works [162, 159] found that spatial information can be used to identify interactions. For instance, for `human-ride-horse` one can imagine the actor's skeleton as legs widely open (on horse sides), and the bounding box center of `human` is usually on top of that of `horse`. However, none of the existing works consider this mechanism in temporal domain: when riding a horse the `human` should be moving with `horse` as a whole. We argue that this temporality is an important property and should be utilized as well.

The spatial-temporal masking pose module is presented at Figure 5.3(b). Given $M$ human trajectories, we first generate $M$ spatial-temporal pose features with a trained human pose prediction model. On frame $t$, the predicted human pose $h_{i,t} \in \mathbb{R}^{17 \times 2}, i = \{1, .., M\}, t = \{1, .., T\}$ is defined as 17 joint points mapped to the original image. We transform $h_{i,t}$ into a skeleton on a binary mask

$$f_h : \{h_{i,t}\} \in \mathbb{R}^{17 \times 2} \to \{\bar{h}_{i,t}\} \in \mathbb{R}^{1 \times H \times W} \tag{5.2}$$

by connecting the joints using lines, where each line has a distinct value $x \in [0, 1]$. This helps the model to recognize and differentiate different poses.

For each of $M \times (N-1)$ valid human-object pairs on frame $t$, we also generate two spatial masks

$$s_{i,t} \in \mathbb{R}^{2 \times H \times W}, i = \{1, ..., M \times (N-1)\} \tag{5.3}$$

corresponding to human and object respectively, where the values inside of each bounding box are ones and outsides are zeroed-out. These masks enable our model to predict HOIs with reference to important spatial information.

For each pair, we concatenate the skeleton mask $\bar{h}_{i,t}$ and spatial masks $s_{i,t}$ along the first dimension to get the initial spatial masking pose feature $p_{i,t} \in \mathbb{R}^{3 \times H \times W}$:

$$p_{i,t} = [s_{i,t}; \bar{h}_{i,t}]. \tag{5.4}$$

We then down-sample $\{p_{i,t}\}$, feed into two 3D convolutional layers with spatial and temporal pooling, and flatten to obtain the final spatial-temporal masking pose feature $\{\bar{p}_{i,t}\}$.

Table 5.1: A comparison of our benchmark VidHOI with existing STAD (AVA [44]), image-based (HICO-DET [11] and V-COCO [48]) and video-based (CAD-120 [79] and Action Genome [69]) HOID datasets. VidHOI is the only dataset that provides temporal information from video clips and complete multi-person and interacting-object annotations. VidHOI also provides the most annotated keyframes and defines the most HOI categories in the existing video datasets. †Two less categories as we combine `adult`, `child` and `baby` into a single category, `person`. *cats.*: categories, *insts.*: instances, *obj.*: objects.

| Dataset | Video dataset? | Localized objs.? | Video hours | #Videos | #Labeled images | #Obj. cats. | #Pred. cats. | #HOI cats. | #HOI insts. |
|---|---|---|---|---|---|---|---|---|---|
| HICO-DET [11] | ✗ | ✓ | - | - | 47K | 80 | 117 | 600 | 150K |
| V-COCO [48] | ✗ | ✓ | - | - | 10K | 80 | 25 | 259 | 16K |
| AVA [44] | ✓ | ✗ | 108 | 437 | 3.7M | - | 49 | 80 | 1.6M |
| CAD-120 [79] | ✓ | ✓ | 0.57 | 0.5K | 61K | 13 | 6 | 10 | 32K |
| Action Genome [69] | ✓ | △ | 82 | 10K | 234K | 35 | 25 | 157 | 1.7M |
| **VidHOI** | ✓ | ✓ | 70 | 7122 | **7.3M** | 78† | 50 | **557** | 755K |

### 5.3.4 Prediction

We fuse the aforementioned features, including correctly-localized visual features $\bar{v}$, spatial-temporal masking pose features $p$, and instance trajectories $j$ by concatenating them along the last axis

$$v_{\text{so}} = [\bar{v}_s; \bar{v}_u; \bar{v}_o; j_s; j_o; \bar{p}_{so}], \quad (5.5)$$

where we slightly abuse the notation to denote the subscriptions $s$ as the subject, $o$ as the object and $u$ as their union region. $v_{\text{so}}$ is then fed into two linear layers with the final output size being the number of interaction classes in the dataset. Since VidHOID is essentially a multi-label learning task, we train the model with per-class binary cross entropy loss.

During inference, we follow the heuristics in image-based HOID [11] to sort all the possible pairs by their softmax scores and evaluate on only top 100 predictions.

## 5.4 Experiments

### 5.4.1 VidHOI Benchmark

#### 5.4.1.1 Dataset and Performance Metric

Before diving into our proposed benchmark, we first explain why the recently-proposed Action Genome [69] is also not a feasible choice. First, the authors acknowledged that the dataset is still incomplete and contains incorrect labels [68]. Second, Action Genome is produced by annotating Charades [137], which is originally designed for activity classification where each clip contains only one "actor" performing predefined tasks; should any other people show up, there are neither any bounding box nor interaction label about them. Finally, the videos are purposedly-generated by volunteers, which are rather unnatural. In contrast, VidHOI are based on VidOR [131] which is densely annotated with all humans and predefined objects showing up in each frame. VidOR is also more challenging as the videos are non-volunteering user-generated and thus jittery at times. A comparison of VidHOI and the existing STAD and HOID datasets is presented in Table 5.1.

VidOR is originally collected for video visual relationship detection where the evaluation is trajectory-based. The volumetric Interaction Over Union (vIOU) between a trajectory and a ground truth needs to be over 0.5 before considering its relationship prediction; however, how to obtain accurate trajectories with correct start- and end-timestamp remains challenging [144, 133]. We notice that some image-based HOID datasets (*e.g.*, HICO-DET [11] and V-COCO [48]) as well as STAD datasets (*e.g.*, AVA [44]) are using a keyframe-centered evaluation strategy, which bypasses the aforementioned issue. We thus adopt the same and follow AVA to sample keyframes at a 1 FPS frequency, where the annotations on the keyframe at timestamp $t$ are assumed to be fixed for $t \pm 0.5$s. In detail, we first filter out those keyframes without presenting at least one valid human-object pair, followed by transforming the labels from video clip-based to keyframe-based to align with common HOID metrics (*i.e.*, frame mAP). We follow the original VidOR split in [131] to divide VidHOI into a training set comprising 193,911 keyframes in 6,366 videos and a validation set[1] with 22,808 keyframes in 756 videos. As shown in Figure 5.4, there are 50 relation classes including actions (*e.g.*, `push`, `pull`, `lift`, etc.) and spatial relations (*e.g.*, `next to`, `behind`, etc.). While half (25) of the predicate classes are temporal-related, they account for merely ∼5% of the dataset.

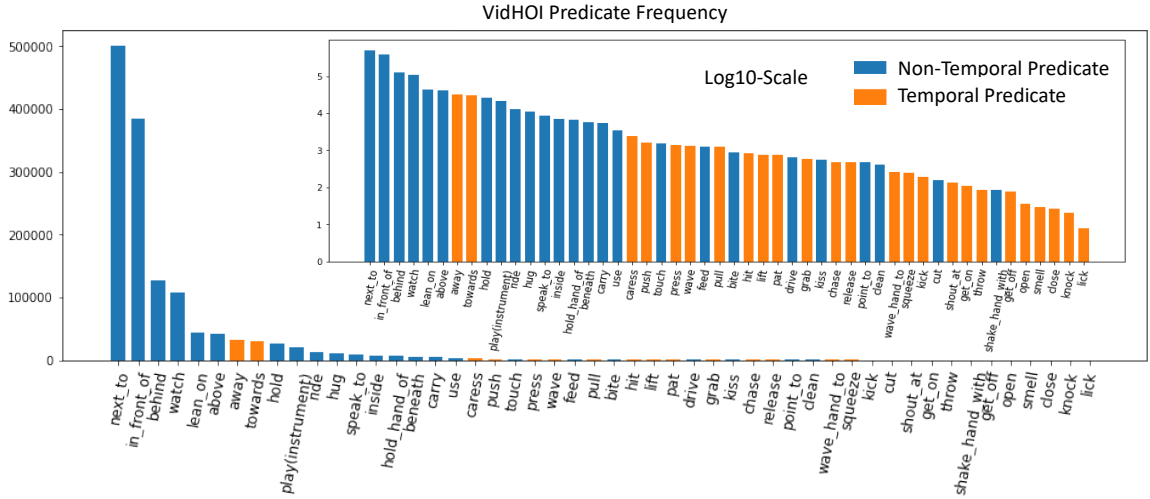

Figure 5.4: Predicate distribution of the VidHOI benchmark shows that most of the predicates are non-temporal-related.

Following the evaluation metric in HICO-DET, we adopt mean Average Precision (mAP), where a true positive HOI needs to meet three below criteria: (a) both the predicted human and object bounding boxes have to overlap with the ground truth boxes with IOU over 0.5, (b) the predicted target category need to be matched and (c) the predicted interaction is correct. Over 50 predicates, we follow HICO-DET to define HOI categories as 557 triplets on which we compute mean AP. By defining HOI categories with triplets we can bypass the polysemy problem [199], *i.e.*, the same predicate word can represent very different meaning when pairing with distinct objects, *e.g.*, `person-fly-kite` and `person-fly-airplane`. We report the mean AP over three categories: (a) **Full**: all 557 categories are evaluated, (b) **Rare**: 315 categories with less than 25 instances in the dataset, and (c) **Non-rare**: 242 categories with more than or equal to 25 instances in the dataset. We also examine the models in two evaluation modes: `Oracle` models are trained and tested with ground truth trajectories, while models in `Detection` mode are tested with predicted trajectories.

---

[1]The VidOR testing set is not available publicly.

### 5.4.1.2 Implementation Details

We adopt ResNet-50 [52] as our 2D backbone for the preliminary experiments, and utilize ResNet-50-based SlowFast [38] as our 3D backbone for all the other experiments. SlowFast contains the Slow and Fast pathways, which correspond to the texture details and the temporal information, respectively, by sampling video frames in different frequencies. For a 64-frame segment centered at the keyframe, $T = 32$ frames are alternately sampled to feed into the Slow pathway; only $T/\alpha$ frames are fed into the Fast pathway, where $\alpha = 8$ in our experiments. We use FastPose [37] to predict human poses and adopt the predicted trajectories generated by a cascaded model of video object detection, temporal NMS and tracking algorithm [144]. Like object detection is to 2D HOI detection, trajectory generation is an essential module but not a main focus of this work. If a bounding box is not available in neighboring frames (*i.e.*, the trajectory is shorter than $T$ or not continuous throughout the segment), we fill it with the whole-image as a box. We train all models from scratch for 20 epochs with the initial learning rate $1 \times 10^{-2}$, where we use step decay learning rate to reduce the learning rate by $10\times$ at the $10^{\text{th}}$ and $15^{\text{th}}$ epoch. We optimize our models using synchronized SGD with momentum of 0.9 and weight decay of $10^{-7}$. We train each 3D video model with eight NVIDIA Tesla V100 GPUs with batch size being 128 (*i.e.*, 16 examples per GPU), except for the full model where we set batch size as 112 due to the memory restriction. We train the 2D model with a single V100 with batch size being 128.

During training, following the strategy in SlowFast we randomly scale the shorter side of the video to a value in [256, 320] pixels, followed by random horizontal flipping and random cropping into $224 \times 224$ pixels. During inference, we only resize the shorter side of the video segment to 224 pixels.

### 5.4.1.3 Full Results

Since we aim to deal with a) the lack of temporal-aware features in 2D HOID methods, b) the feature inconsistency issue in common 3D HOID methods and c) the lack of a VidHOID benchmark, we mainly compare with the 2D model [159] and its naive 3D variant on VidHOI to understand if our ST-HOI addresses these issues effectively. We also compare with two existing approaches: VidVRD-MMF [144] and SeRVo-HOI [47]. VidVRD-MMF was originally designed for a closely related tasks, VidVRD [133], which is a clip-level visual relationship detection task (as we discussed in Section 2.3). VidVRD-MMF is a multi-modal feature fusion model which comprises three stages: i) object trajectory detection, ii) relation instance prediction and iii) segment association. Note that the only stage we can compare is the second one (*i.e.*, relation instance prediction), since we use the same trajectory generation module as in our baselines, and the segment association stage does not apply to VidHOID. On the other hand, Gupta et al. [47] propose SeRVo-HOI which is a VidHOID pipeline with a similar architecture to our ST-HOI but it differs in that it fuses spatial-temporal information with 3D-CNNs at the end of the network (without temporal pooling). In addition, they employ a stronger visual backbone, *i.e.*, ResNext-101 [171] and they explore different choices of loss functions (*e.g.*, propensity-weighted loss [67], focal loss [100], etc.) to address the long-tailed problem.

Table 5.2: Results of the baselines, our full model (ST-HOI) and two models adapted from VidVRD and VidHOIR, respectively, on the VidHOI validation set (numbers in mAP). There are two evaluation modes: Detection and Oracle, which differ only in the use of predicted or ground truth trajectories during inference. "%" means the full mAP change compared to the *2D model*.

|  | Model | Full | Non-rare | Rare | % |
|---|---|---|---|---|---|
| *Oracle* | 2D model [159] | 14.1 | 22.9 | 11.3 | - |
|  | 3D model | 14.4 | 23.0 | 12.6 | 2.1 |
|  | Ours: ST-HOI | **17.6** | **27.2** | **17.3** | **24.8** |
|  | LIGHTEN [145, 47] | 13.4 | - | - | −5.0 |
|  | VidVRD-MMF [144] | 14.1 | 21.9 | 11.6 | 0.0 |
|  | SeRVo-HOI (vanilla) [47] | 19.5 | 28.8 | 19.5 | 38.3 |
|  | SeRVo-HOI (best) [47] | **21.2** | **29.7** | **20.9** | **50.4** |
| *Detection* | 2D model [159] | 2.6 | 4.7 | 1.7 | - |
|  | 3D model | 2.6 | 4.9 | 1.9 | 0.0 |
|  | Ours: ST-HOI | **3.1** | **5.9** | **2.1** | **19.2** |
|  | VidVRD-MMF [144] | 2.6 | 4.4 | 1.7 | 0.0 |
|  | SeRVo-HOI (vanilla) [47] | 4.0 | 6.4 | 3.2 | 53.8 |
|  | SeRVo-HOI (best) [47] | **4.8** | **6.9** | **4.1** | **84.6** |

The performance comparison between our full ST-HOI model and baselines (**2D model**, **3D model**) are presented in Table 5.2. Table 5.2 shows that **3D model** only has a marginal improvement compared to **2D model** (overall ∼2%) under all settings in both evaluation modes. On the other hand, our ST-HOI achieve significant performance boost compared to both the 2D and 3D models, *e.g.*, leading to a relative 25% higher mAP than 2D model. Overall, the performance gap between `Detection` and `Oracle` models is significant, indicating the room for improvement in trajectory generation. Another interesting observation is that the *Full mAPs* are very close to the *Rare mAPs*, especially under `Oracle` mode, showing that the long-tailed effect over HOIs is strong (but is common and natural).

Comparing to other methods, the table shows that the multi-modal feature fusion model, VidVRD-MMF, has similar performance to our 2D/3D baseline models. This inferior result compared to ours is presumably because of their choice of features: they omit visual features which has been shown important in broadly vision tasks. In contrast, SERVO-HOI [47] achieves state-of-the-art results on our benchmark thanks to their stronger backbone and debiasing loss functions.

### 5.4.1.4 Ablation Study: Feature Choices

We also present ablation studies on our different features (modules) including trajectory features (**T**), correctly-localized visual features (**V**) and spatial-temporal masking pose features (**P**). As shown in Table 5.3, adding trajectory features (*i.e.*, **Ours-T**) leads to a much larger 23% improvement in `Oracle` mode or 15% in `Detection` mode, showing the importance of correct spatial-temporal information. We also find that by adding additional temporal-aware features (*i.e.,* **V** and **P**) to **Ours-T**, increasingly higher mAPs are attained. Our full model (**Ours-T+V+P**) reports the best mAPs in `Oracle` mode, achieving the highest ∼25% relative improvement. We notice that the performance of **Ours-T+V** is close to that of **Ours-T** under `Oracle` setting, which is possibly because the ground truth trajectories (**T**) have provided enough "correctly-localized" information so that the correct features do not help much. We also note that the performance of **Ours-T+P** is slightly higher than that of **Ours-T+V+P** under `Detection` mode, which is assumably due to the same, aforementioned reason and the inferior performance resulting from the predicted trajectories. Notably, the reported numbers in the `Oracle` mode represent an upper bound performance of the proposed methods. The fact that the results under the `Oracle` mode are significantly better than those under the `Detection` mode (*i.e.*, 17.6 vs 3.2) leaves plenty of room for improvement of trajectory detection models.

Table 5.3: Ablation study on spatial and temporal feature choices on VidHOI validation set (numbers in mAP). **T**: Trajectory features. **V**: Correctly-localized visual features. **P**: Spatial-temporal masking pose features. "%" means the full mAP change compared to the *2D model*. \T means no trajectory is used (same as the *3D model*).

|  | Model | Full | Non-rare | Rare | % |
|---|---|---|---|---|---|
| *Oracle* | Ours-\T | 14.4 | 23.0 | 12.6 | 2.1 |
| | Ours-T | 17.3 | 26.9 | 16.8 | 22.7 |
| | Ours-T+V | 17.3 | 26.9 | 16.3 | 22.7 |
| | Ours-T+P | 17.4 | 27.1 | 16.4 | 23.4 |
| | Ours-T+V+P | **17.6** | **27.2** | **17.3** | **24.8** |
| *Detection* | Ours-\T | 2.6 | 4.9 | 1.9 | 0.0 |
| | Ours-T | 3.0 | 5.5 | 2.0 | 15.4 |
| | Ours-T+V | 3.1 | 5.8 | 2.0 | 19.2 |
| | Ours-T+P | **3.2** | **6.1** | 2.0 | **23.1** |
| | Ours-T+V+P | 3.1 | 5.9 | **2.1** | 19.2 |

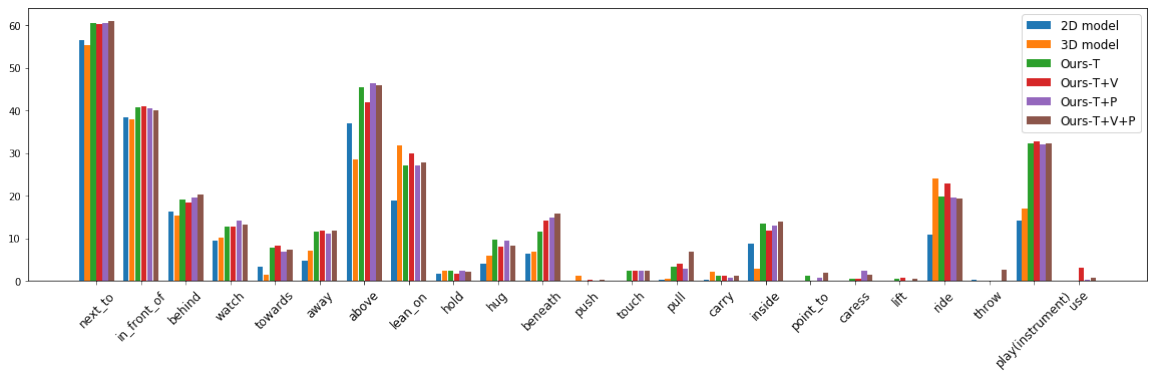

Figure 5.5: Performance comparison in predicate-wise AP (pAP). The performance boost after adding trajectory features is observed for most of the predicates. Interestingly, both spatial (*e.g.*, `next to`, `behind`) and temporal (*e.g.*, `towards`, `away`) predicates benefit from the temporal-aware features. Predicates sorted by the number of occurrence. Models in `Oracle` mode.

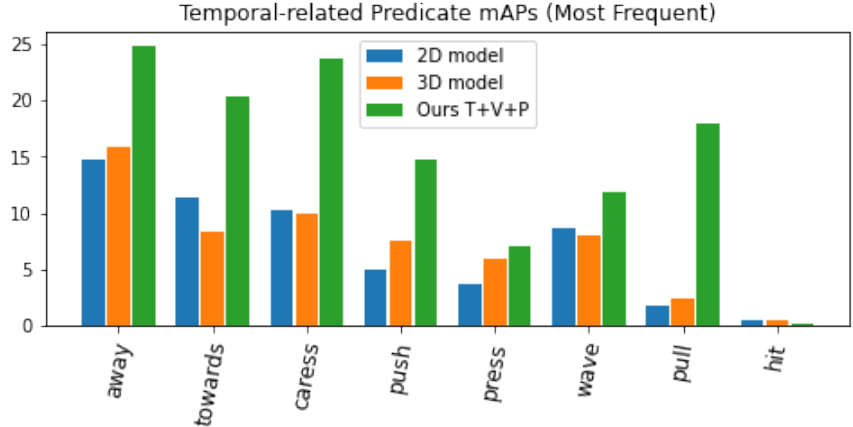

Figure 5.6: Results (in predicate-wise AP) of the baselines and our full model w.r.t. top frequent temporal predicates.

To understand the effect of temporal features on individual predicates, we compare with predicate-wise AP (pAP) shown in Figure 5.5. We observe that, again, under most of circumstances naively replacing 2D backbones with 3D ones does not help video HOI detection. Both temporal predicates (*e.g.*, `towards`, `away`, `pull`) and spatial (*e.g.*, `next_to`, `behind`, `beneath`) predicates benefit from the additional temporal-aware features in ST-HOI. These findings verify our main idea about the essential use of trajectories and trajectory-based features. In addition, each additional features do not seem to contribute equally for different predicates. For instance, we see that while **Ours-T+V+P** performs the best on some predicates (*e.g.*, `behind` and `beneath`), our sub-models achieve the highest mAP on other predicates (*e.g.*, `watch` and `ride`). This is assumedly because predicate-wise performance is heavily subject to the number of examples, where major predicates have 10-10000 times more examples than minor ones (as shown in Figure 5.4).

### 5.4.1.5 Analysis: Spatial versus Temporal HOIs

Since the majority of HOI examples are spatial-related (∼95%, as shown in Figure 5.4), the results above might not be suitable for demonstrating the temporal modeling ability of our proposed model. We thus focus on the performance on only temporal-related predicates in Figure 5.6, which demonstrates that **Ours-T+V+P** greatly outperforms the baselines regarding the top frequent temporal predicates.

Table 5.4: Results of temporal-related and spatial (non-temporal) related triplet mAP. T%/S% means relative temporal/spatial mAP change compared to 2D model [159]. **T**: Trajectory features. **V**: Correctly-localized visual features. **P**: Spatial-temporal masking pose features. \T means no trajectory is used (same as the *3D model*).

|  |  | Temporal | T% | Spatial | S% |
|---|---|---|---|---|---|
| *Oracle* | 2D model [159] | 8.3 | - | 18.6 | - |
|  | Ours-\T | 7.7 | -7.2 | 20.9 | 12.3 |
|  | Ours-T | **14.4** | **73.5** | 24.7 | 32.8 |
|  | Ours-T+V | 13.6 | 63.9 | 24.6 | 32.3 |
|  | Ours-T+P | 12.9 | 55.4 | **25.0** | **34.4** |
|  | Ours-T+V+P | **14.4** | **73.5** | **25.0** | **34.4** |
| *Detection* | 2D model [159] | 1.5 | - | 2.7 | - |
|  | Ours-\T | 1.6 | 6.7 | 2.9 | 7.4 |
|  | Ours-T | 1.8 | 20.0 | **3.3** | **23.6** |
|  | Ours-T+V | 1.8 | 20.0 | **3.3** | **23.6** |
|  | Ours-T+P | 1.8 | 20.0 | **3.3** | **23.6** |
|  | Ours-T+V+P | **1.9** | **26.7** | **3.3** | **23.6** |

Table 5.4 presents the triplet mAPs of spatial- or temporal-only predicates, showing **Ours-T** significantly improves the **2D model** on temporal-only mAP by relative +73.9%, in sharp contrast to -7.1% of the **Ours-\T** (**3D model**) in `Oracle` mode. Similar to our observation with Table 5.2, **Ours-T** performs on par with **Ours-T+V+P** for temporal-only predicates; however, it falls short of spatial-only predicates, showing that spatial/pose information is still essential for detecting spatial predicates. Overall, these results demonstrate the outstanding spatial-temporal modeling ability of our approach.

We also compare the performance with respect to some HOI triplets in Figure 5.7. Similar to the results on predicate-wise mAP, we also observe the large gap between naive 2D/3D models and our models with the temporal features. ST-HOI variants are more accurate in predicting especially temporal-aware HOIs (`hug/lean_on-person` and `push/pull-baby_walker`). We also see in some examples that **Ours-T+V+P** does not perform the best among all the variants, *e.g.*, `lean_on-person`), which is similar to the phenomenon we observed in Figure 5.5.

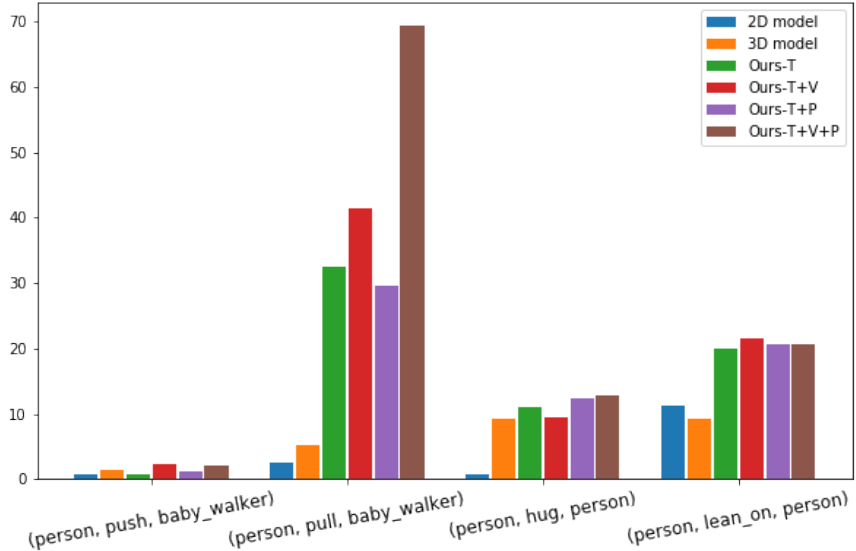

Figure 5.7: Performance comparison (in AP) of some temporal-related HOIs in VidHOI validation set. Compared to 2D model, 3D model (Ours-\T) only shows limited improvement for the presented examples, while our ST-HOI variants provide huge performance boost. Models are in `Oracle` mode.

#### 5.4.1.6 Qualitative Results

To understand the effectiveness of our proposed method, we visualize two video HOI examples of VidHOI predicted by the **2D model** [159] and **ST-HOI** (both in `Oracle` mode) in Figure 5.8. Each (upper and lower) example is a 5-second video segment (*i.e.*, five keyframes) with a HOI prediction table where each entry means either True Positive (TP), False Positive (FP), False Negative (FN) or True Negative (TN) for both models. The upper example shows that, compared to the **2D model**, **ST-HOI** makes more accurate HOI detection by successfully predicting `hold_hand_of` at $T_4$ and $T_5$. Moreover, **ST-HOI** is able to predict interactions that requires temporal information, such as `lift` at $T_1$ in the lower example. However, we can see that there is still room for improvement for **ST-HOI** in the same example, where `lift` is not detected in the following $T_2$ to $T_4$ frames. Overall, our model produces less false positives throughout the dataset, which in turn contributes to its higher mAP and pAP.

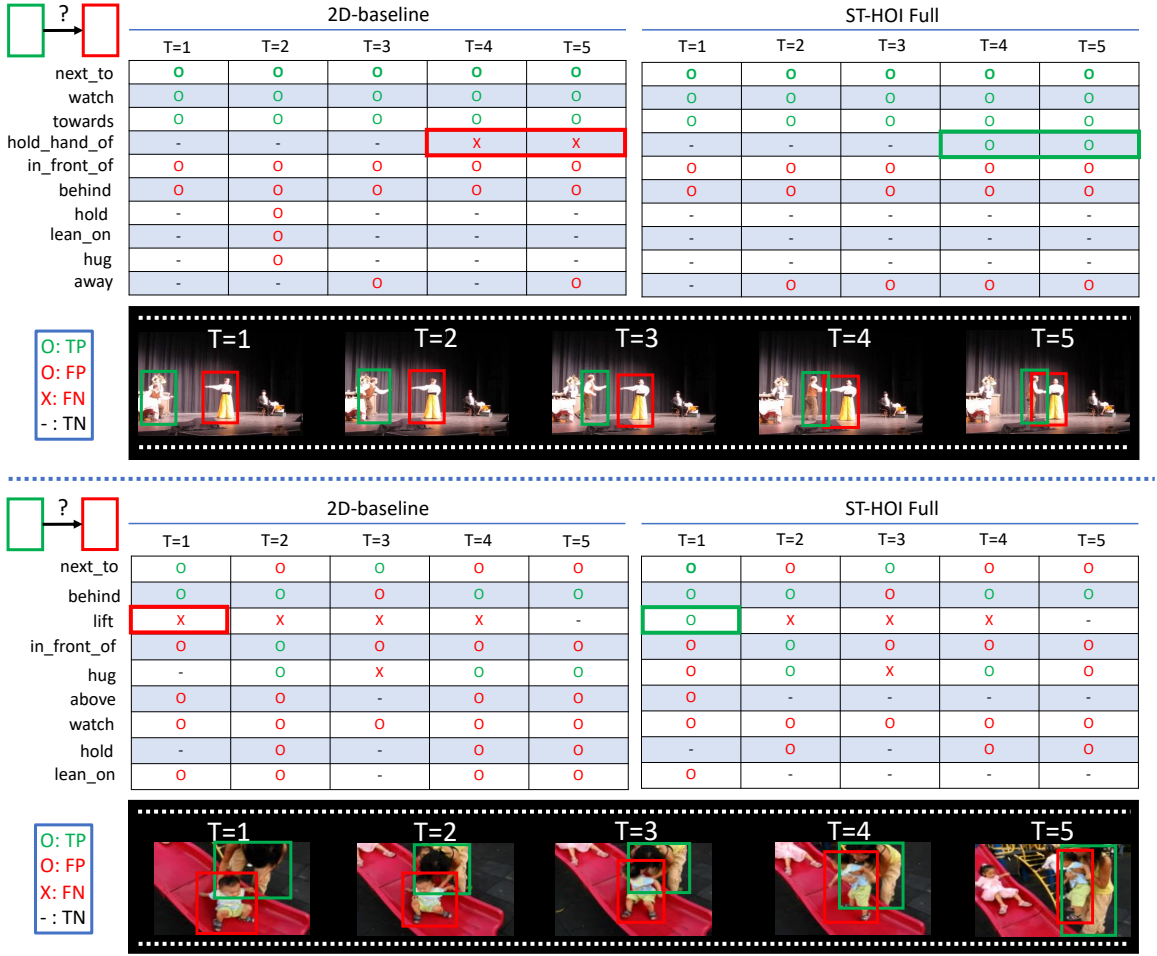

Figure 5.8: Examples of video HOIs predicted by the 2D model [159] and our ST-HOI, both in `Oracle` mode. Each consists of five consecutive keyframes sampled in 1 Hz, where an entry in tables denotes whether a predicate between the subject (human; a green box) and the object (also human in both cases; a red box) is detected correctly (True Positive) or not (False Positive or False Negative). Compared to the 2D baseline, our model predicts more accurate temporal HOIs (*e.g.*, `hold_hand_of` in $T_4$ and $T_5$ of the upper example and `lift` in $T_1$ of the lower example). ST-HOI also produces less false positives in both examples.

### 5.4.2 ImageNet-VidVRD Benchmark

#### 5.4.2.1 Dataset and Performance Metric

Besides our proposed VidHOI benchmark, we also experiment the proposed baseline model on a video relationship detection benchmark, ImageNet-VidVRD [133]. Note that ImageNet-VidVRD does not focus on human-centric interactions, but on all kinds of relationships. ImageNet-VidVRD curated 1,000 videos from ILSVRC2016-VID [126], where we follow [133, 154, 103, 14] to use 800 of them for training and the other 200 for evaluation. There are 35 object classes and 132 predicate types annotated, along with the objects' trajectories, in the dataset.

Following [133, 121, 154, 103, 14], we evaluate our model with two tasks: relation detection and relation tagging.

- **Relation detection** can be viewed as an extension of the Scene Graph Detection (SGDet) task of SGG into video domain, in that it aims to generate not only visual relationship triplets of (`subject, predicate, object`) but also trajectories of subjects and objects, where both the trajectories have to adequately overlap the ground truth ones by having volumetric Intersection over Union (vIoU) greater than or equal to 0.5 [133]. We use Recall@K ($K \in \{50, 100\}$) which computes the fraction of correct predictions among top $k$ confident predictions, and mean Average Precision (mAP) which measures the overall performance.

- **Relation tagging** is a relatively easier task which only requires a video VRD model to outputs a set of visual relation triplets without localizing object trajectories. Since the video-level visual relations here are more likely to be intensively annotated than frame-level VRD datasets (*e.g.*, VG dataset [80]), precision@K as a fairer and less passive metric can be used here.

#### 5.4.2.2 Implementation Details

Since ImageNet-VidVRD consists of not only humans, we leave the masking pose features (Section 5.3.3) out. We adopt a three-stage VidVRD pipeline [103] including trajectory proposal, relationship pair proposal and relationship classification, where our ST-HOI's spatial-temporal features are taken in by the relationship classifier (which is MLPs). For fairer comparison, we follow [151, 103, 14] to utilize I3D [10] instead of SlowFast (while the backbone is still ResNet-50) to extract 3D convolutional visual features, and we incorporate the additional visual features from the object detector (Faster R-CNN [123]) used in the trajectory proposal stage.

Table 5.5: Comparison of video visual relationship detection performance on ImageNet-VidVRD dataset [133].

| Method | Relation Detection | | | Relation Tagging | | |
|---|---|---|---|---|---|---|
| | R@50 | R@100 | mAP | P@1 | P@5 | P@10 |
| VidVRD [133] | 5.5 | 6.4 | 8.6 | 43.0 | 28.9 | 20.8 |
| GSTEG [154] | 7.1 | 8.7 | 9.5 | 51.5 | 39.5 | 28.2 |
| VRD-GCN [121] | 7.4 | 8.8 | 14.2 | 59.5 | 40.5 | 27.9 |
| VRD-GCN+siamese [121] | 8.1 | 9.3 | 16.3 | 57.5 | 41.0 | 28.5 |
| TRACE [151] | 9.1 | 11.2 | 17.6 | 61.0 | 45.3 | 33.5 |
| VidVRD-MMF [144] | 9.5 | 10.4 | 19.0 | 57.5 | 41.4 | 29.5 |
| BSTS [103] | 11.2 | 13.7 | 18.4 | 60.0 | 43.1 | 32.2 |
| Social Fabric [14] | **13.7** | **16.9** | **20.1** | **62.5** | **49.2** | **38.5** |
| ST-HOI (Ours-T+V) | 9.1 | 11.4 | 14.8 | 55.5 | 38.9 | 28.9 |

#### 5.4.2.3 Results

We compare our proposed baseline with state-of-the-art video VRD methods including VidVRD [133], GSTEG [154], VRD-GCN [121] (and its variant), TRACE [151], BSTS [103], VidVRD-MMF [144] and Social Fabric [14]. As shown in Table 5.5, our approach, ST-HOI (Ours-T+V), achieves decent performance on all the metrics compare to the other alternatives. We note that it might be hard to fairly compare our methods with video VRD methods, since our method is not designed for optimizing video-level relation detection and only serve as a classification head, but not for trajectory proposal and relation instance association. Moreover, different combinations of features including visual features, language clues, relative motion features and 3D Conv features are being arbitrarily used in these works [151, 14, 121, 144], making a fair comparison between results difficult. However, as also found in [14], additional spatial-temporal features are beneficial to the model performance.

## 5.5 Summary

In this chapter, we study how machines can learn *human-centered*, structured representations in videos, which are potentially useful in real-word applications such as unmanned store systems and pedestrian detection. Because of the lack of an idea VidHOID setting (as discussed in Section 2.3), we propose a new task named VidHOID which aims to detect HOIs in videos in a keyframe-centered, spatial-temporal-localized manner. Specifically, we established a new VidHOID benchmark dubbed VidHOI and introduced a keyframe-centered detection evaluation strategy. We then proposed a spatial-temporal baseline ST-HOI which exploits trajectory-based temporal features including correctly-localized visual features, spatial-temporal masking pose features and trajectory features, solving the second problem. With quantitative and qualitative experiments on VidHOI, we showed that our model provides a huge performance boost compared to both the 2D and 3D baselines and is effective in differentiating temporal-related interactions. Our proposed relation classifier is also shown to achieve decent results on a VidVRD benchmark. We expect that the proposed baseline and the dataset would serve as a solid starting point for the relatively underexplored VidHOID task.

# Chapter 6

# Conclusions

## 6.1 Summary

In this thesis, we developed novel approaches to learning structured representations of visual scenes. We first demonstrated that rich, external visual-linguistic knowledge along with our novel visual and spatial feature modules are beneficial for learning stronger visual relationship predictors (Chapter 3). We then turned our focus to scene graph generation for more holistic, structured representation, and we showed how unbiased scene graphs can be recovered under a more proper positive-unlabeled learning setting, which we tackled with a performant positive-unlabeled learning algorithm tailored for scene graph generation, achieving new state-of-the-art debiasing performance (Chapter 4). Furthermore, extending VRD and SGG into the temporal domain where interactions between humans and objects are of interest, we proposed a new benchmark where our spatial-temporal human-object interaction detection model serves as a solid baseline (Chapter 5).

Overall, these findings facilitate the learning of structured scene representations from different aspects: performance, informativeness and spatial-temporal perception. We summarize the contributions of this thesis as follows:

- **Visual relation detection with external multimodal knowledge.** We introduced a novel Transformer-based multi-modal VRD architecture named *Relational Visual-Linguistic Bidirectional Encoder Representation from Transformers* (RVL-BERT) which is enriched by the visual-linguistic knowledge from large-scale external datasets. Equipped with attention-guided visual feature module and spatial feature module, RVL-BERT achieves state-of-the-art VRD performance on the SpatialSense dataset and competitive results on the VRD and VG datasets. We also demonstrated, quantitatively and qualitatively, the effectiveness of the external knowledge and the proposed attention-guided visual module (Chapter 3).

- **Unbiased scene graph generation.** We proposed to view SGG as a positive-unlabeled (PU) learning problem, on which the labeling biases (including reporting bias and bounded rationality) can be concretely taken into account and then removed. We introduced a model-agnostic PU learning algorithm dubbed *Dynamic Label Frequency Estimation* (DLFE) which dynamically estimates the labeling probability values that are indispensable for offsetting the biases. Applying DLFE to existing SGG methods has shown to result in state-of-the-art unbiased SGG performance, specifically, +5 averaged mean recall points (24% $\rightarrow$ 29%) or +21 tail recall points (17% $\rightarrow$ 38%) than the previous state-of-the-art on the challenging VG dataset. DLFE is also shown to contribute to significantly more informative scene graphs (Chapter 4).

- **Human-object interaction detection in videos.** We studied how to construct structured human-center scene representations of videos (*i.e.*, *video human-object interaction detection* (VidHOID)), which has still remained underexplored. Specifically, we introduced a keyframe-centered, larger-scale VidHOID benchmark named *VidHOI* enabling more accurate performance analysis. We then proposed a strong VidHOID baseline called *Spatial-Temporal Human-Object Interaction* (ST-HOI) which exploits box trajectories along with frame-wise human and object features. ST-HOI outperform the 2D/3D baseline models by obtaining 74% relatively or 6.1% absolutely higher mAP (8.3% $\rightarrow$ 14.4%) for temporal-aware HOIs, and 34% relatively or 6.4% absolutely higher overall mAP (18.6% $\rightarrow$ 25.0%) for all kinds of HOIs.

## 6.2 Open Challenges and Future Work

In this thesis, we put an emphasis on some critical challenges of constructing better structured representations including the performance and informativeness of SGG/VRD and the ability to take the temporal domain into account. Nevertheless, some open challenges and limitations still remain in our proposed approaches. We foresee that the remaining issues might be great directions for future work.

- **Architecture.** We note that the proposed models in Chapter 3 and 5 might be further improved.

    – Our proposed RVL-BERT (Chapter 3) is designed to resemble a VRD classifier such that it only takes in and make prediction for one query object pair each time. This design limits the model efficiency during inference since BERT-based model is notoriously large (160M+ parameters in our case). A promising alternative to this design is a multi-modal BERT taking in all proposals altogether (by formulating into a SGG problem), where self-attention exploits relationships among all object pairs. This design has also been seen in recent SGG works [177, 186].

    – As discussed in Chapter 4, our DLFE makes a data assumption named *Selected Completely At Random* (SCAR). While SCAR has been widely used in PU learning literature [36] and is shown to be useful in our work, it might not be suitable for real-world datasets as it over-simplifies the selection bias by assuming label frequency is independent of the attribute $x$. On the other hand, while *Selected At Random* (SAR) is more relaxed by keeping label frequencies dependent of $x$ (*i.e.*, label frequency is essentially *propensity score*), we normally do not have propensity score provided in datasets. Bekker et al. [5] propose a method for PU learning under only the SAR assumption; however, it still makes additional assumptions such as a label frequency depends on only a *known* subset of attributes $x_e$. This not only remains to be a relatively strong assumption in our view, but also is unrealistic to choose a subset of attributes (which are essentially contextualized visual relation candidate features) to apply in SGG data. Recently, Bekker et al. [7] propose a EM-based method to jointly optimize both a propensity score estimation model and a classification model on a propensity-weighted dataset, which is only based on the data *separability* assumption. We leave this possible extension for future work.

    – The proposed ST-HOI (in Chapter 5) exploits multiple spatial and tem-

poral features including trajectories, human poses and Tube-of-interest pooled features. However, ours approach remains to be simple since we do not incorporate spatial-temporal feature refining via message passing algorithms [120, 178, 28]. In addition, as we discussed in Chapter 5, SeRVo-HOI [47] outperforms our model by utilizing a stronger CNN backbone and skewed-distribution-aware loss functions. There is thus room for improvement of the proposed baselines presented in this thesis; however, we note that we also aim to propose the new task (*i.e.*, VidHOID) along with the benchmark (*i.e.*, VidHOI), which, in conjunction with the baselines, comprise our contribution.

- **Practicability.** We find that there are some practicability issues regarding our ST-HOI.

    – Since we aim to showcase the importance of temporal information for ST-HOI in VidHOID (Chapter 5), we simplify the training setting by only training with ground truth trajectories (while evaluating with predicted/ground truth trajectories). This setting serves as a performance upper bound and is in constrast to that of *video visual relation detection* (VidVRD) [133, 144, 132] where predicted trajectories are being used. We conjecture that, with predicted trajectories, VidHOID models would be more practicable for real-word scenarios. However, we note that accurate trajectory generation remains very challenging because i) the action boundaries labeling could be inconsistent due to inter-annotator disagreement [44, 63], and ii) complex scenes due to constantly changing camera views and intermittent trajectories for some video datasets such as VidOR [131]. Therefore, high label quality and more spatial temporal variance-robust trajectory generation models are very much favorable for real-word use of VidHOID.

    – The "keyframes" used in VidHOI dataset (Chapter 5) are, following the same strategy of AVA dataset [44], uniformly sampled from video clips at a sampling frequency of 1 Hz. Therefore, the sampled keyframes are not guaranteed to be the most important frame or the middlemost frame of the annotated HOIs (trajectories). This might give rise to some issues,

*e.g.*, learning a keyframe-centered model where some neighboring but unrelated frames are taken into account. To tackle this problem, during labeling for each HOI trajectories, one might sample a "keyframe" from the most important or the middlemost frame.

- **Evaluation Fairness.** We note that there are some fairness issue in the metrics we used (following the convention) in our works. While the DLFE proposed in Chapter 4 achieves significantly stronger unbiased SGG performance, this method is not effective for single-class ranking metric, *e.g.*, mean average precision (mAP) used in object detection [101]. This is because that, with the SCAR assumption, DLFE transforms predicted probabilities conditioned only on predicate class but not object pair features. Therefore, the ranking of examples of the same class does not change. We suggest that researchers be extra careful of the metrics when dealing with PU learning datasets, and we refer the interested readers to the more detailed survey on PU learning [6].

In conclusion, we believe that the studies in this thesis pave a way for learning stronger, more informative or temporal-aware visual relation-based representations of visual scenes. We anticipate future work to learn more robust representations by addressing the aforementioned limitations and challenges.

# Publications during PhD Study